\pdfoutput=1
\documentclass[11pt]{article}
\usepackage{fancyreport}
\graphicspath{{figures/}{figures/memory/}{figures/benchmarks/}{figures/appendix/}{figures/method/}}
\makeatletter
\renewcommand{\and}{\end{tabular}\hskip 0.3em plus .1fil\begin{tabular}[t]{c}}
\makeatother

\titlelogos{\includegraphics[height=10mm]{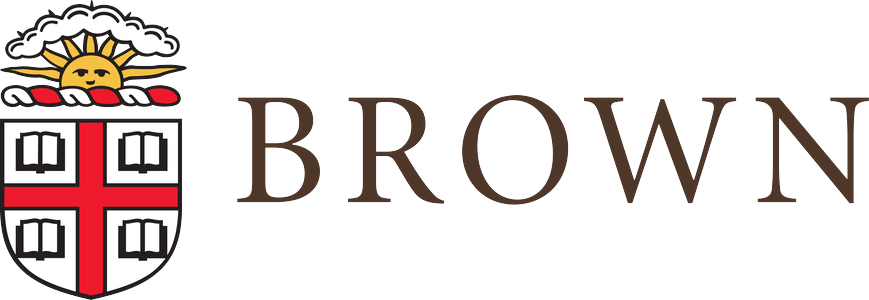}\hspace{11mm}\raisebox{.75mm}{\includegraphics[height=8.5mm]{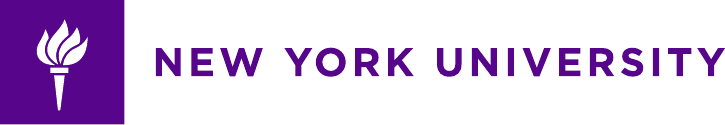}}\hspace{11mm}\includegraphics[height=10mm]{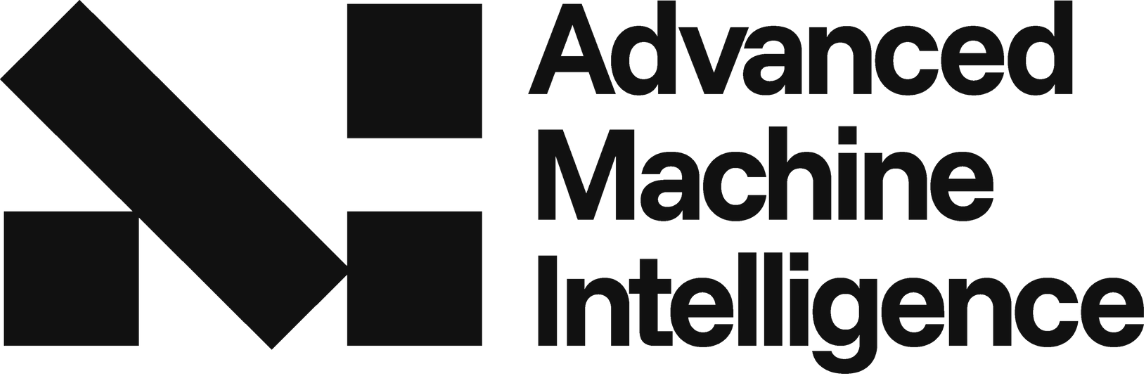}}
\newcommand{\linkicon}[1]{\raisebox{-.17em}{\includegraphics[height=1em]{logos/#1}}}

\usepackage{mathtools}
\usepackage[numbers]{natbib}
\usepackage{colortbl}
\usepackage{multirow}
\usetikzlibrary{arrows.meta,positioning,calc, fit, backgrounds}

\usepackage{pifont}       
\usepackage{xcolor}       
\usepackage{colortbl}     
\usepackage{booktabs}     
\usepackage{array}        
\newcolumntype{L}[1]{>{\raggedright\arraybackslash}p{#1}}
\definecolor{oursrow}{RGB}{230, 245, 230}  
\definecolor{headerblue}{RGB}{30, 60, 120}

\definecolor{cfblue}{RGB}{40, 82, 140}
\definecolor{cfpurple}{RGB}{105, 70, 150}
\definecolor{cfgreen}{RGB}{40, 125, 95}
\definecolor{cforange}{RGB}{190, 110, 35}
\definecolor{cfgray}{RGB}{85, 85, 85}
\definecolor{modfill}{HTML}{D7E3F6}
\definecolor{modline}{HTML}{6E8DC0}
\definecolor{maskblue}{HTML}{6FA8DC}
\definecolor{obsfill}{HTML}{DCDCDC}
\definecolor{obsline}{HTML}{9A9A9A}
\definecolor{stline}{HTML}{8A8A8A}
\definecolor{regfill}{HTML}{F6CFCF}
\definecolor{regline}{HTML}{C97B7B}
\definecolor{regtext}{HTML}{A23232}
\definecolor{msefill}{HTML}{CFE6C6}
\definecolor{mseline}{HTML}{6FA15F}
\definecolor{msetext}{HTML}{2E6B2E}
\definecolor{ipcfill}{HTML}{F8DCB4}
\definecolor{ipcline}{HTML}{D79A4E}
\definecolor{ipctext}{HTML}{9C5A12}
\definecolor{arrowcol}{HTML}{5A5A5A}
\definecolor{panelline}{HTML}{3A3A3A}

\newcommand{\pmark}{\textcolor{orange!90!black}{\ding{51}\kern-0.5ex\textsubscript{*}}}

\definecolor{problemcolor}{RGB}{255,230,230}
\definecolor{solutioncolor}{RGB}{230,255,230}
\definecolor{keyinsightcolor}{RGB}{230,240,255}
\definecolor{warningcolor}{RGB}{255,245,220}

\newcommand{\R}{\mathbb{R}}
\newcommand{\Do}{\mathrm{do}}
\newcommand{\ind}{\perp\!\!\!\perp}

\providecommand{\vo}{\bm{o}}

\AtBeginEnvironment{proposition}{\crefalias{theorem}{proposition}}
\AtBeginEnvironment{corollary}{\crefalias{theorem}{corollary}}
\AtBeginEnvironment{lemma}{\crefalias{theorem}{lemma}}
\AtBeginEnvironment{definition}{\crefalias{theorem}{definition}}
\AtBeginEnvironment{conjecture}{\crefalias{theorem}{conjecture}}
\crefname{definition}{definition}{definitions}
\crefname{conjecture}{conjecture}{conjectures}
\minitocsetup{Intro, Why Counterfactuals?, Natura Non Facit Saltus, Experiments, Conclusion}

\title{Counterfactual World Models:\\[8pt]JEPAs That Reason About ``What If''}
\author{%
  \begin{tabular}[t]{c}
    Randall Balestriero$^*$\\
    \footnotesize Brown University\\
    \footnotesize AMI Labs
  \end{tabular}
  \and
  \begin{tabular}[t]{c}
    Yann LeCun$^*$\\
    \footnotesize New York University\\
    \footnotesize AMI Labs
  \end{tabular}
  \and
  \begin{tabular}[t]{c}
    Kerda Varaku$^*$\\
    \footnotesize Independent Researcher
  \end{tabular}
}
\date{\small
  \href{https://github.com/galilai-group/counterfactual-WM}{\linkicon{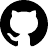}\,\color{navblack}github.com/galilai-group/counterfactual-WM}%
  \hspace{2em}%
  \href{https://huggingface.co/spaces/galilai-group/cwm_demo}{\linkicon{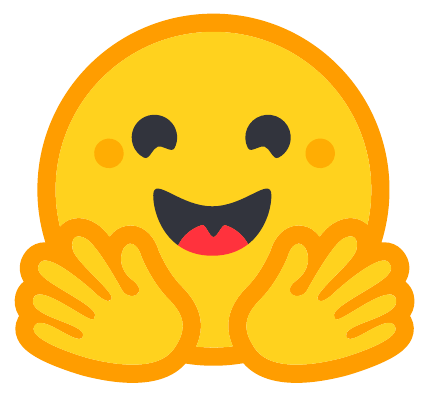}\,\color{navblack}huggingface.co/spaces/galilai-group/cwm\_demo}}

\teaser[\textbf{One observed outcome lets a world model predict what the other
actions would have done.} A boat picks a heading $a'$ while a hidden current, varying slowly in space and time, pushes it; we plot
({\bf top}) its distance to the center of the view and ({\bf bottom}) how far
the current turns its course. Gray: every future the current could produce;
dotted: what would truly have happened in the event that occurred; diamond: the
one outcome observed. A standard joint-embedding predictive architecture (JEPA) world model (green) predicts the average
future and cannot tell which current occurred. Our counterfactual world model (orange)
predicts each alternative as the observed outcome plus the same model's
predicted difference between the two headings, and closely tracks the true
counterfactual, without counterfactual data or retraining. Right: the outcome
observed under the executed heading $a_t$ (top), and what would truly have
happened had the boat headed $-126^\circ$ instead (bottom, gray line), a frame
no model sees. Arrows mark the hidden current field; outlines show where each model
puts the boat (details in \cref{app:teaser}).
\label{fig:teaser}]{\centering
\includegraphics[width=.99\linewidth]{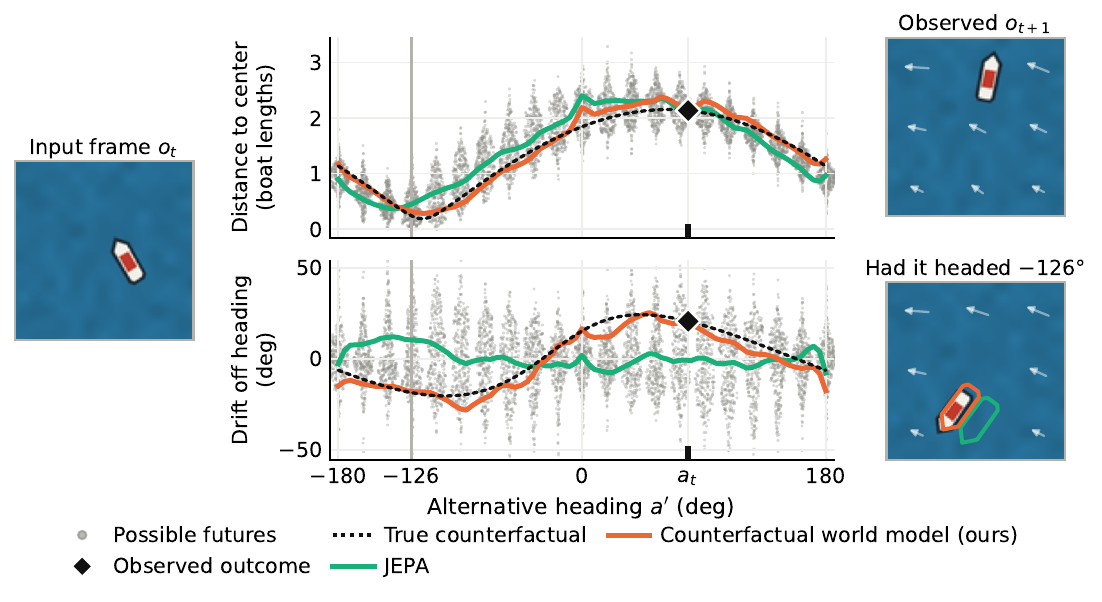}}

\begin{document}
\maketitle
\begin{abstract}
World models operate in a world they cannot fully observe. 
When a hidden disturbance, such as a current pushing a boat, changes the outcome of an action, a counterfactual question follows: what would have happened under a different action, subject to the same disturbance? Its answer separates bad decisions from bad luck. 
We prove that world models trained on factual transitions cannot answer it on their own, whatever their data or capacity. 
For many physical systems, the observed outcome and a simple property of the dynamics suffice: \emph{natura non facit saltus}, nature makes no leaps. 
When actions and disturbances act as separate forces, their interaction is second-order in time, so the counterfactual outcome is, to first order, the observed outcome plus the difference between the world model's predictions under the alternative and factual actions. 
This yields a counterfactual world model with no counterfactual data, extra parameters, or retraining. 
For transitions of duration $h$, we prove an $O(h^2)$ bound on the counterfactual error relative to the disturbance variance, establish a matching minimax lower bound from unpaired factual data, and characterize failures at discontinuities, coordinate folds, and disturbances that scale the command. 
We validate these predictions from pixels with JEPA world models on frozen DINOv2 and masked autoencoder (MAE) encoders. 
On four rendered simulators, our approach improves counterfactual $R^2$ from $38$--$46\%$ to $86$--$96\%$ with the same world models; on a fifth, an underwater vehicle, it reaches $90$--$94\%$ with sixteen times more factual data. 
Four further frozen encoders, DINOv3, SigLIP~2, CLIP and EVA-02, give $81$--$95\%$ on the same simulators. 
In hindsight planning, it finds the action that would have reached the goal in $35$--$69\%$ of events, against $14$--$23\%$ for an oracle planner that knows the simulator but not the outcome.
\end{abstract}

\navsection[Intro]{Introduction}

World models predict the consequences of actions and are becoming a core component of agents that interact with the world. From latent dynamics models for control
\citep{ha2018recurrent,hafner2019learning,hafner2020dream,hafner2025mastering,schrittwieser2020mastering,hansen2024tdmpc2}
to joint-embedding predictive architectures (JEPAs) that predict in the feature space of self-supervised encoders
\citep{lecun2022path,assran2023self,bardes2024vjepa,assran2025vjepa2,zhou2024dinowm,sobal2025learning},
they allow agents to anticipate outcomes before acting.

Yet the real world is never fully observed. Consider the boat of \cref{fig:teaser}: it picks a heading while an unseen current pushes it. Before acting, no world model can know the current it is about to meet, so the best it can do is predict where the boat ends up on average over currents (green). The observed outcome (diamond) lands elsewhere, and in doing so reveals how far the current actually pushed the boat. A new question then arises: \emph{where would the boat have ended up under a different heading, in that same current?} (dotted). This is an \emph{individual counterfactual}. Asked about another heading, a standard world model returns the average over currents again (green) and discards what the outcome revealed. The same question arises whenever a gust of wind or a miscalibrated actuator alters an outcome. Its answer is what one needs to debug a failure, to tell a bad decision from bad luck, to assign credit to an action, and to find in hindsight the action that would have reached the goal, which we call \emph{hindsight planning}.

This question is fundamentally different from factual prediction: as we prove, a world model trained on factual transitions cannot answer it on its own. A world model learns the conditional consequences of actions from observed transitions, whereas an individual counterfactual asks about an alternative action under the \emph{same hidden disturbance}. An unpaired factual dataset does not record outcomes under two actions for the same initial state and realized disturbance. Consequently, even a perfect model of factual transitions does not determine how alternative outcomes are coupled, i.e., which outcome under the alternative action belongs to the observed event: distinct counterfactual worlds can induce exactly the same factual distribution (\cref{thm:individual-cf-minimax,thm:factual-gauge-nonidentification}). Neither a more accurate predictor nor a more expressive generative model resolves this ambiguity, and no factual test can reveal it. Without additional structure, individual counterfactuals are not identified.

For many physical systems, nature provides this structure. \emph{Natura non facit saltus}: nature makes no leaps. When the action and the hidden disturbance enter as separate forces, as the boat's propulsion and the current do, each changes the state gradually, and each alters the effect of the other only through the change of state it has already caused. Their \emph{interaction}, i.e., how much the effect of changing the action depends on the disturbance, is therefore second-order in the transition duration. To first order, changing the action changes the outcome by the same amount in every event, and this amount is what a factual world model predicts between the two actions. Separate channels are an assumption on how the disturbance enters, not a consequence of smoothness: a worn actuator whose effect grows with the command is smooth, yet it interacts with the action at first order (\cref{sec:effects}), and we test this failure in \cref{sec:stress}. Under separate channels, this observation leads to a simple construction. Let $z_t=f(o_t)$ denote the representation of an observation, and let $\mu(z_t,a)$ be a factual world model trained to predict the next representation under action $a$. Given an observed transition $(z_t,a_t,z_{t+1})$, we estimate the outcome under an alternative action $a'$ by
\begin{equation}
    \boxed{
    \widehat z_{t+1}(a')
    =
    z_{t+1}
    +
    \mu(z_t,a')
    -
    \mu(z_t,a_t).
    }
    \label{eq:counterfactual-jepa}
\end{equation}
The observed outcome records the effect of the realized disturbance; the world model predicts the effect of changing the action. The construction requires only two evaluations of an ordinary factual predictor and one residual correction: no paired counterfactual data, additional parameters, explicit disturbance inference, or retraining. We call the resulting predictor a \emph{counterfactual world model}. On the boat of \cref{fig:teaser}, it shifts the world model's prediction for the new heading by the drift that the current caused in the observed event (from the green to the orange outline in the bottom-right frame), and follows the true counterfactual (orange against dotted).

In additive worlds, residual reuse is the unique counterfactual rule that is exact for every factual transition (\cref{prop:residual-unique}). Outside this regime, its error is governed by the interaction between actions and hidden disturbances, quantified by the mixed derivative of the dynamics (\cref{thm:interaction-bound}). For short physical transitions with separate action and disturbance channels, this interaction is second-order in the time step $h$, leaving only an $O(h^2)$ fraction of the disturbance uncertainty unexplained (\cref{thm:short-time-residual,cor:relative-rate}). This guarantee extends through smooth representations and to systems with hidden velocities (\cref{cor:inertial-residual}). The rate cannot be uniformly improved: we establish a matching minimax lower bound, with the sharp leading constant, for estimators using unpaired factual transitions and one observed outcome (\cref{thm:minimax-rate}). Residual reuse therefore achieves the best possible asymptotic rate under the stated information and physical assumptions. How short is short enough depends on the system, and we measure it (\cref{sec:theory-checks}). The same analysis identifies the limits of residual reuse: disturbances that scale the command, discontinuous contacts, and folded representations can invalidate it.

Residual correction is familiar in control theory, where disturbance observers, offset-free model predictive control, and iterative learning control reuse errors from observed trajectories
\citep{ohishi1987microprocessor,muske2002disturbance,pannocchia2003disturbance,arimoto1984bettering,bristow2006survey}.
Likewise, additive-noise structural causal models provide a classical route to individual counterfactuals
\citep{pearl2009causality,peters2017elements}.
What has remained unclear is why this construction should be justified for learned world models, what can be recovered without paired counterfactual observations, and whether the principle survives nonlinear representations learned from pixels. Our analysis connects the identification problem of causal inference, the disturbance structure of control, and the representations of modern world models. In particular, factual training of the predictor is sufficient under the stated shared-input conditions, without additional propensity weighting or encoder retraining
(\cref{prop:irreducible,cor:frozen-encoder-mean,prop:shared-inputs}).

We test these claims using rendered physical simulators and frozen self-supervised image encoders, DINOv2 \citep{oquab2023dinov2} in two sizes and masked autoencoders (MAE) \citep{he2022masked}, and repeat the main comparison with four further frozen encoders, DINOv3 \citep{simeoni2025dinov3}, SigLIP~2 \citep{tschannen2025siglip2}, CLIP \citep{radford2021learning} and EVA-02 \citep{fang2023eva02}. We score predicted physical quantities, such as joint angles and positions, by their $R^2$ against the true counterfactual outcome, i.e., the fraction of its variance they explain (\emph{physical counterfactual $R^2$}). On four environments, with DINOv2 and MAE, counterfactual world models improve physical counterfactual $R^2$ from $38$--$46\%$ to $86$--$96\%$ using the same factual predictors, with the correction applied after a readout of these quantities fitted on factual frames; applied to the image features themselves, which needs no state labels, it reaches $73$--$94\%$. On the fifth, the BlueROV2 underwater vehicle, they reach $90$--$94\%$ with sixteen times more factual data. They outperform even an oracle predictor that knows the true simulator and its full state, but not the disturbance that occurred, and therefore predicts the average outcome. With the four further encoders, correction after the readout reaches $81$--$95\%$ on the first four environments. In hindsight planning, the counterfactual model reaches the goal in $35$--$69\%$ of events, compared with $14$--$23\%$ for the true-simulator planner without access to the observed outcome. The experiments also recover the predicted failure modes, including contacts, folded coordinates, and disturbances that multiply the command. These results demonstrate that the observed outcome can be more informative for individual counterfactual reasoning than an otherwise perfect predictive model.

Our contributions are threefold. 
\begin{itemize}
\item \textbf{First}, we establish the nonidentification of individual counterfactuals from factual world models, including a minimax impossibility result and an explicit gauge ambiguity that persists even with exact factual likelihoods: the latent noise of a generative model can be relabeled separately for each action without changing any factual prediction (\cref{sec:need}).
\item \textbf{Second}, we derive residual reuse from the short-time structure of physical dynamics, prove its exactness under additivity, characterize its interaction error and failure modes, and establish its minimax rate-optimality from unpaired factual data (\cref{subsec:short-time-residual,thm:minimax-rate}).
\item \textbf{Third}, we validate the resulting counterfactual world models from pixels, across seven frozen image encoders, physical disturbances, hindsight-planning tasks, and controlled violations of the assumptions (\cref{sec:residual-experiments}).
\end{itemize}

\navsection[Why Counterfactuals?]{World Models Need Individual Counterfactuals}
\label{sec:need}

World models and causal models describe the same underlying object from different perspectives: how a system responds to actions in the presence of unobserved disturbances. We first distinguish the quantities learned by ordinary world models from the individual counterfactuals needed after observing an outcome (\cref{sec:effects}). We then show why individual counterfactuals cannot be identified from factual transitions alone, even by a perfect generative model (\cref{sec:counterfactual}). Finally, we identify the additional structure under which counterfactual reasoning becomes possible (\cref{sec:structure-needed}).

\subsection{World Models Estimate Average Effects, Surprises Ask for Individual Ones}
\label{sec:effects}

\paragraph{From physical dynamics to world models.}
Let $s_t$ denote the physical state of a system, $a_t$ the action applied at time $t$, and $\varepsilon_t$ an unobserved disturbance. The transition obeys
\begin{equation}
    s_{t+1}=F_s(s_t,a_t,\varepsilon_t).
    \label{eq:structural-transition}
\end{equation}
This single equation admits three interpretations. In control theory, $F_s$ describes a disturbed dynamical system, or \emph{plant}. In causal inference, it is a structural causal equation: the action is an intervention, and $\varepsilon_t$ is the exogenous variable held fixed when constructing a counterfactual \citep{pearl2009causality}. In world modeling, it is the unknown physical mechanism that a learned predictor attempts to approximate. Connections between dynamical systems and structural causal models have been formalized through differential equations \citep{mooij2013ode,rubenstein2018deterministic,peters2022causal}.

The physical state is generally not observed directly. Instead, an agent receives an observation $o_t$, such as an image, and represents it through $z_t=f(o_t)$. A representation-based world model learns a predictor $\mu(z_t,a_t)$ of the next representation $z_{t+1}$. Joint-embedding predictive architectures (JEPAs) predict in this feature space rather than reconstructing pixels \citep{lecun2022path}. The encoder can be learned through a joint-embedding objective or obtained from another pretraining method, including one that reconstructs pixels such as a masked autoencoder. Since the world model encodes the frames and predicts in feature space, it is a JEPA either way.

A structural transition $z_{t+1}=F(z_t,a_t,\varepsilon_t)$ at the representation level requires that $z_t$ contain sufficient information for the relevant dynamics. This is not automatic: a single image may hide velocity or other state variables. We distinguish the physical mechanism $F_s$ from its representation-level counterpart and state the required sufficiency assumptions where they are used.\footnote{\citet{bear2023cwm} use ``counterfactual world modeling'' for masked prediction in visual perception. Here the term refers to counterfactual outcomes under alternative actions and shared disturbances.}

\paragraph{Prediction, intervention, and counterfactuals.}
The structural equation supports three distinct queries. Ordinary \emph{prediction} asks what outcome is likely under the behavior policy that generated the observations, $P(s_{t+1}\mid s_t)$. An \emph{interventional} query asks what would happen if an action were imposed, $P(s_{t+1}\mid s_t,\Do(a_t=a'))$, averaging over the disturbances that might occur. This is the question answered by an ideal action-conditioned world model before acting. An \emph{individual counterfactual} asks a different question. Suppose the system executed $a_t$, experienced a disturbance $\varepsilon_t$, and produced the observed outcome $s_{t+1}$. What would the outcome have been under the alternative action $a'$, while the initial state and disturbance remained unchanged? The answer is
\[
    s_{t+1}(a')=F_s(s_t,a',\varepsilon_t),
\]
where $\varepsilon_t$ is the same disturbance that produced the observed transition. The distinction is essential because an intervention considers possible disturbances, whereas an individual counterfactual concerns the disturbance that actually occurred. In causal terminology, counterfactual reasoning uses the observed outcome to infer information about the realized exogenous variables before changing the action.

\paragraph{From conditional to individual effects.}
The distinction can be expressed using potential outcomes. Let $Y(a)$ be the outcome that would result under action $a$, and let $Z$ denote pre-action covariates available to the model. Three quantities are commonly distinguished:
\begin{align}
    \mathrm{ATE}(a',a)
    &=\E[Y(a')-Y(a)]
    &&\text{(average treatment effect)},\label{eq:ate-definition}\\
    \mathrm{CATE}(a',a\mid z)
    &=\E[Y(a')-Y(a)\mid Z=z]
    &&\text{(conditional average treatment effect)},\label{eq:cate-definition}\\
    \mathrm{ITE}_i(a',a)
    &=Y_i(a')-Y_i(a)
    &&\text{(individual treatment effect)}.\label{eq:ite-definition}
\end{align}
The ATE averages over the population, the CATE averages over events sharing the same pre-action information, and the ITE concerns one particular event. Estimating conditional effects from observations is a central problem in causal representation learning and treatment-effect estimation
\citep{shalit2017estimating,lim2018forecasting,bica2020estimating,melnychuk2022causal}. To connect these definitions to a world model, suppose actions are conditionally exchangeable given the model input $Z$, i.e., once $Z$ is known, the action taken carries no further information about the potential outcomes (no hidden confounding).\footnote{For example, when the behavior policy selects actions using only $Z$ and any action randomness is independent of the system's hidden disturbances.} Suppose also that actions have sufficient overlap, i.e., the logged actions cover every queried action at every value of $Z$. Under these conditions, squared-loss prediction of $Y$ from $(Z,A)$ identifies the conditional interventional mean
\[
    \mu(z,a)=\E[Y(a)\mid Z=z].
\]
Consequently, evaluating the predictor at two actions gives
\[
    \mu(z,a')-\mu(z,a)
    =\mathrm{CATE}(a',a\mid z).
\]
Without these conditions, a factual predictor instead learns an observational conditional mean, which can differ from the conditional interventional mean. We return to the relevant shared-input conditions in \cref{prop:irreducible,prop:shared-inputs}.

Even when the conditional interventional means are identified exactly, they do not determine the individual effect. The CATE averages over disturbances compatible with $Z=z$, while the ITE compares two outcomes under the same realized disturbance. Equivalently, the ATE and CATE depend on marginal distributions of potential outcomes, whereas the ITE depends on how those outcomes are \emph{coupled} across actions, i.e., on their joint distribution, which says which outcome under $a'$ belongs to the same event as a given outcome under $a$. This missing coupling is the fundamental obstacle to individual counterfactual world modeling.

\paragraph{Individual and average effects differ only through interaction.}
We say that the action and the disturbance \emph{interact} when the effect of changing the action depends on which disturbance occurred. Without interaction, the ITE is the same in every event and therefore equals the CATE; with interaction, the two differ by an amount set by the unobserved disturbance. For the boat of \cref{fig:teaser}, a uniform current would drift the boat by the same amount whatever its heading, so turning would move the boat's position by the same amount in every event, and its counterfactual position would be the observed one plus the CATE. Its current varies slowly in space, so the drift depends only slightly on where the boat goes, and the interaction is small; \cref{sec:where-transport} measures how it grows as the eddies shrink. A worn actuator that delivers only a fraction of the command is the opposite case: how far a larger command moves the system depends on the wear, so no single correction is right for every event. Interaction is what separates the coupling from a plain shift: without it, the coupling reduces to a shift by the CATE, which factual data identify, and the interaction measures how far the true coupling departs from that shift. Factual data alone cannot tell the two cases apart: the two mechanisms of the next subsection (\cref{sec:counterfactual}) share their factual laws, yet the first has no interaction and the second does. \Cref{subsec:short-time-residual} shows when physics keeps the interaction small.

\subsection{No Factual World Model Identifies Individual Counterfactuals}
\label{sec:counterfactual}
\label{subsec:cf-information-source}

Can a sufficiently accurate or expressive world model recover the missing coupling and estimate individual treatment effects? In general, it cannot. Unpaired factual data reveal the outcome distribution associated with each action, but do not record outcomes under two actions for the same realized disturbance. We first establish that this ambiguity persists with infinite data, even under perfect generative world modeling.

\paragraph{Factual marginals do not determine counterfactuals.}
Consider the potential outcomes $Y(a)$ and $Y(a')$. An arbitrarily large factual dataset can identify their marginal laws (we write \emph{law} for the probability distribution of a random variable) under the appropriate identification and coverage conditions. It does not, however, reveal their joint law. Different joint distributions can share exactly the same marginals while assigning different outcomes to the same individual under alternative actions. For example, let $U$ be uniformly distributed on $[0,1]$. Consider two mechanisms with the same outcome under action $a$,
\[
    Y(a)=U,
\]
but different outcomes under $a'$:
\[
    Y(a')=U
    \qquad\text{or}\qquad
    Y(a')=1-U.
\]
In both mechanisms, the outcome is uniformly distributed under either action. Their factual and interventional distributions are therefore identical. Yet observing $Y(a)=u$ gives counterfactual outcomes $u$ and $1-u$, respectively; neither factual distribution distinguishes the two. The following proposition states why these marginals do not identify individual treatment effects.

\begin{proposition}[Marginals do not identify counterfactual matchings]
\label{prop:marginals-not-counterfactual}
Two structural models can agree on every action-conditional outcome law while predicting different alternatives for the same observed transition.
(Proof in \cref{proof:marginals-not-counterfactual}.)
\end{proposition}

The ambiguity is not due to limited data, imperfect optimization, or insufficient model capacity. Even when all interventional laws are known exactly, the minimax risk for estimating an individual counterfactual remains bounded away from zero, as does that of the individual treatment effect (\cref{thm:individual-cf-minimax,cor:ite-minimax}). The two mechanisms can also be smooth; smoothness of the factual outcome laws alone does not identify their coupling. This is an instance of the causal hierarchy separating interventions from counterfactuals \citep{bareinboim2022pearl}, and of the classical nonidentification of the joint distribution of potential outcomes from experimental marginals \citep{heckman1997heterogeneity}. A deterministic world model predicting the conditional mean therefore cannot recover the individual counterfactual merely by becoming more accurate. But perhaps a generative world model could do better, by sampling a latent disturbance and reusing it under another action. The next result shows why factual training does not identify this operation either.

\paragraph{Perfect generative models still leave the coupling free.}
Suppose the factual outcome law at state $s$ and action $a$ is represented as the push-forward
\[
    P_{s,a}=(T_{s,a})_\#\nu,
\]
where $\nu$ is a reference noise distribution and $T_{s,a}$ is a measurable bijection. That is, an outcome is generated by drawing noise from $\nu$ and passing it through $T_{s,a}$; $P_{s,a}$ is the resulting law of the outcome, the \emph{push-forward} of $\nu$ by $T_{s,a}$. An invertible generative model can infer the latent noise associated with an observed outcome $y$ through $T_{s,a}^{-1}(y)$ and decode it under an alternative action $a'$. Its implied counterfactual is thus the composition ($\circ$) of these two maps,
\[
    T_{s,a'}\circ T_{s,a}^{-1}(y).
\]
This appears to solve the individual counterfactual problem: infer the noise from what happened, then reuse it to predict what would have happened. The difficulty is that the inferred noise has no uniquely identified correspondence across actions. Consider any family of $\nu$-preserving bijections $G_{s,a}$. Reparameterizing the noise separately for each action leaves the factual laws unchanged, but can change the counterfactual mapping, as we formalize below.

\begin{theorem}[Factual gauge nonidentification]
\label{thm:factual-gauge-nonidentification}
Let $P_{s,a}=(T_{s,a})_\#\nu$ with measurable bijections $T_{s,a}$. For every measurable family of $\nu$-preserving bijections $G_{s,a}$, the mechanisms
\[
    \widetilde T_{s,a}=T_{s,a}\circ G_{s,a}
\]
induce exactly the same factual laws, but their counterfactual coupling is
\[
    \widetilde T_{s,a'}\circ\widetilde T_{s,a}^{-1}
    =
    T_{s,a'}\circ G_{s,a'}\circ
    G_{s,a}^{-1}\circ T_{s,a}^{-1}.
\]
Factual likelihood, calibration, or any criterion depending only on $\{P_{s,a}\}$ therefore cannot identify the same-disturbance counterfactual. The gauge cancels when it is action-independent. Within the class of invertible transports from the common reference law, every factual-law-preserving reparameterization is of this form.

(Proof in \cref{sec:factual-gauge-revision}.)
\end{theorem}

The theorem identifies the precise missing information: factual training constrains the distribution of latent noise at each action, but not how a particular realization of that noise corresponds across actions. This is an action-dependent \emph{gauge freedom}, a term borrowed from physics for a change of description, here a relabeling of the noise, that leaves every observable prediction unchanged. The freedom is real whenever $\nu$ admits a $\nu$-preserving bijection other than the identity, as any continuous noise law does, e.g., through a reflection of a centered Gaussian; only deterministic dynamics, with $\nu$ a point mass, leave nothing to relabel. A normalizing flow with an exact likelihood can infer a latent variable from an observed transition, yet factual data alone do not establish that this latent variable represents the same physical disturbance under a different action. Related nonidentification results for bijective causal models appear in
\citet{nasresfahany2023counterfactual,nasresfahany2023nonidentifiability}; see also neural causal models in \citet{xia2023neural}. The same limitation applies beyond invertible generators. A diffusion or latent-variable world model may sample realistic outcomes under alternative actions, but its factual objective does not, by itself, determine which alternative outcome corresponds to the particular event that was observed. Individual counterfactuals therefore require information about the correspondence across actions that factual predictive accuracy does not measure.

\subsection{Every Counterfactual Needs Structure Beyond Factual Data}
\label{sec:structure-needed}

The preceding results establish a boundary between learning factual distributions and identifying individual counterfactuals. Factual data can identify interventional quantities under the appropriate assumptions, but the coupling between potential outcomes requires additional structure. This structure may come from paired observations, a known mechanism, restrictions on the disturbance, or assumptions on how actions affect the system.

\begin{table}[t]
\centering\small
\setlength{\tabcolsep}{4pt}
\renewcommand{\arraystretch}{1.15}
\begin{tabular}{L{4.6cm}L{3.5cm}L{3.4cm}L{3.4cm}}
\toprule
\textbf{Query from factual transitions} &
\textbf{Additional requirement} &
\textbf{Status} &
\textbf{Where} \\
\midrule
Interventional mean $\E[Y(a')\mid H]$ &
Exchangeability, overlap &
Identified by factual regression &
\cref{prop:irreducible,prop:shared-inputs}; \citep{rubin1974estimating,hernan2020causal,shimodaira2000improving}\\
Interventional distribution $P(Y(a')\mid H)$ &
Same &
Identified &
\citep{pearl2009causality,peters2017elements}\\
\addlinespace
Individual counterfactual, unrestricted coupling &
None &
\textbf{Not identified}; positive minimax risk &
\cref{thm:individual-cf-minimax}; \citep{heckman1997heterogeneity,bareinboim2022pearl}\\
\ldots\ with exact factual likelihood &
None &
\textbf{Not identified}; action-dependent gauge &
\cref{thm:factual-gauge-nonidentification}; \citep{nasresfahany2023counterfactual}\\
\addlinespace
\ldots\ under additive response &
Shared additive disturbance &
Exact by residual reuse &
\cref{eq:additive-response}; \citep{hoyer2008nonlinear,peters2017elements}\\
\ldots\ under monotone scalar response &
Rank preservation &
Identified by rank transport &
\citep{chernozhukov2005iv,athey2006identification,nasresfahany2023counterfactual,wu2025learning}\\
\ldots\ under smooth force-like physics &
Short duration, sufficient features &
Residual error $O(h^2)$ relative to disturbance &
\cref{thm:interaction-bound,thm:short-time-residual,cor:relative-rate}\\
\ldots\ with paired alternatives or known mechanism &
Paired data or simulator &
Identified &
\citep{baradel2020cophy,buesing2019woulda,oberst2019counterfactual}\\
\addlinespace
\ldots\ through contacts, multiplicative faults, folded coordinates &
Additional structure needed &
Residual reuse can fail &
\cref{sec:stress,app:short-time-residual}\\
\bottomrule
\end{tabular}
\caption{\textbf{What factual world models identify, and what individual counterfactuals require.}
Under suitable exchangeability and coverage assumptions, factual transitions identify interventional quantities, but not the correspondence between outcomes under different actions. Individual counterfactuals become identifiable through additional assumptions or paired observations. For smooth physical dynamics, residual reuse instead provides an approximation with a controlled error and explicit failure modes.}
\label{tab:map}
\end{table}

\paragraph{How existing methods resolve the ambiguity.}
Different approaches to counterfactual reasoning provide the missing coupling in different ways. Structural causal models impose restrictions on how exogenous variables generate outcomes \citep{pearl2009causality,peters2017elements}. Additive-noise models provide an especially simple case: the observed residual identifies the disturbance contribution, which can then be reused under a different action \citep{hoyer2008nonlinear}. Monotone scalar responses identify counterfactuals through rank preservation: if a scalar outcome increases with a scalar disturbance under every action, an event keeps its quantile in the outcome distribution across actions, so the counterfactual is the outcome at the same quantile under $a'$ (rank transport). Suitable bijective mechanisms can support more general transport constructions
\citep{chernozhukov2005iv,athey2006identification,nasresfahany2023counterfactual,wu2025learning}.
Alternatively, paired observations or a known simulator can provide the correspondence between actions directly
\citep{baradel2020cophy,janny2022filtered,buesing2019woulda,oberst2019counterfactual,pawlowski2020dscm}.

Counterfactual reasoning has also been studied in reinforcement learning and world modeling, including data augmentation
\citep{lu2020sample,pitis2020counterfactual,pitis2022mocoda},
credit assignment \citep{mesnard2021counterfactual},
explanation \citep{tsirtsis2021counterfactual},
and counterfactual action-sequence optimization
\citep{tsirtsis2023finding,haugh2023counterfactual}.
Observed trajectory residuals have been reused to correct model errors in model-based control
\citep{abbeel2006using}.
More recent approaches study counterfactual prediction in driving world models through abduction, action, and prediction
\citep{zhang2026driving},
noise-coupled branching in interactive video world models
\citep{ma2026twin},
and action-discriminative latent models for counterfactual model predictive control
\citep{qiu2026adwm}.
These works differ in their assumptions, supervision, and target counterfactual queries, but share (i) the need for some correspondence between alternative outcomes, (ii) expert knowledge in some structural model, or (iii) access to simulators for paired data generation.

\paragraph{In practice, the coupling is the hard requirement.}
Exchangeability and overlap hold whenever the logged actions depend only on what the world model sees, plus independent exploration that covers the queried actions (\cref{prop:shared-inputs}); an operator who perceives more than is recorded can break them. The coupling requirements of \cref{tab:map} are harder to meet. Paired outcomes and known mechanisms are rarely available outside simulation, and factual data can never confirm additivity or rank preservation, since the coupling they assert is not identified (\cref{thm:factual-gauge-nonidentification}). We therefore do not postulate additivity itself: \cref{subsec:short-time-residual} derives it, up to second order, from one physical assumption, separate action and disturbance channels, with an explicit error bound, and \cref{sec:stress} tests where it breaks.

\paragraph{Physical dynamics constrain counterfactual outcomes.}
Physical dynamics can constrain the correspondence between outcomes under different actions without learning it or using a known simulator. When actions and disturbances enter through separate force-like channels, the effect of changing an action is approximately independent of the realized disturbance over a sufficiently short transition. The approximation error is governed by their interaction, rather than by the magnitude of the disturbance itself. This distinction matters for learned world models. Instead of attempting to infer the true exogenous variable or identify a generative coupling from factual data, we can reuse the discrepancy between the observed outcome and the factual prediction. The world model then provides only the change attributable to the alternative action. The resulting counterfactual depends on an observable residual and two ordinary predictions, not on a learned correspondence between latent noise variables.

The next section formalizes this principle. We first establish why residual reuse is exact in additive systems, then quantify the error introduced by action--disturbance interactions. Finally, we show that the short-time structure of physical dynamics controls this interaction and that the resulting approximation achieves the minimax-optimal rate from unpaired factual data (\cref{subsec:short-time-residual}).

\navsection[Natura Non Facit Saltus]{Natura Non Facit Saltus: Counterfactuals from a World Without Leaps}
\label{subsec:short-time-residual}

The previous section established that factual transition laws do not identify individual counterfactuals (\cref{sec:need}). Identification requires additional structure on how the same disturbance affects outcomes under different actions. Here we consider physical dynamics in which actions and disturbances enter through separate force-like channels. Their effects add in the instantaneous dynamics, while their interaction accumulates over time. We first derive residual reuse and characterize its exactness under additive responses. We then express its error in terms of the action--disturbance interaction. For smooth physical systems, this interaction is second-order in the transition duration, yielding a minimax-optimal counterfactual approximation. Finally, we establish the conditions under which a factual world model trained by squared loss provides the required action contrast.

\subsection{Residual Reuse for Individual Counterfactuals}
\label{sec:bridge}
\label{sec:recipe}

Consider an observed transition $(H_t,a_t,z_{t+1})$, where $H_t$ contains the current observation embedding and, optionally, past embeddings and actions. Let $\mu(H_t,a)$ predict the next embedding under action $a$. The counterfactual model uses the observed outcome and the predicted difference between the factual and alternative actions.

\begin{definition}[Counterfactual world model]
\label{def:counterfactual-jepa}
Given a factual predictor $\mu$ and an observed transition $(H_t,a_t,z_{t+1})$, the counterfactual world model estimates the outcome under action $a'$ by
\begin{equation}
    \widehat z_{t+1}(a')
    =
    z_{t+1}
    +
    \mu(H_t,a')
    -
    \mu(H_t,a_t).
    \label{eq:counterfactual-jepa-definition}
\end{equation}
For a frozen readout $\pi$ mapping embeddings to physical quantities, we may instead apply the correction in the readout coordinates:
\begin{equation}
    \widehat s_{t+1}(a')
    =
    \pi(z_{t+1})
    +
    \pi(\mu(H_t,a'))
    -
    \pi(\mu(H_t,a_t)).
    \label{eq:counterfactual-jepa-chart}
\end{equation}
An alternative is to decode the transported embedding, $\pi(\widehat z_{t+1}(a'))$. The two constructions generally differ when $\pi$ is nonlinear.
A nonlinear readout can introduce error even when the factual prediction $\mu$ is exact, so guarantees in readout coordinates also require control of the error in the decoded action contrast.
\end{definition}

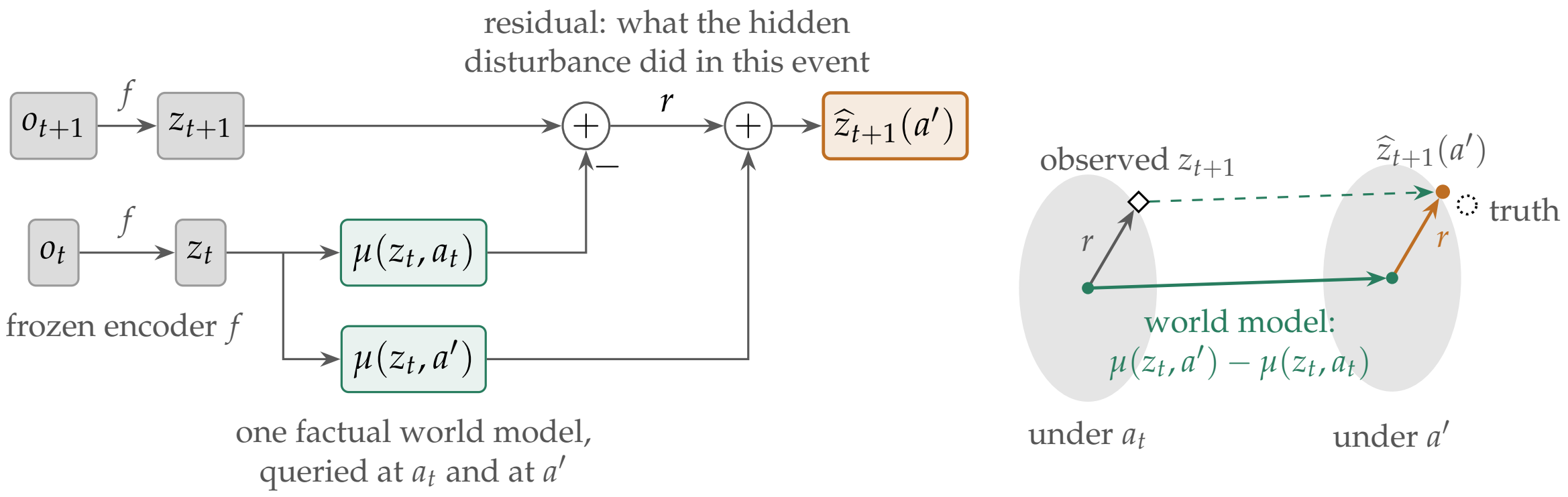
\begin{figure}[t]
\centering
\begin{tikzpicture}[font=\small, >={Stealth[length=2mm]},
  box/.style={rounded corners=2pt, minimum height=6.5mm, inner sep=3pt, line width=.6pt},
  obs/.style={box, draw=obsline, fill=obsfill},
  wm/.style={box, draw=cfgreen, fill=cfgreen!10},
  cf/.style={box, draw=cforange, fill=cforange!14, line width=.9pt},
  sum/.style={circle, draw=arrowcol, line width=.6pt, inner sep=0pt, minimum size=4.6mm},
  flow/.style={->, draw=arrowcol, line width=.6pt},
  note/.style={font=\footnotesize, text=cfgray, align=center}]
\node[obs] (otp) at (0,1.25) {$o_{t+1}$};
\node[obs] (ztp) at (1.45,1.25) {$z_{t+1}$};
\node[obs] (ot) at (0,0) {$o_t$};
\node[obs] (zt) at (1.45,0) {$z_t$};
\node[wm] (m1) at (3.55,0) {$\mu(z_t,a_t)$};
\node[wm] (m2) at (3.55,-1.05) {$\mu(z_t,a')$};
\node[sum] (s1) at (5.25,1.25) {$+$};
\node[sum] (s2) at (6.85,1.25) {$+$};
\node[cf] (out) at (8.3,1.25) {$\widehat z_{t+1}(a')$};
\draw[flow] (otp) -- node[above, note]{$f$} (ztp);
\draw[flow] (ot) -- node[above, note]{$f$} (zt);
\draw[flow] (zt) -- (m1);
\draw[flow] ($(zt.east)!.5!(m1.west)$) |- (m2);
\draw[flow] (ztp) -- (s1);
\draw[flow] (m1) -| node[pos=.92, right, inner sep=1.5pt]{$-$} (s1);
\draw[flow] (s1) -- node[above]{$r$} (s2);
\draw[flow] (m2) -| (s2);
\draw[flow] (s2) -- (out);
\node[note, above=1.5mm of s1, xshift=8mm] {residual: what the hidden\\disturbance did in this event};
\node[note, below=1.2mm of m2] {one factual world model,\\queried at $a_t$ and at $a'$};
\node[note, below=1.2mm of ot, xshift=7mm] {frozen encoder $f$};
\begin{scope}[xshift=10.2cm, yshift=-.35cm]
  \coordinate (A) at (0,0);
  \coordinate (B) at (.5,.85);
  \coordinate (C) at (3.0,.1);
  \coordinate (D) at ($(C)+(B)-(A)$);
  \coordinate (Y) at ($(D)+(.24,-.13)$);
  \fill[obsfill!75] (A) ellipse (.68 and 1.12);
  \fill[obsfill!75] (C) ellipse (.68 and 1.12);
  \draw[->, cfgreen, line width=.9pt, shorten >=1.5pt] (A) -- (C);
  \draw[->, cfgreen, line width=.6pt, dashed, shorten >=2.5pt, shorten <=3pt] (B) -- (D);
  \draw[->, arrowcol, line width=.8pt, shorten >=2.5pt] (A) -- (B);
  \draw[->, cforange, line width=.9pt, shorten >=2pt] (C) -- (D);
  \fill[cfgreen] (A) circle (1.7pt);
  \fill[cfgreen] (C) circle (1.7pt);
  \draw[black, fill=white, line width=.6pt] (B) +(0,2.8pt) -- +(2.8pt,0) -- +(0,-2.8pt) -- +(-2.8pt,0) -- cycle;
  \draw[black, densely dotted, line width=.7pt] (Y) circle (2.6pt);
  \fill[cforange] (D) circle (2pt);
  \node[note, anchor=north] at (0,-1.17) {under $a_t$};
  \node[note, anchor=north] at (3.0,-1.17) {under $a'$};
  \node[note, anchor=north, text=cfgreen] at ($(A)!.5!(C)+(0,-.12)$) {world model:\\$\mu(z_t,a')-\mu(z_t,a_t)$};
  \node[note, anchor=east] at ($(A)!.5!(B)+(-.05,.02)$) {$r$};
  \node[note, anchor=west, text=cforange] at ($(C)!.5!(D)+(.05,-.04)$) {$r$};
  \node[note, anchor=south] at ($(D)+(-.1,.1)$) {$\widehat z_{t+1}(a')$};
  \node[note, anchor=south] at ($(B)+(0,.1)$) {observed $z_{t+1}$};
  \node[note, anchor=west] at ($(Y)+(.08,-.04)$) {truth};
\end{scope}
\end{tikzpicture}
\caption{\textbf{The observed outcome carries the event, the world model carries the action.} Left: the counterfactual world model of \cref{def:counterfactual-jepa}. The observed outcome supplies the residual $r=z_{t+1}-\mu(z_t,a_t)$, what the hidden disturbance did in this event; the factual world model, queried at the executed and at the alternative action, supplies what changing the action does. Nothing is trained for the counterfactual. Right: the same rule in outcome space. Gray: the outcomes that the disturbance could produce under each action, centered at the world model's predictions (green). The observed outcome (diamond) is one of them; residual reuse adds the same residual to the prediction under $a'$ (orange). Its gap to the true counterfactual (dotted) is the gap between the individual and the conditional average effect, which vanishes without interaction (\cref{eq:residual-contrast-identity}). Readout transport (\cref{eq:counterfactual-jepa-chart}) applies the same arithmetic after a readout $\pi$ of physical quantities.}
\label{fig:recipe}
\end{figure}

\Cref{fig:recipe} illustrates the rule. Both constructions use the factual predictor without retraining but differ in the coordinates in which the residual is transported. As shown below, these coordinates determine the size of the counterfactual approximation error. They also differ in what they need: \eqref{eq:counterfactual-jepa-definition} uses only the encoder and the factual predictor, and therefore applies to any pretrained representation, whereas \eqref{eq:counterfactual-jepa-chart} needs a readout fitted on physical-state labels of factual frames, in exchange for coordinates in which the interaction can be much smaller (\cref{sec:where-transport}). To analyze the rule, fix a source representation $z$ and a transition duration $h>0$ and denote by $Y_h(a)=F_h(z,a,\varepsilon)$ the outcome under action $a$ and realized disturbance $\varepsilon$. Counterfactual outcomes $Y_h(a)$ and $Y_h(a')$ are defined using the same $\varepsilon$. Assume that the disturbance has a common conditional law $\nu_z$ across actions, and that the queried actions have factual support. The population squared-loss predictor is then $\mu_h(z,a)=\E_{\nu_z}[F_h(z,a,\varepsilon)]$.
The required identification conditions for learning this mean from factual transitions are given in \cref{sec:interventional}.
Given the observed outcome $y=Y_h(a)$, define residual reuse with the exact factual predictions
\begin{equation}
    T_R(z,a,y,a')
    =
    y+\mu_h(z,a')-\mu_h(z,a).
    \label{eq:residual-transport}
\end{equation}
The correction keeps the factual residual $y-\mu_h(z,a)$ and adds it to the predicted outcome under $a'$. In causal inference, this is the counterfactual construction associated with an additive-noise model \citep{pearl2009causality,peters2017elements}. In control, related residual corrections appear in disturbance observers and offset-free control \citep{muske2002disturbance,pannocchia2003disturbance}. Let $\Delta_h=Y_h(a')-Y_h(a)$ be the individual action contrast, i.e., the change in outcome caused by switching from $a$ to $a'$ in the same event. Since
$\mu_h(z,a')-\mu_h(z,a)=\E_{\nu_z}\Delta_h$,
substitution into \eqref{eq:residual-transport} gives the exact identity
\begin{align}
    T_R-Y_h(a')
    &=
    \E_{\nu_z}\Delta_h-\Delta_h,
    \label{eq:residual-contrast-identity}\\
    \E_{\nu_z}\|T_R-Y_h(a')\|^2
    &=
    \operatorname{tr}\operatorname{Var}_{\nu_z}(\Delta_h).
    \label{eq:residual-contrast-risk}
\end{align}
Here $\E_{\nu_z}$ averages over the disturbance $\varepsilon\sim\nu_z$, $\operatorname{Var}_{\nu_z}$ is the covariance matrix under that law (written $\operatorname{Cov}_{\nu_z}$ for $\varepsilon$ itself), and $\operatorname{tr}$ is the matrix trace, so $\operatorname{tr}\operatorname{Var}_{\nu_z}$ sums the variances of all coordinates. The residual-reuse error is therefore the variance of the individual action effect across disturbances, i.e., of the gap between the ITE and the CATE of \cref{sec:effects}. In particular, residual reuse is exact for a fixed action pair if and only if the action contrast is independent of the disturbance, up to $\nu_z$-null sets. Consider an additive response
\begin{equation}
    F_h(z,a,\varepsilon)
    =
    m_h(z,a)+q_h(z,\varepsilon).
    \label{eq:additive-response}
\end{equation}
The same disturbance term appears under both actions, so $\Delta_h
    =
    m_h(z,a')-m_h(z,a)$
is independent of $\varepsilon$, giving $T_R=Y_h(a')$ for every realized transition, even if $q_h$ is nonlinear, multidimensional, or noninjective.

The observed residual identifies the disturbance contribution needed for the counterfactual: $y-\mu_h(z,a)=q_h(z,\varepsilon)-\E_{\nu_z}q_h(z,\varepsilon)$. Since this centered contribution is shared across actions, we can reuse it without recovering $\varepsilon$ or learning $q_h$. Even when several disturbances produce the same observed outcome, they all give the same action contrast, so the counterfactual remains identified although the disturbance itself is not.

\begin{proposition}[Residual reuse is the only rule exact in every additive world]
\label{prop:residual-unique}
Let $T(z,a,y,a')$ be any counterfactual rule computed from the factual laws and an observed transition. If $T$ is exact in every world with an additive response
\[
    F_h(z,a,\varepsilon)
    =
    m_h(z,a)+q_h(z,\varepsilon),
\]
a common disturbance law and supported queries, then in each such world
\[
    T(z,a,y,a')
    =
    y+\mu_h(z,a')-\mu_h(z,a)
\]
for every supported action pair and every observed outcome $y$.

(Proof in \cref{proof:residual-unique}.)
\end{proposition}

Thus, additive structure identifies an individual counterfactual from factual conditional means and one observed outcome, without identifying a latent disturbance or a generative coupling. The next result quantifies the error when the action contrast depends on the disturbance.

\subsection{Counterfactual Error Is Controlled by Action--Disturbance Interactions}

The preceding subsection established that residual reuse is exact in additive worlds and is the only rule with that guarantee. We now quantify its error beyond additivity: by \eqref{eq:residual-contrast-risk}, the error is the variation of the action contrast across disturbances, i.e., the action--disturbance interaction of \cref{sec:effects}, which we bound through the mixed action--disturbance derivative of the transition map.

Fix a source $z$ and suppress the fixed duration $h$ in the notation. Let $\mathcal A\subseteq\R^p$ and $\mathcal U\subseteq\R^m$ be closed convex sets containing the supported actions and the support of $\nu_z$, with $\E_{\nu_z}\|\varepsilon\|^2<\infty$.

Suppose $F(z,a,\varepsilon)$ is continuously differentiable in $\varepsilon$, $\partial_\varepsilon F$ is continuously differentiable in $a$, and the mixed derivative satisfies
\begin{equation}
    \bigl\|
    \partial_a\partial_\varepsilon F(z,a,\varepsilon)[u,v]
    \bigr\|
    \le
    K_z\|u\|\|v\|
    \label{eq:mixed-derivative-bound}
\end{equation}
for all $a\in\mathcal A$, $\varepsilon\in\mathcal U$ and directions $u,v$.

\begin{theorem}[Interaction bound]
\label{thm:interaction-bound}
Under these conditions, every supported action pair satisfies
\begin{equation}
    \begin{aligned}
    \E_{\nu_z}\|T_R-Y(a')\|^2
    &=
    \operatorname{tr}\operatorname{Var}_{\nu_z}(\Delta)\\
    &\le
    K_z^2\|a'-a\|^2
    \operatorname{tr}\operatorname{Cov}_{\nu_z}(\varepsilon).
    \end{aligned}
    \label{eq:interaction-bound}
\end{equation}

If, in addition, the response under $a'$ separates disturbances,
\[
    \|F(z,a',\varepsilon)-F(z,a',\varepsilon')\|
    \ge
    \kappa_z\|\varepsilon-\varepsilon'\|,
    \qquad \kappa_z>0,
\]
then, whenever the disturbance variance is nonzero,
\begin{equation}
    \rho_z(a,a')
    :=
    \frac{
    \operatorname{tr}\operatorname{Var}_{\nu_z}(\Delta)
    }{
    \operatorname{tr}\operatorname{Var}_{\nu_z}(Y(a'))
    }
    \le
    \left(
    \frac{K_z\|a'-a\|}{\kappa_z}
    \right)^2.
    \label{eq:interaction-ratio}
\end{equation}

(Proof in \cref{proof:interaction-bound}.)
\end{theorem}

The separation condition gives $\operatorname{tr}\operatorname{Var}_{\nu_z}(Y(a'))\ge\kappa_z^2\operatorname{tr}\operatorname{Cov}_{\nu_z}(\varepsilon)$, yielding the relative bound. The mixed derivative measures how the response to a disturbance changes with the action. When it vanishes, the action contrast is disturbance-independent and residual reuse is exact. When it is small, the counterfactual error is controlled even if the disturbance itself is large. The normalized quantity $\rho_z(a,a')$ compares residual reuse with an ideal predictor that does not observe the factual outcome. Indeed, the minimum squared risk of predicting $Y(a')$ from $z$ and $a'$ alone is
\[
    \inf_c \E_{\nu_z}\|c-Y(a')\|^2
    =
    \operatorname{tr}\operatorname{Var}_{\nu_z}(Y(a')).
\]
The numerator of $\rho_z$ is the risk of residual reuse with exact factual predictions. Therefore,
\[
    \rho_z<1
    \quad\Longleftrightarrow\quad
    \text{residual reuse improves on the exact factual predictor}.
\]
At $\rho_z=0$ the counterfactual is exact; at $\rho_z>1$ the residual correction is worse than ignoring the observed outcome. The bound also identifies the relevant dimensionless quantity, $\frac{K_z\|a'-a\|}{\kappa_z}$.
A small mixed derivative is not sufficient on its own: the outcome must also distinguish disturbances, as quantified by $\kappa_z$. In particular, a representation that suppresses first-order disturbance variation may have a small denominator in \eqref{eq:interaction-ratio} even when the underlying physical dynamics are smooth.

This coordinate dependence motivates transporting the residual after a physical readout, as in \eqref{eq:counterfactual-jepa-chart}.

A learned predictor contributes error only through its action contrast: errors shared by the factual and alternative predictions cancel. To separate this estimation error from the physical interaction, let
\[
    \widehat\mu(z,a)=\mu(z,a)+e(z,a)
\]
and write
\[
    e_\Delta(z,a,a')=e(z,a')-e(z,a).
\]
For a fitted predictor fixed independently of the query disturbance,
\begin{align}
    \widehat T_R-Y(a')
    &=
    T_R-Y(a')+e_\Delta(z,a,a'),\\
    \E_{\nu_z}\|\widehat T_R-Y(a')\|^2
    &=
    \operatorname{tr}\operatorname{Var}_{\nu_z}(\Delta)
    +
    \|e_\Delta(z,a,a')\|^2.
    \label{eq:residual-estimation-risk}
\end{align}
With exact factual predictions, the residual-reuse error has zero mean, so its cross term with the fixed error in the predicted action contrast vanishes. Equation \eqref{eq:residual-estimation-risk} therefore separates the error due to the physical interaction from that due to learning the action contrast; for a predictor fitted on random training data, the identity holds conditional on that fit whenever the query disturbance still follows $\nu_z$.

\paragraph{What factual data reveal about the interaction.}
\label{sec:factual-interaction}
The ratio $\rho_z$ depends on the same-event contrast $\Delta$, and therefore on the coupling that factual data do not identify (\cref{sec:counterfactual}). Computing it requires paired outcomes, from a simulator or from trials that repeat a persistent disturbance. In our experiments, it is an evaluation diagnostic computed from simulator outcomes (\cref{sec:theory-checks}), not a quantity available before deployment. Factual data bound it only from below. The true joint law of the two centered outcomes $Y(a)-\E_{\nu_z}Y(a)$ and $Y(a')-\E_{\nu_z}Y(a')$ is one coupling of their laws, so $\operatorname{tr}\operatorname{Var}_{\nu_z}(\Delta)$ is at least the smallest mean squared distance between two random vectors with these laws (the squared Wasserstein distance), and, by the reverse triangle inequality, at least
\[
    \Bigl(
    \sqrt{\operatorname{tr}\operatorname{Var}_{\nu_z}(Y(a'))}
    -
    \sqrt{\operatorname{tr}\operatorname{Var}_{\nu_z}(Y(a))}
    \Bigr)^2.
\]
Both bounds involve only factual laws; the second needs only the expected squared residual of a factual world model under each action, which a second regression on factual data estimates. An outcome spread that changes with the action, as under a gain fault with commands of different magnitudes, therefore flags a large interaction from factual data alone. The converse fails: in the mechanism $Y(a)=U$, $Y(a')=1-U$ of \cref{sec:counterfactual}, both centered laws are uniform on $[-\frac12,\frac12]$, so both bounds vanish, yet $\rho_z=4$. Certifying a small interaction therefore needs another source of information: where the disturbance enters (\cref{subsec:short-time-residual}), a simulator, or paired trials.

The interaction bound controls $\operatorname{tr}\operatorname{Var}_{\nu_z}(\Delta)$. We next show why this variance is small for short physical transitions.

\subsection{Physical Dynamics Yield Minimax-Optimal Counterfactuals}

Residual reuse is exact under additivity, and the interaction bound quantifies its error beyond that regime. In physical systems, the principle behind our title, \emph{natura non facit saltus} (nature makes no leaps), takes a concrete form: separate action and disturbance forces accumulate over time and interact only through subsequent changes of state. This yields approximate additivity over a short transition, with a counterfactual error rate that we prove is minimax optimal given unpaired factual data and one observed outcome. In Newtonian and Lagrangian mechanics, this separate-force structure takes the form
\begin{equation}
    M(q)\ddot q
    =
    \tau_a(q,a)
    +
    \tau_\varepsilon(q,\dot q,\varepsilon)
    -
    c(q,\dot q).
    \label{eq:generalized-force-additivity}
\end{equation}
Here $q$ denotes configuration, $M(q)$ the mass matrix, $\tau_a$ and $\tau_\varepsilon$ the action and disturbance forces, and $c$ the remaining mechanical terms; dots denote time derivatives, so $\dot q$ is the velocity and $\ddot q$ the acceleration. At a fixed position and velocity, the two force contributions add; their interaction arises as the action changes the state at which the disturbance acts, and conversely. This separate-channel structure is assumed, not derived from smoothness: the worn actuator of \cref{sec:effects} is smooth yet interacts with the action at first order, and we list it among the violations at the end of this subsection.

To formalize this, consider the continuous-time system, an ordinary differential equation (ODE),
\begin{equation}
    \dot s
    =
    v_s(s,a)+b_s(s,\varepsilon),
    \qquad
    z=\mathcal E(s)=f(o(s)),
    \label{eq:short-time-physical-ode}
\end{equation}
with action and disturbance held fixed over a transition of duration $h$. The observation $o(s)$ is encoded by the same frozen encoder $f$ used throughout; $\mathcal E=f\circ o$ denotes their composition so that derivatives with respect to the physical state $s$ are explicit.

For a sufficiently smooth vector field, Taylor expansion gives
\begin{equation}
    s_{t+h}
    =
    s_t
    +
    h\bigl[v_s(s_t,a)+b_s(s_t,\varepsilon)\bigr]
    +
    O(h^2).
    \label{eq:physical-first-order}
\end{equation}
The first-order effect of the action does not depend on the disturbance, and the first-order effect of the disturbance does not depend on the action. Their interaction enters through higher-order terms.

Applying the smooth composition $\mathcal E=f\circ o$ gives the corresponding expansion in representation coordinates,
\[
    z_{t+h}
    =
    z_t
    +
    hD\mathcal E(s_t)v_s(s_t,a)
    +
    hD\mathcal E(s_t)b_s(s_t,\varepsilon)
    +
    O(h^2).
\]
Here $D\mathcal E(s_t)$ is the Jacobian of $\mathcal E$ at $s_t$, which maps a small change of physical state to the resulting change of embedding. The additive action and disturbance terms in the physical dynamics remain separate at first order after applying the smooth map $\mathcal E$, regardless of the encoder pretraining objective. The representation must still contain enough information to determine these first-order responses; conditioning on an insufficient $z_t$ can invalidate the expansion.

The following theorem states the property directly for the transition map. It collects all terms beyond first order into a remainder $R_h$ and assumes that its root-mean-square size over disturbances, $\|R_h\|_{L^2(\nu_z)}=(\E_{\nu_z}\|R_h\|^2)^{1/2}$, is at most $Ch^2$ for a constant $C$ that does not depend on $h$.

\begin{theorem}[Short-time residual counterfactuals]
\label{thm:short-time-residual}
Assume, uniformly on the supported source/action domain, that
\begin{equation}
    F_h(z,a,\varepsilon)
    =
    z
    +
    h\,v(z,a)
    +
    h\,w(z,\varepsilon)
    +
    R_h(z,a,\varepsilon),
    \qquad
    \|R_h(z,a,\varepsilon)\|_{L^2(\nu_z)}
    \le Ch^2,
    \label{eq:short-time-expansion}
\end{equation}
with all terms square integrable and a common disturbance law $\nu_z$ across actions.

Then, for every supported action pair,
\begin{equation}
    \|T_R-Y_h(a')\|_{L^2(\nu_z)}
    \le 2Ch^2,
    \qquad
    \E_{\nu_z}\|T_R-Y_h(a')\|^2
    \le 4C^2h^4.
    \label{eq:short-time-residual-rate}
\end{equation}

For a fixed fitted predictor $\widehat\mu_h=\mu_h+e_h$,
\begin{equation}
    \|\widehat T_R-Y_h(a')\|_{L^2(\nu_z)}
    \le
    2Ch^2
    +
    \|e_h(z,a')-e_h(z,a)\|.
    \label{eq:short-time-learned-mean}
\end{equation}

(Proof in \cref{proof:short-time-residual}.)
\end{theorem}

The cancellation follows directly from the action contrast. Subtracting the expansions under $a$ and $a'$ gives
\begin{align}
    \Delta_h
    &=
    F_h(z,a',\varepsilon)-F_h(z,a,\varepsilon)\\
    &=
    h\bigl[v(z,a')-v(z,a)\bigr]
    +
    R_h(z,a',\varepsilon)-R_h(z,a,\varepsilon).
    \label{eq:short-time-action-contrast}
\end{align}
The first-order disturbance term $h\,w(z,\varepsilon)$ cancels. Consequently,
\[
    \operatorname{tr}\operatorname{Var}_{\nu_z}(\Delta_h)
    =
    O(h^4),
\]
which gives the squared-error rate through \eqref{eq:residual-contrast-risk}.

For \eqref{eq:short-time-physical-ode}, the expansion follows from Taylor's formula when the vector fields are continuously differentiable, $\mathcal E$ is twice continuously differentiable, and the relevant derivative and moment bounds hold uniformly (\cref{app:short-time-residual}). The corresponding terms are
\[
    v(z,a)=D\mathcal E(s)v_s(s,a),
    \qquad
    w(z,\varepsilon)=D\mathcal E(s)b_s(s,\varepsilon).
\]
This covers force-like disturbances with separate action and disturbance channels, including temporally persistent disturbances when conditioned on an appropriate source context.

The absolute error rate does not by itself establish the value of conditioning on the observed outcome. As $h\to0$, all outcomes approach their source state, so even predictors that ignore both the action and disturbance can have small absolute error (\cref{prop:small-step-baselines}). We therefore consider the interaction ratio relative to the disturbance-induced variance.

\begin{corollary}[Residual reuse explains all but an $O(h^2)$ fraction]
\label{cor:relative-rate}
Under \cref{thm:short-time-residual}, suppose the first-order disturbance effect separates disturbances:
\[
    \|w(z,\varepsilon)-w(z,\varepsilon')\|
    \ge
    \kappa\|\varepsilon-\varepsilon'\|,
\]
with $\kappa>0$, and let
\[
    \sigma^2
    =
    \operatorname{tr}\operatorname{Cov}_{\nu_z}(\varepsilon)>0.
\]
For $h<\kappa\sigma/(\sqrt{2}\,C)$,
\begin{equation}
    \rho_h(a,a')
    =
    \frac{
    \E_{\nu_z}\|T_R-Y_h(a')\|^2
    }{
    \operatorname{tr}\operatorname{Var}_{\nu_z}(Y_h(a'))
    }
    \le
    \left(
    \frac{2Ch}{\kappa\sigma-\sqrt{2}\,Ch}
    \right)^2
    =
    O(h^2).
    \label{eq:relative-rate}
\end{equation}

(Proof in \cref{proof:relative-rate}.)
\end{corollary}

The numerator is $O(h^4)$ by \cref{thm:short-time-residual}, while the disturbance-induced variance in the denominator is of order $h^2$ under the separation condition. Hence the relative risk decreases as $h^2$. This is the relevant guarantee: residual reuse removes an increasing fraction of the disturbance uncertainty as the transition duration decreases. The corollary says how fast the error vanishes, not how short is short enough. Its bound holds only below the duration stated in the corollary, and guarantees an improvement over ignoring the outcome only at shorter durations still; both limits shrink when the remainder constant $C$ grows or when the disturbance moves the outcome less (smaller $\kappa\sigma$), and these constants depend on the system. Whether the transitions of a given system are short enough is therefore an empirical question. Factual data cannot settle it, since the interaction is not identified (\cref{sec:counterfactual}); \cref{sec:theory-checks} measures it on our simulators.

The first-order argument also extends to certain partially observed inertial systems. With pose-only observations, an unknown initial velocity contributes to the transition, while applied forces affect position at order $h^2$. The interaction between the force contribution and the disturbance appears at higher order. Under the conditions of \cref{cor:inertial-residual}, the mean-squared error of residual reuse is $O(h^6)$ despite the hidden velocity. If, in addition, distinct disturbances move the pose apart at order $h^2$, as in the separation condition of \cref{cor:relative-rate}, the variance they induce is at least of order $h^4$, and the relative rate is again $O(h^2)$. This is a specific property of inertial dynamics, rather than a general guarantee for arbitrary insufficient representations.

\paragraph{Minimax optimality.}
The short-time bound gives a rate achievable by residual reuse. We next ask whether another estimator could achieve a uniformly smaller risk using the same information. The estimator may know the exact factual transition laws at every action and duration and may condition on the observed factual outcome. It does not observe paired outcomes under different actions for the same disturbance.

\begin{theorem}[Residual reuse is minimax rate-optimal]
\label{thm:minimax-rate}
Let $\mathcal F_C$ contain every world satisfying the assumptions of \cref{thm:short-time-residual}, with remainder constant $C$ for all $0<h\le h_0$ and actions in $[-1,1]$.

Give an estimator the exact factual transition law at every source, action and duration, together with the observed outcome $Y_h(-1)$ of a query event at a fixed source. Then, for every $0<h\le h_0$,
\begin{equation}
    \begin{aligned}
    \frac{4C^2h^4}{1+C^2h^2}
    &\le
    \inf_\delta\sup_{M\in\mathcal F_C}
    \E_M\bigl\|\delta-Y_h^M(1)\bigr\|^2\\
    &\le
    \sup_{M\in\mathcal F_C}
    \E_M\bigl\|T_R^M-Y_h^M(1)\bigr\|^2
    =
    4C^2h^4.
    \end{aligned}
    \label{eq:minimax-rate}
\end{equation}
The lower bound already holds within smooth autonomous ODEs with separate action and disturbance channels.

(Proof in \cref{app:minimax-rate}.)
\end{theorem}

The lower and upper bounds agree at leading order as $h\to0$. Thus, residual reuse is minimax rate-optimal, with the sharp leading constant, over the stated class of short-time models. The corresponding relative rate is also optimal under the conditions of \cref{cor:minimax-relative}.

The lower-bound construction uses two worlds with identical factual transition laws but different same-disturbance couplings. The ambiguity is a rotation of the hidden disturbance whose sign cannot be recovered from the factual laws, as in related latent-rotation nonidentification results \citep{nasresfahany2023counterfactual}. The construction is realized by smooth ODEs whose second-order remainder attains the risk bound. It follows that additional model capacity or a more accurate factual likelihood cannot uniformly improve the rate without additional assumptions or information.

This is a worst-case result over $\mathcal F_C$. A known physical mechanism, paired counterfactual observations, or repeated observations of a shared disturbance may permit more accurate estimators for particular systems.

\paragraph{Limits of the short-time approximation.}
The expansion \eqref{eq:short-time-expansion} requires the first-order action response to be independent of the disturbance. Three types of violations are relevant.

First, disturbances that directly multiply the command introduce an action--disturbance interaction at first order. For $\dot s=a\varepsilon$, the counterfactual error is proportional to $h(a-a')\varepsilon$, and the relative risk can fail to decrease as $h$ shrinks (\cref{app:counterexamples}). A state-dependent disturbance, such as uncertain damping, is different: it may satisfy the separate-channel ODE model, although the relevant time scale depends on the plant dynamics.

Second, contacts, clipping and actuator saturation can introduce nonsmooth or instantaneous interactions. A transition that crosses a contact boundary can violate the uniform expansion required by the theorem.

Third, the representation can obscure the required structure. Folded coordinates may suppress the first-order disturbance response, while insufficient observations may mix states with different action contrasts. Smoothness of the encoder alone does not resolve state aliasing (\cref{app:counterexamples}). Raw pixels are an extreme case: intensities change sharply at object edges, so the second-order remainder of \cref{thm:short-time-residual} is very large in pixel coordinates, and a pixel residual does not transport at any duration we test (\cref{sec:where-transport}).

These cases do not contradict the interaction bound: they either increase the interaction, reduce disturbance separation, or violate the assumptions of the short-time expansion. We evaluate them separately in \cref{sec:stress}.

The duration $h$ in these bounds is that of the counterfactual itself, from the action to the observed outcome. Over longer durations, second-order terms can accumulate and the bounds loosen, unless the dynamics settle to an additive equilibrium (\cref{sec:theory-checks}); our experiments therefore apply the correction one transition at a time. Reusing an observed residual for later transitions is a different question, which depends on whether the disturbance persists (\cref{sec:closed-loop}).

\subsection{Reusing Pretrained Encoders and Factual Predictors}
\label{sec:interventional}
\label{subsec:std-limitations}
\label{subsec:bias}
\label{subsec:noise-cost}

The preceding guarantees use the population conditional mean $\mu_h$, whereas a counterfactual world model uses the action contrast supplied by a learned predictor. We show that $\mu_h$ can be learned by ordinary squared-loss training on a frozen pretrained encoder, regardless of its pretraining objective, and that an available factual predictor can be reused when it already supplies the required contrast. We also state when the predicted action contrast equals the conditional average effect of changing the action; by \eqref{eq:residual-estimation-risk}, only the error in this contrast contributes additional counterfactual risk.

\paragraph{Squared-loss prediction on frozen representations.}
Let $f$ be a fixed encoder and $H_t$ the model context already introduced in \cref{sec:bridge}, a fixed measurable function of observations up to $t$ and past actions, such as a sequence of frame embeddings and controls. With
\[
    X_t=(H_t,\va_t),
    \qquad
    Y_t=z_{t+1}=f(\vo_{t+1}),
\]
we retain $\mu$ for the population conditional mean and $\widehat\mu$ for a fitted predictor, suppressing the fixed transition duration. The encoder can come from any pretraining objective: once $f$ and the context construction are frozen, squared loss identifies the conditional mean of the next embedding.

\begin{theorem}[Squared loss identifies the encoded conditional mean]
\label{prop:irreducible}
For fixed measurable context $H_t$ and encoder $f$, assume $\E\|Y_t\|^2<\infty$ and let
\[
    \mu(x)=\E[Y_t\mid X_t=x].
\]
Every measurable square-integrable predictor $\widehat\mu$ satisfies
\begin{equation}
    \E\|Y_t-\widehat\mu(X_t)\|^2
    =
    \E\|Y_t-\mu(X_t)\|^2
    +
    \E\|\widehat\mu(X_t)-\mu(X_t)\|^2.
    \label{eq:mse-pythagorean}
\end{equation}
Hence $\mu$ is the unique population squared-loss minimizer up to equality almost surely on the training support, and
\begin{equation}
    \E\!\left[
    2\{\mu(X_t)-Y_t\}\mid X_t
    \right]=0.
    \label{eq:mse-unbiased-score}
\end{equation}
The minimum risk is
\begin{equation}
    \min_{\widehat\mu}\E\|Y_t-\widehat\mu(X_t)\|^2
    =
    \E\!\left[
    \operatorname{tr}\operatorname{Var}(Y_t\mid X_t)
    \right].
    \label{eq:irreducible-error}
\end{equation}

Moreover, for any measurable positive weight $w(X_t)$ satisfying
\[
    0<w(X_t)<\infty\ \text{a.s.},
    \qquad
    0<\E[w(X_t)]<\infty,
    \qquad
    \E[w(X_t)\|Y_t\|^2]<\infty,
\]
the set of population minimizers is unchanged:
\begin{equation}
    \arg\min_{\widehat\mu}
    \E\!\left[w(X_t)\|Y_t-\widehat\mu(X_t)\|^2\right]
    =
    \{\widehat\mu:\widehat\mu(X_t)=\mu(X_t)\ \text{a.s.}\}.
    \label{eq:mse-reweighting}
\end{equation}

(Proof in \cref{proof:irreducible}.)
\end{theorem}

The conditional-expectation decomposition applies to any fixed encoder: pretraining determines the features being predicted, while squared-loss training identifies their conditional mean on the factual support. We can therefore reuse released representations without requiring the original pretraining objective or predictor to match the world-modeling task.

\begin{corollary}[A released encoder can be reused]
\label{cor:frozen-encoder-mean}
For any frozen encoder checkpoint $f$, a predictor trained by squared loss on
\[
    \bigl(
    H_t,
    \va_t,
    f(\vo_{t+1})
    \bigr)
\]
has population target $\mu$. If $\mu$ belongs to the predictor class and training attains the minimum population risk, the fitted predictor equals $\mu$ almost surely on the training support.

The conclusion is independent of the encoder pretraining objective or the availability of its original predictor, and does not establish representation sufficiency or counterfactual identification.

(Proof in \cref{proof:irreducible}.)
\end{corollary}

Frozen JEPA encoders, masked autoencoders, contrastively trained representations and supervised checkpoints can thus supply the features for a factual world model. Our experiments fit a new squared-loss predictor on these frozen features; when a pretrained world model already provides the required action contrast for the chosen context and target representation, both its encoder and predictor can instead be reused without retraining. An arbitrary pretrained predictor is not guaranteed to estimate this contrast accurately: its contrast error enters \eqref{eq:residual-estimation-risk}, and counterfactual validity still requires supported actions, the identification conditions below and the preceding structural assumptions.

\paragraph{When factual predictions identify action effects.}
Squared-loss training identifies $\mu(H_t,a)=\E[Y_t\mid H_t,A_t=a]$, whereas our counterfactual construction requires the interventional contrast $\E[Y_t(a')-Y_t(a)\mid H_t]$. The difference $\mu(H_t,a')-\mu(H_t,a)$ identifies this contrast under consistency, conditional exchangeability and positivity \citep{rubin1974estimating,hernan2020causal}. Conditional exchangeability requires
\[
    Y_t(a)\ind A_t\mid H_t
\]
for each action $a$, meaning that the chosen action carries no further information about the potential outcome after conditioning on the model context. This holds when the behavior policy uses only information available to the world model and independent randomization, as formalized below; positivity additionally requires the queried actions to have support at the relevant contexts.

\begin{proposition}[Shared inputs rule out confounding]
\label{prop:shared-inputs}
Suppose the behavior policy chooses
\[
    \va_t=\pi_b(g(H_t),\xi_t),
\]
where $g$ is measurable and the policy randomization $\xi_t$ is conditionally independent of the potential outcomes $(Y_t(a))_a$ given $H_t$.

Then
\[
    Y_t(a)\ind\va_t\mid H_t
\]
for every action $a$, and, on supported actions,
\[
    \mu(H_t,a)=\E[Y_t(a)\mid H_t].
\]

(Proof in \cref{proof:shared-inputs}.)
\end{proposition}

A controller choosing actions from the observation history supplied to the world model satisfies this condition when its additional randomization is independent of the potential outcomes, even if the physical disturbance is unobserved. If a teleoperator instead adjusts the command using wind information unavailable to the model, the logged action carries additional information about the disturbance and can confound the command effect with the wind effect. Action coverage remains a separate requirement: a deterministic controller may satisfy the shared-input condition while choosing only one action at each context, leaving alternative action contrasts unsupported.

\begin{table}[t]
\centering\small
\setlength{\tabcolsep}{4pt}
\renewcommand{\arraystretch}{1.15}
\begin{tabular}{L{3.7cm}L{2.3cm}L{3.9cm}L{3.9cm}}
\toprule
\textbf{Assumption} & \textbf{Stated in} & \textbf{Holds, e.g.} & \textbf{Fails, e.g.}\\
\midrule
Logged actions depend only on what the world model sees, plus independent randomization &
\cref{prop:shared-inputs} &
an autopilot steering from the same camera, with exploration noise &
a skipper who feels the current and steers against it\\
The logged actions cover the queried ones &
\cref{sec:interventional} (positivity) &
random exploration over the full rudder range &
an autopilot that never turns hard, queried about a hard turn\\
\addlinespace
The disturbance adds its own force instead of scaling the command &
\cref{eq:generalized-force-additivity,eq:short-time-physical-ode} &
a current on a boat, a gust on a drone, an unknown load on a robot arm &
a worn motor that delivers a fraction of the command; braking on ice\\
No contact, impact or saturation within the transition &
\cref{thm:short-time-residual} &
a vehicle or an arm moving freely through water or air &
a pusher that may or may not touch the block; a ball hitting a wall\\
The transition is short compared with the interaction &
\cref{cor:relative-rate} &
control at tens of hertz; servos that settle to their setpoint &
a weak damping fault, whose window narrows with its size; long horizons\\
\addlinespace
The encoded context determines the response to the action &
\cref{thm:short-time-residual,cor:inertial-residual} &
a frame showing the boat's pose, with past frames for its velocity &
an occluded object; a single frame of a fast-moving one\\
The coordinates separate disturbances, without folds &
\cref{thm:interaction-bound} &
positions; $\sin(\mathrm{pitch})$ near level &
$\cos(\mathrm{pitch})$; raw pixels\\
\bottomrule
\end{tabular}
\caption{\textbf{Assumptions of residual reuse.} What residual reuse needs, where we state it formally, and a real system where it holds or fails. The last five are tested in \cref{sec:residual-experiments}. Factual data can reveal missing action coverage and flag some large interactions, such as gain faults (\cref{sec:factual-interaction}), but cannot confirm the first assumption or that the interaction is small.}
\label{tab:assumptions}
\end{table}

\paragraph{Learning the contrast without propensity weighting.}
Under exchangeability and overlap, ordinary squared-loss regression identifies the conditional interventional mean without inverse-propensity weighting: by \eqref{eq:mse-reweighting}, positive input weights leave the unrestricted population minimizer unchanged. This invariance concerns the target, while weighting can affect finite-sample accuracy and the approximation selected by a misspecified predictor class \citep{shimodaira2000improving}. Propensity-based and doubly robust methods remain useful for other causal targets, including marginal functionals involving nuisance estimation \citep{chernozhukov2018}, but do not by themselves identify the same-event coupling missing from factual data (\cref{sec:counterfactual}).

For counterfactual prediction, the fitted world model only needs to estimate $\mu(H_t,a')-\mu(H_t,a)$ accurately, since errors shared across actions cancel in \eqref{eq:residual-estimation-risk}; an action-dependent bias, however, enters the correction directly. Thus, a frozen pretrained encoder and an ordinary factual predictor supply the learned part of our construction whenever their supported action contrast has the required interventional interpretation. The observed outcome supplies the realized residual, and the physical dynamics control how accurately it transfers between actions in the chosen coordinates. We test these requirements below and, separately, test on three large external corpora whether predictors fitted on frozen DINOv2-S and V-JEPA~2 features leave any input-predictable residual (\cref{app:frozen-mean-checks}).

\paragraph{Assumptions at a glance.}
\Cref{tab:assumptions} collects the assumptions of residual reuse, where each is stated formally, and a real system where it holds or fails. The first two make the predicted action contrast causal, the next three keep the action--disturbance interaction small, and the last two concern the coordinates in which the residual is reused. In our benchmarks, the factual action of the evaluated transition is drawn uniformly from the command support, independently of the disturbance, so the first two hold by design; \cref{sec:residual-experiments} tests the other five.

\navsection[Experiments]{Counterfactual World Models: Empirical Validation}
\label{sec:residual-experiments}
We first test the physical approximation using simulator outcomes, then evaluate counterfactual prediction from pixels with frozen encoders and predictors trained on factual transitions. We examine how the disturbance mechanism and output coordinates affect residual reuse, and conclude with hindsight planning and the effect of disturbance persistence. No counterfactual outcome is used for training or model selection.

\subsection{Validating the Physical Approximation}
\label{sec:theory-checks}
To isolate the interaction error from prediction and readout errors, we evaluate residual reuse in physical coordinates with simulator conditional means estimated by Monte Carlo. For each source state and action pair, we compute
\[
    \rho(a,a')
    =
    \frac{
    \operatorname{tr}\operatorname{Var}
    \bigl(Y(a')-Y(a)\bigr)
    }{
    \operatorname{tr}\operatorname{Var}(Y(a'))
    }.
\]
By \eqref{eq:residual-contrast-risk}, the numerator is the counterfactual risk with exact factual predictions, while the denominator is the minimum risk without the factual outcome. Residual reuse improves on this reference when $\rho<1$.

\begin{figure}[t]
\centering
\includegraphics[width=\linewidth]{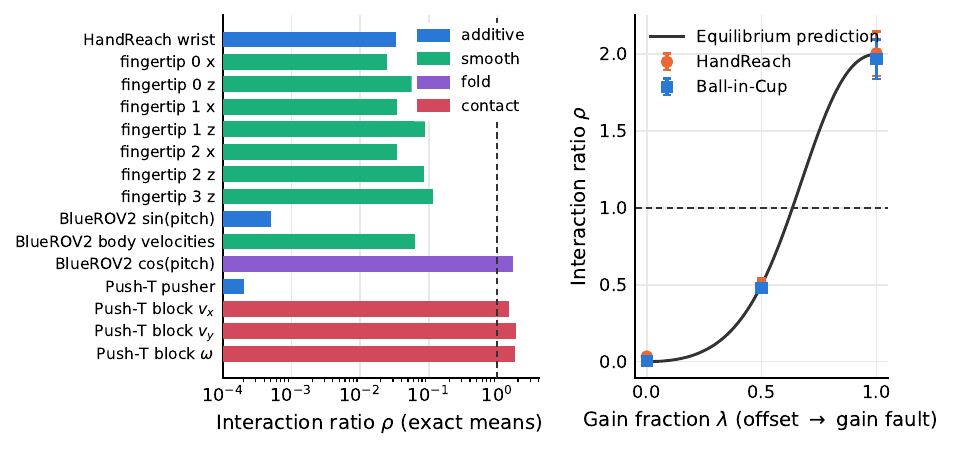}
\caption{\textbf{Counterfactual error follows the action--disturbance interaction.} Left: interaction ratios in physical coordinates; the dashed line marks the risk of the optimal predictor without the factual outcome. Right: changing an actuator offset into a gain fault increases the interaction ratio as predicted by the equilibrium model. Both panels use simulator outcomes and Monte Carlo conditional means, without image encoders or learned predictors. Intervals are bootstrapped over source states. Complete results are in \cref{tab:stress-physical,tab:dial}.}
\label{fig:theory-checks}
\end{figure}

\paragraph{Interaction determines counterfactual accuracy.}
For actuator offsets scored in the directly affected coordinates, $\rho\le0.033$ (\cref{fig:theory-checks}, left). Small interactions also occur outside additive responses: HandReach fingertip positions and BlueROV2 body velocities have $\rho\le0.12$, with ideal residual-reuse $R^2$ of $94.6$--$98.9\%$ against references near $50\%$ without the outcome. Folded pitch coordinates and contact-mediated block velocities instead have $\rho>1$, so residual reuse increases error even with exact factual predictions (\cref{tab:stress-physical}).

We also vary the disturbance continuously from an actuator offset to a gain fault,
\[
    \sigma\varepsilon
    \left[
    (1-\lambda)+\lambda\frac{a}{a_{\max}}
    \right],
    \qquad \lambda\in[0,1],
\]
while calibrating the physical parameters before evaluation to keep the reference without the outcome near $50\%$ $R^2$. The equilibrium model predicts $\rho=0$, $0.5$ and $2$ at $\lambda=0$, $0.5$ and $1$ (derivation in \cref{app:gain-fault-analysis}). HandReach and Ball-in-Cup follow this prediction: at $\lambda=0.5$, their measured ratios are $0.50$ and $0.48$; at $\lambda=1$, they are $2.01$ and $1.97$ (\cref{fig:theory-checks}, right). The error therefore follows the action--disturbance interaction as the disturbance changes from additive to multiplicative.

\begin{figure}[t]
\centering
\includegraphics[width=\linewidth]{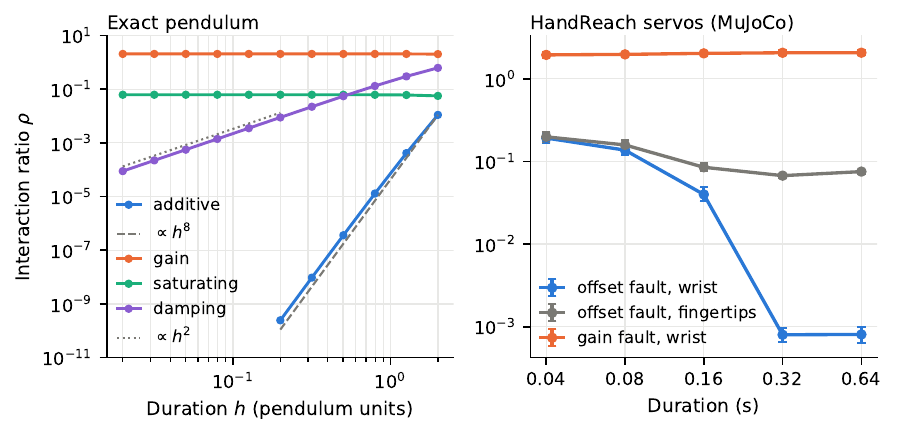}
\caption{\textbf{Approximate additivity can arise at short durations or after servo settling.} Left: a damped pendulum, $\ddot\theta=-\sin\theta-0.2c\dot\theta+u$, with additive force $u=a+\varepsilon$, gain fault $u=a(1+\varepsilon)$, saturation $u=1.2\tanh((a+\varepsilon)/1.2)$, or damping fault $c=1+1.5\varepsilon$, $u=a$; otherwise $c=1$. Evaluation uses 400 states and 256 disturbances, scoring pose with shared initial velocity. Right: native HandReach, with the prefix fixed and duration varied, using 256 source states and 64 disturbance draws. Both panels use simulator dynamics only.}
\label{fig:interaction-duration}
\end{figure}

\paragraph{Short transitions and additive equilibria.}
The short-time result predicts $\rho_h=O(h^2)$ under separate smooth action and disturbance channels. On a damped pendulum, an uncertain damping coefficient gives the predicted $h^2$ dependence, whereas an additive force gives the faster $h^8$ decrease (\cref{fig:interaction-duration}, left). Gain faults and actuator saturation instead maintain ratios near $2.06$ and $0.06$ over the tested durations. These mechanisms distinguish a small interaction from one that decreases as the transition becomes shorter.

Native position servos provide a second route to approximate additivity. In HandReach, the offset interaction ratio decreases from $0.19$ at $0.04\,\mathrm{s}$ to $0.0008$ at $0.32\,\mathrm{s}$ as the actuators settle to an approximately additive setpoint; fingertip coordinates settle near $\rho=0.07$, while gain faults remain near $2$ (\cref{fig:interaction-duration}, right). This improvement at longer durations comes from the equilibrium response. It is distinct from the short-time guarantee and shows why the dynamics, duration and measured coordinates must be considered together.

\subsection{Counterfactual Prediction from Pixels}
\label{sec:setup}
\begin{figure}[t]
\centering
\includegraphics[width=\linewidth]{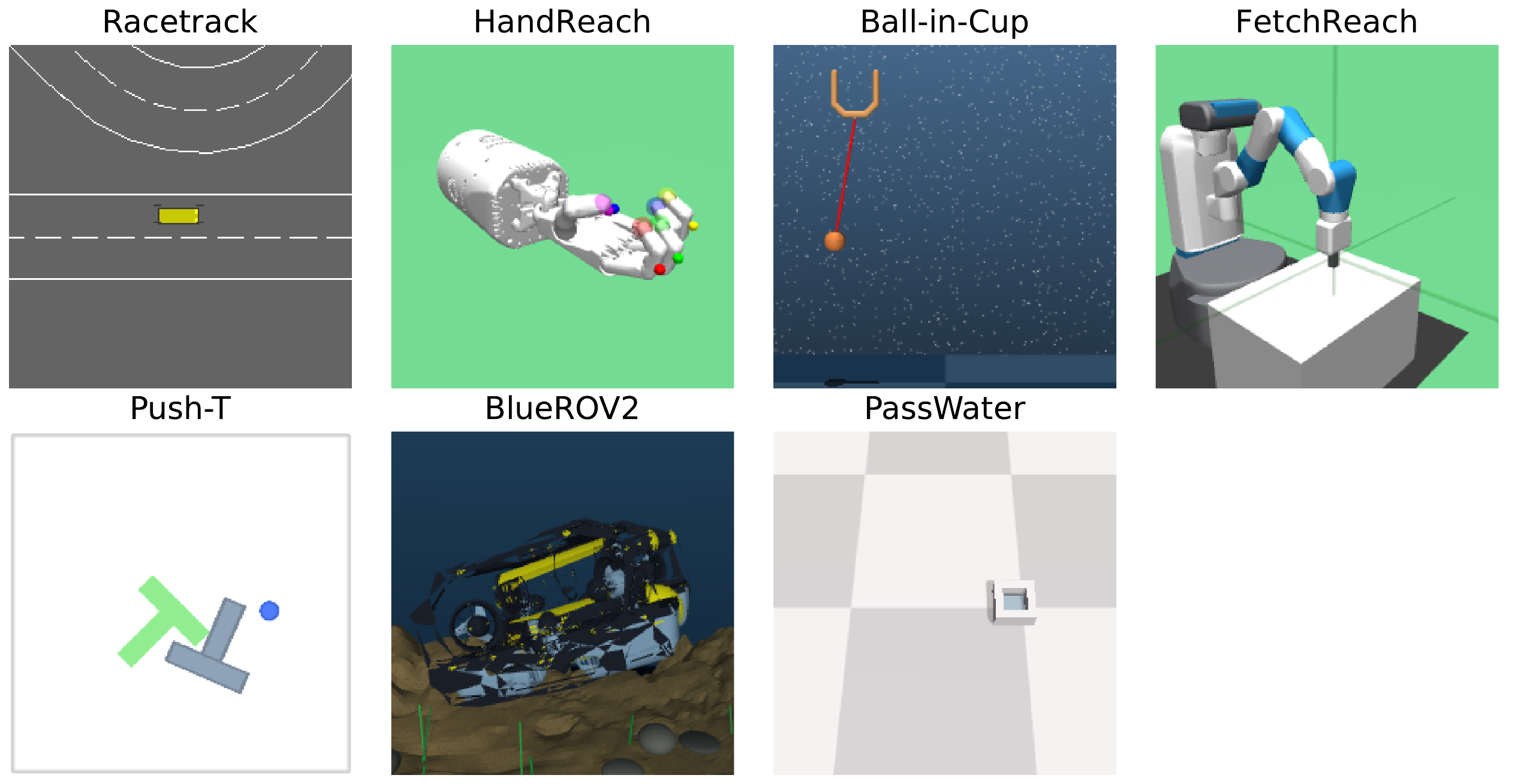}
\caption{\textbf{Seven rendered physical environments with hidden disturbances.} Racetrack, HandReach, Ball-in-Cup, FetchReach, Push-T, BlueROV2 and liquid transport (PassWater), shown after two behavior-policy transitions. Counterfactual branches share the source state and realized disturbance. Environment design, scored coordinates and observability checks are detailed in \cref{app:independent-benchmark-design,app:pinned-protocol}.}
\label{fig:environment-overview}
\end{figure}

\paragraph{Environments and factual training.}
We use seven native simulators rendered at $224\times224$: HandReach, Ball-in-Cup, FetchReach, liquid transport (PassWater), Racetrack, Push-T and BlueROV2 (\cref{fig:environment-overview}). Each transition receives a bounded disturbance with a physical meaning, such as an offset in a servo target, force, position goal or thrust. After two transitions of a task-directed behavior policy, we sample a factual action and three distinct alternative actions uniformly from the declared command support. Evaluation branches share the same source state and realized disturbance. We calibrate the environments so that both actions and disturbances affect the outcome and verify image-based recovery of the scored quantities; the complete design and observability protocol is in \cref{app:independent-benchmark-design,app:pinned-protocol}. The primary comparisons use HandReach, Ball-in-Cup, FetchReach, PassWater and BlueROV2; Racetrack and Push-T are retained as diagnostics.

Images are encoded by frozen, publicly released self-supervised vision transformers, without fine-tuning: DINOv2 \citep{oquab2023dinov2}, trained by self-distillation, in its small and base sizes (DINOv2-S, DINOv2-B), and the masked autoencoder MAE \citep{he2022masked}, trained to reconstruct masked image patches, in its base size (MAE-B). These two cover the two main families of self-supervised pretraining, joint embedding and reconstruction, and DINOv2 is the encoder of recent JEPA world models \citep{zhou2024dinowm}. Each frame is represented by the encoder's class-token (CLS) embedding. Four further frozen encoders, DINOv3-B \citep{simeoni2025dinov3}, \mbox{SigLIP~2-B} \citep{tschannen2025siglip2}, CLIP-B \citep{radford2021learning} and EVA-02-B \citep{fang2023eva02}, go through the same pipeline, each with the default pooling of its released checkpoint (\cref{tab:encoder-breadth,app:encoders}). A predictor with two attention blocks of width $1{,}024$ receives $k\in\{1,2,3\}$ frame embeddings, past actions and the query action. It is fitted on all recorded factual transitions using
\begin{equation}
 \mathcal L_{\mathrm{factual}}
 =\E_{\mathrm{train}}\Bigl\|
 d_\theta(H_t^{(k)},a_t)-\frac{z_{t+1}-z_t}{r}\Bigr\|^2,
 \qquad \widehat\mu_k(H,a)=z_t+r\,d_\theta(H,a),
 \label{eq:factual-loss}
\end{equation}
where $r$ scales feature displacements using fitting data only. Architecture, regularization and checkpoints are selected by held-out factual error (\cref{app:stress}). The counterfactual prediction is
\begin{equation}
 \widehat z_{t+1}^{\mathrm{CF}}
 =z_{t+1}+\widehat\mu_k(H_t^{(k)},a')-\widehat\mu_k(H_t^{(k)},a_t).
 \label{eq:history-residual-rule}
\end{equation}
Residual correction adds no parameters or training stage. Each world model is thus a JEPA: it predicts the next embedding, never pixels, including with MAE-B, whose pretraining reconstructs pixels. Encoder specifications are in \cref{app:encoders}.

\paragraph{Evaluation and baselines.}
Frozen nonlinear readouts $\pi$ are fitted on factual frames with physical-state targets. They provide physical units for evaluation and coordinates in which we can apply residual correction directly, as in \eqref{eq:counterfactual-jepa-chart}; we compare this construction with decoding the corrected embedding. No counterfactual label enters fitting or model selection. We report physical counterfactual $R^2$ on prespecified coordinates: one minus the squared error of the predicted coordinates, summed over coordinates and queries, divided by the summed variance of their true counterfactual values. Intervals are bootstrapped over source states and training seeds.

The \emph{trained predictor}, $\pi(\widehat\mu_k(H,a'))$, shares all weights with the counterfactual model but does not observe the factual outcome. The \emph{oracle predictor (no outcome)} knows the full simulator source state and averages the true dynamics over the disturbance law, giving the optimal squared-error prediction with that information. \emph{Factual copying}, $\pi(z_{t+1})$, tests whether the alternative action matters, while shuffling factual outcomes across disturbance draws tests whether the correction uses information specific to the observed event.

\begin{figure}[t]\centering
\includegraphics[width=\linewidth]{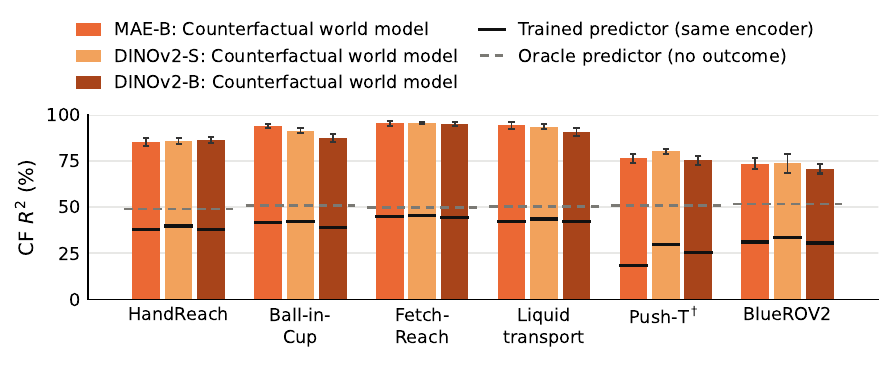}
\caption{\textbf{Residual correction improves counterfactual prediction with the same factual predictor.} CF $R^2$: physical counterfactual $R^2$. Bars: correction after the readout, with bootstrap intervals over source states and training seeds. Black ticks: the same predictor without the factual outcome. Dashed lines: the oracle predictor with the full simulator source state but no outcome. All predictors use one frame, $8$k factual episodes and three seeds. $^\dagger$Push-T is diagnostic. Complete values, feature-space correction, ensembles and larger BlueROV2 collections are in \cref{tab:main-cf}.}
\label{fig:main-cf}
\end{figure}

\paragraph{Counterfactual accuracy.}
\label{sec:pinned-results}
With one frame and $8$k factual episodes per environment, correction after the readout reaches $85.6$--$95.7\%$ $R^2$ on HandReach, Ball-in-Cup, FetchReach and PassWater, compared with $38.1$--$45.6\%$ for the same trained predictors and $49.1$--$51.1\%$ for the oracle predictors without the outcome (\cref{fig:main-cf}). Factual copying reaches $6.7$--$43.5\%$, and shuffling outcomes reduces counterfactual accuracy to between $-14.6$ and $6.3\%$ (\cref{tab:main-cf}). Thus, the improvement uses the realized disturbance information in the factual outcome; it cannot be explained by copying that outcome or improving the underlying predictor. The four further encoders give the same picture: on the same four environments, wherever they pass our disturbance-recovery criterion, correction after the readout reaches $81.0$--$95.3\%$, against $34.7$--$45.4\%$ for their trained predictors, and DINOv3-B gives the highest values of all seven encoders on HandReach and Ball-in-Cup, $88.8$ and $95.0\%$ (\cref{tab:encoder-breadth}).

The interaction ratios of \cref{fig:theory-checks,tab:stress-physical} are computed from simulator outcomes in physical coordinates, so they are the same for every encoder. An encoder changes the counterfactual error only through the error of the learned action contrast, the second term of \eqref{eq:residual-estimation-risk}, and through the readout: across the seven encoders, correction after the readout ranges over $81.0$--$88.8\%$ on HandReach, against $98.3\%$ with exact simulator means.

BlueROV2 requires more factual data. With the same $8$k episodes, correction after the readout reaches $70.7$--$74.2\%$ with MAE-B and DINOv2, and $75.5\%$ with DINOv3-B, against $51.8\%$ for the oracle predictor without the outcome. Correction in feature space reaches only $46.1$--$53.0\%$. The complete comparison, including seed ensembles, larger data budgets and the diagnostic Push-T results, is in \cref{tab:main-cf}.

\paragraph{Correction without state labels.}
The two constructions of \cref{def:counterfactual-jepa} need different supervision. Correction in feature space, \eqref{eq:counterfactual-jepa-definition}, uses only the frozen encoder and the factual predictor, so it applies to any pretrained representation. Correction after the readout, \eqref{eq:counterfactual-jepa-chart}, also needs the readout, fitted on physical-state labels of factual frames. We score both through the same readout, which any evaluation in physical units requires. Without labels, correction in feature space reaches $73.4$--$94.1\%$ on HandReach, Ball-in-Cup, FetchReach and PassWater, against $38.1$--$45.6\%$ for the trained predictors, so most of the gain needs no state supervision. The readout adds $1.4$--$4.3$ points on the first three and $12.1$--$17.7$ points on PassWater. On BlueROV2, the readout is needed: correction in feature space stays near the oracle predictor without the outcome at the standard data size, and reaches $65.9$--$72.2\%$ with sixteen times the data, against $90.1$--$93.5\%$ after the readout (\cref{tab:main-cf}). \Cref{sec:where-transport} traces this gap to the coordinates: the full feature vector is not additive, and the readout selects coordinates in which the action and the disturbance nearly add.

\paragraph{Factual predictor quality.}
\label{sec:stronger-predictor}
Equation \eqref{eq:residual-estimation-risk} separates the interaction error from the error in the predicted action contrast. Selecting the predictor configuration by factual validation loss reduces factual error by $8$--$17\%$ in every tested cell and improves or preserves counterfactual accuracy in five of six environments (\cref{tab:wm-quality}). Push-T provides an exception: factual feature error improves while the predicted pusher action contrast becomes less accurate. Factual loss over the full feature vector therefore provides an imperfect measure of the action contrast required by residual reuse.

\begin{figure}[t]\centering
\includegraphics[width=\linewidth]{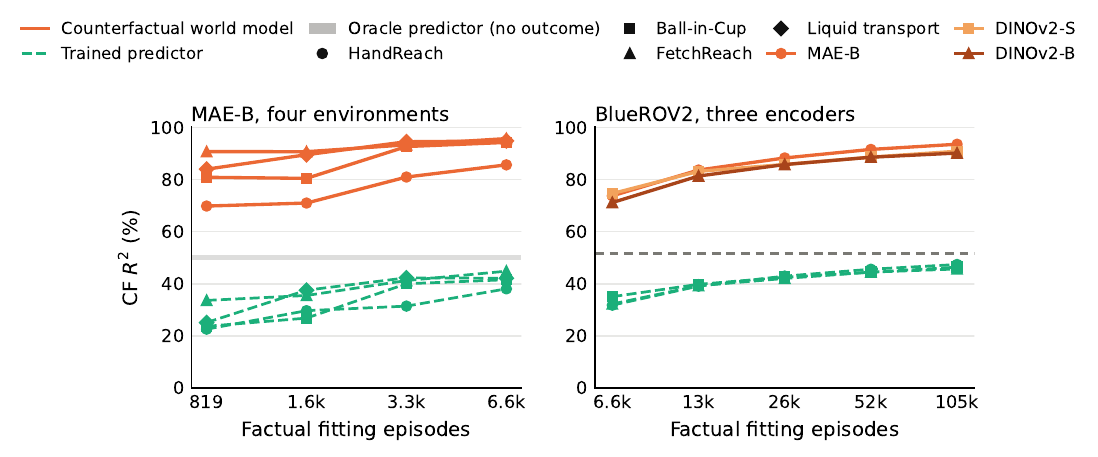}
\caption{\textbf{More factual data improves counterfactual accuracy without changing the correction.} Solid: counterfactual prediction after the readout. Dashed: the same trained predictor without the outcome. Left: MAE-B on four environments; the gray band spans their oracle references without the outcome. Right: BlueROV2 with three encoders and nested factual collections; the gray dashed line is its oracle reference. Fitting episodes are $80\%$ of each factual collection, the rest being held out, so the standard $8$k-episode collection gives $6{,}553$ (\cref{app:factual-fitting}). One frame, three seeds; complete results in \cref{tab:data-size,tab:main-cf}.}
\label{fig:scaling}
\end{figure}

Increasing the factual data improves counterfactual prediction without changing the correction. With MAE-B, an eightfold increase raises readout-corrected $R^2$ from $69.8$ to $85.6\%$ on HandReach, $80.8$ to $94.2\%$ on Ball-in-Cup, $90.7$ to $95.7\%$ on FetchReach and $83.9$ to $94.8\%$ on PassWater (\cref{fig:scaling}). BlueROV2 reaches $85.7$--$88.2\%$ across encoders with four times the standard data and $90.1$--$93.5\%$ with sixteen times the data; its oracle predictor without the outcome remains at $51.8\%$. These results identify factual prediction error as a practical limitation even when the physical interaction is small.

\begin{figure}[t]\centering
\includegraphics[width=.98\linewidth]{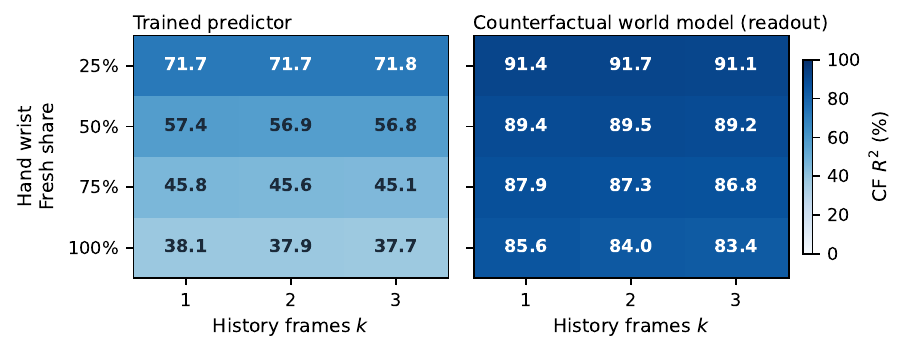}
\caption{\textbf{The observed outcome retains information as the disturbance becomes less predictable from history.} HandReach with frozen MAE features, $k=1$--$3$ history frames and three training seeds. Scores are physical counterfactual $R^2$ after the readout. Physical parameters and action support are fixed; changing the fresh disturbance share also changes the visited-state distribution. Bootstrap intervals and copying results are in \cref{tab:pinned-hand-iid}.}
\label{fig:pinned-hand-iid-history}
\end{figure}

\paragraph{Observation history.}
\label{sec:history-main}
In this ablation, each disturbance has two components (\cref{eq:pinned-driver}): a \emph{fresh} one, independent and identically distributed (IID) over transitions, and a \emph{persistent} one, correlated across transitions. Past frames can reveal the persistent component, whereas only the current outcome carries information about the fresh one. The main benchmark is fully fresh. On HandReach, increasing the fresh share of disturbance variance from $25$ to $100\%$ reduces the trained predictor's $R^2$ from $71.7$ to $38.1\%$, whereas the counterfactual model goes only from $91.4$ to $85.6\%$ (\cref{fig:pinned-hand-iid-history}). With half the variance persistent, increasing the context from one to three frames changes trained-predictor accuracy by $-2.3$ to $+2.1$ points and counterfactual accuracy by $-4.5$ to $+1.6$ points. The population benefit of additional history is therefore not consistently realized by these finite fitted models. The formal risk comparison and complete history results are given in \cref{prop:history-risk-order,sec:pinned-ablations}.

\subsection{Conditions and Limits of Residual Reuse}
\label{sec:where-transport}
The interaction analysis predicts how residual reuse changes with the measured coordinates and the disturbance mechanism. We test both using the same factual-training procedure, with a separate factual predictor fitted for each disturbance condition.

\paragraph{Nonlinear responses and external forces.}
The HandReach predictor also produces accurate fingertip counterfactuals through a second readout fitted on factual frames. On the seven prespecified fingertip coordinates, MAE-B reaches $83.6\%$ $R^2$ with correction in feature space, against $39.8\%$ for the trained predictor, $-1.3\%$ for factual copying and $53.0\%$ for the oracle predictor without the outcome. The simulator reference with exact means reaches $96.5\%$ (\cref{app:stress}). These nonlinear coordinates therefore retain much of the benefit found on the directly actuated wrist.

\begin{figure}[t]\centering
\includegraphics[width=\linewidth]{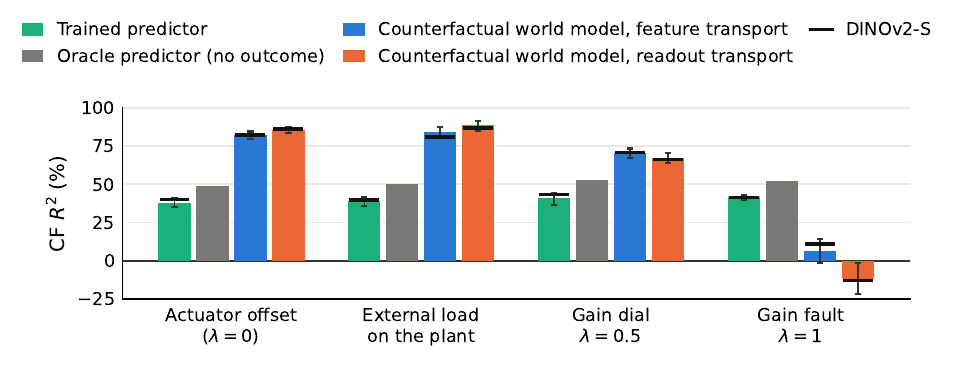}
\caption{\textbf{The learned correction succeeds under an external load and fails under a gain fault.} HandReach wrist $R^2$ for the trained predictor, oracle predictor without the outcome, and counterfactual model with correction in feature or readout coordinates. MAE-B bars include bootstrap intervals; DINOv2-S results are black ticks. Each disturbance condition uses its own factual data, frozen readouts and three predictors selected by factual loss. Complete results are in \cref{tab:plant-learned,tab:dial-learned}.}
\label{fig:stress-learned}
\end{figure}

Residual reuse also works when the disturbance acts on the plant rather than through the command. An external load applied at the HandReach palm reaches the wrist through the configuration-dependent force Jacobian. Its measured interaction is small, $\rho\le0.001$, and the learned counterfactual reaches $88.7\%$ with MAE-B and $86.9\%$ with DINOv2-S after the readout, against $39$--$40\%$ for the trained predictors and $50\%$ for the oracle predictor without the outcome (\cref{fig:stress-learned}). The disturbance's point of entry therefore does not determine whether correction succeeds; its interaction with the action does.

\paragraph{Output coordinates.}
The same physics can give different interaction errors in different coordinates. For BlueROV2, $\sin(\mathrm{pitch})$ is additive with $\rho=0$, whereas the folded coordinate $\cos(\mathrm{pitch})$ gives $\rho=1.67$. For the actuator-offset conditions, applying correction after the physical readout improves on decoding the corrected embedding in every environment, by $1.4$ to $25.9$ points (\cref{tab:main-cf}, including diagnostic Push-T). The readout thus supplies both physical quantities and coordinates with smaller action--disturbance interactions; accurate physical counterfactuals do not establish additivity of the full feature vector (\cref{app:readout-error} gives an explicit example).

\begin{figure}[!tb]\centering
\includegraphics[width=\linewidth]{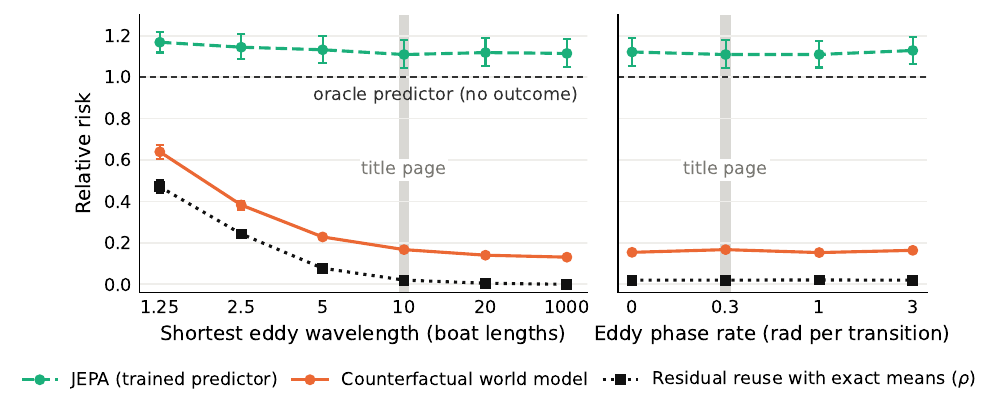}
\caption{\textbf{Residual reuse degrades as the current varies in space, not in time.} Relative risk, i.e., summed squared position error over that of the oracle predictor without the outcome (dashed line), on the title-page boat as its eddies shrink (left, phase rate $0.3$ radians per transition) and as their pattern drifts faster (right, $L=10$ boat lengths). Residual reuse with exact conditional means estimates the interaction ratio $\rho$; the counterfactual world model adds a nearly constant estimation error to it, while the trained predictor (JEPA) stays above the oracle predictor. Gray marks the title-page current. Bars are bootstrap $95\%$ intervals over $1000$ fresh events with $16$ alternative headings each. Diagnostic: the oracle predictor exceeds $60\%$ position $R^2$ (\cref{tab:boat-field}).}
\label{fig:boat-field}
\end{figure}

\paragraph{Currents that vary in space and time.}
A uniform current makes residual reuse exact on the title-page boat, so one may ask how constant the disturbance must be. We vary the eddies of its current along two axes, with their speed and the mean flow fixed: their size, through the shortest wavelength $L$ from $1000$ boat lengths down to $1.25$, and how fast their pattern drifts in time, from $0$ to $3$ radians per transition. Each condition has its own $128$k factual episodes, frozen MAE-B readout and factual predictor (\cref{app:boat-field}). In space, the interaction ratio grows as $1/L^2$, from $2\times10^{-6}$ at $L=1000$ to $0.08$ at $L=5$: the gradient of the current scales as $1/L$, and it sets how differently the current pushes the boat at the endpoints of two headings, one boat length apart. The growth slows once the waves approach the length of the move, and $\rho$ reaches $0.49$ at $L=1.25$. The counterfactual world model follows $\rho$ with a nearly constant estimation error: its relative risk lies $0.12$ to $0.15$ above $\rho$ at every size, from $0.13$ to $0.64$, while the trained predictor stays at $1.11$--$1.17$, slightly above the oracle predictor without the outcome (\cref{fig:boat-field}). In time, $\rho$ stays at $0.020$--$0.022$ and the counterfactual relative risk at $0.15$--$0.17$ for every phase rate: all headings of an event meet the field at the same instants, so the correction depends on how the current differs between the places the actions lead to, not on how it changes over time. A change over time would matter only if a residual observed in one transition were reused in a later one. Residual reuse therefore needs a disturbance that varies little over the distance separating the outcomes of alternative actions. These collections are diagnostic, since the oracle predictor already reaches $86$--$97\%$ position $R^2$ and recovery of the north-south drift falls just below $95\%$ (\cref{tab:boat-field}).

\paragraph{Pixels do not transport.}
Residual reuse needs coordinates in which actions and disturbances add, and raw pixels lack them even with exact means. On the title-page boat (\cref{fig:teaser}) with a uniform current, for which correction after the readout is exact, we transport the observed outcome with Monte Carlo conditional means over the current, so no model is fitted and only the coordinates change. In pixels, the interaction ratio is $2.13$: the current displaces the boat's edges, which lie elsewhere under the alternative heading, so adding the difference of mean images leaves a dark imprint and a blurred copy rather than a displaced boat. Read back through the frozen encoder and readout, these images give a relative risk of $5.86$ against the true position, far worse than ignoring the outcome. The $768$ MAE-B features are not additive in their own metric either ($\rho=1.92$), yet correction in feature space still carries the displacement into the readout (relative risk $0.31$), and correction after the readout reaches $0.002$. Shorter transitions do not rescue pixels: their ratio stays between $1.74$ and $2.18$ from $h=1$ to $h=0.05$ (\cref{app:pixel-transport}). Pixel intensities change sharply at the boat's edges, so the second-order term of \cref{thm:short-time-residual} becomes small only for moves well below a pixel, and even at $h=0.05$ the commanded move spans about $1.4$ pixels. Counterfactual world models therefore reuse residuals in compact coordinates read out from a learned representation; a world model that predicts pixels would need such coordinates before its residual could be reused.

\paragraph{Contacts and multiplicative faults.}
\label{sec:stress}
Push-T block velocities depend on whether the disturbed command produces contact. Their interaction ratios exceed one, and residual reuse with exact simulator means gives $R^2$ values of $5.6$, $-15.5$ and $-44.4\%$, below the references without the outcome at $20$--$39\%$ (\cref{tab:stress-physical}). The failure is present before fitting a visual predictor.

The learned gain-fault experiment follows the interaction increase found in \cref{sec:theory-checks}. At $\lambda=0.5$, correction in feature space reaches $70.5$--$70.6\%$, above the oracle predictor without the outcome at $52.9\%$. At $\lambda=1$, it falls to $6.4$--$11.0\%$, below the trained predictors at $41$--$42\%$; correction after the readout falls further, to between $-12.8$ and $-11.8\%$ (\cref{fig:stress-learned,tab:dial-learned}). These $\lambda>0$ cells are diagnostic, since disturbance recovery from the frozen features falls below $95\%$. A more accurate factual predictor cannot remove the interaction responsible for the exact-mean residual error. Complete physical checks, observability results and the uncertain-damping experiments are in \cref{app:stress,tab:plant-screen}.

\subsection{Planning with the Observed Outcome}
\label{sec:closed-loop}
Hindsight planning asks which command would have reached a goal in the event that just occurred, e.g., which heading would have brought the boat of \cref{fig:teaser} to a mooring under the current it met. A retry, the correction of later commands under a persistent fault and the diagnosis of a failed attempt all need this answer; we test the first two below. We evaluate whether improved counterfactual predictions change the action selected for an observed event. After one factual transition, the planner receives a goal reached by another action under the same source state and disturbance. It selects a command, which we score by replaying that event. The goal is attainable by construction. The trained predictor and counterfactual model share the same frozen models, those of \cref{fig:main-cf} except on BlueROV2, which uses its $32$k-episode predictors; the oracle planners use the true simulator, either averaging over the disturbance law or observing the realized disturbance. All planners use the same cross-entropy search, except that on liquid transport the true-simulator planners use $16$ candidates per round (\cref{app:closed-loop}). Success requires standardized distance to the goal below $0.2$; full protocols are in \cref{app:closed-loop}.

\begin{figure}[t]\centering
\includegraphics[width=\linewidth]{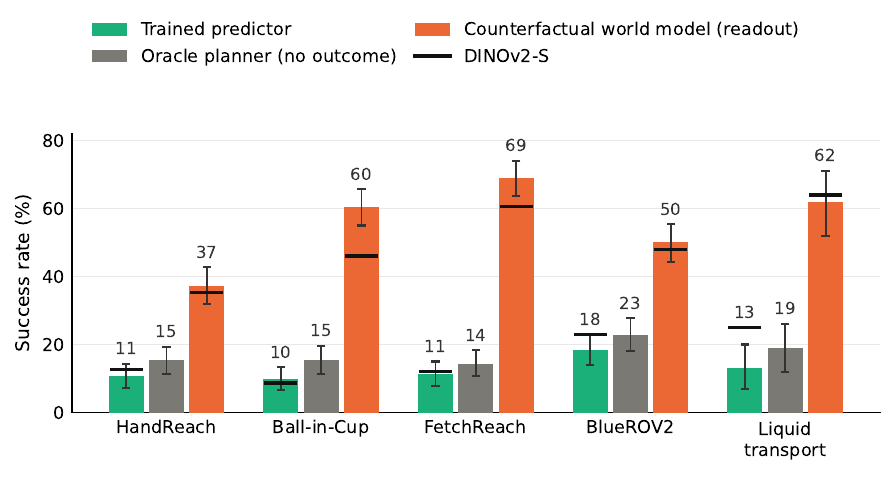}
\caption{\textbf{The observed outcome improves action selection for the same event.} Hindsight-planning success when the chosen command is evaluated under the observed event's source state and disturbance. The trained predictor and counterfactual model share their frozen weights, with the $32$k-episode predictors on BlueROV2; the oracle without the outcome knows the simulator and full source state. MAE-B bars include $95\%$ episode-bootstrap intervals; DINOv2-S results are black ticks. Complete results are in \cref{tab:oneshot}.}
\label{fig:oneshot}
\end{figure}

The counterfactual world model succeeds in $35$--$69\%$ of events, compared with $9$--$25\%$ for the trained predictor and $14$--$23\%$ for the oracle planner without the outcome (\cref{fig:oneshot}). Its gain over the latter is positive in all ten environment--encoder comparisons, with bootstrap intervals above zero. The oracle planner observing the true disturbance succeeds in every event (\cref{tab:oneshot}). Thus, the observed outcome supplies information useful for action selection even when the comparison planner knows the true dynamics. The oracle without the outcome minimizes expected squared distance; it is a reference for that objective, not a universal upper bound on thresholded success.

\paragraph{Repeated attempts and subsequent transitions.}
Two additional attempts on the same event raise counterfactual-model success to $51$--$97\%$, while the trained predictor remains at $9$--$24\%$. Over three consecutive transitions, a persistent fault, one disturbance held for the whole episode, gives an improvement from $4$--$25\%$ to $11$--$51\%$, whereas fresh disturbances, redrawn independently at every transition, give changes of only $-4$ to $6$ points. Without disturbances, residual correction reduces replanning success by $4$--$15$ points, although it still improves same-event retries by $13$--$58$ points (\cref{tab:retry,tab:closed-loop}). The command-specific prediction-error analysis in \cref{fig:closed-loop-why} explains this dependence: the observed residual predicts the error for the same event and for later transitions under a persistent fault, but not for later independent disturbances. The planning benefit therefore depends on whether the disturbance revealed by the outcome also affects the action being evaluated.

\navsection[Conclusion]{Conclusion}
We proposed counterfactual world models, which give world models a capability that
factual training cannot provide on its own, at the cost of two predictor calls
and one addition. The argument has one thread: individual counterfactuals
cannot be learned from unpaired data, physics supplies the missing structure
when disturbances enter through their own channel,
that structure dictates residual reuse, no estimator using unpaired data can
uniformly improve its rate, and the experiments confirm both its successes and
its failures. The rule is valid where the action--disturbance interaction is
small in the chosen representation, which
physics provides for force-like disturbances over short transitions and for
servoed setpoints once they settle: \emph{natura non facit saltus}. Where nature
leaps, as through contacts, or where the disturbance scales the command instead
of adding its own force, the recipe fails, measurably.

\textbf{The residual carries the event, the world model carries the action.}~On
the boat of \cref{fig:teaser}, the residual between the observed and predicted
positions is the drift caused by the current in this event; in HandReach, it is
the wrist offset produced by the miscalibrated servo; in BlueROV2, the pitch
change produced by the thruster offset. The world model supplies only how far the
change of command moves the system. Our benchmarks are calibrated so that the
disturbance carries about half of the counterfactual variance, which no
predictor that ignores the outcome can explain (oracle near $50\%$); on
HandReach, the same predictor explains $38.1\%$ without the residual and $85.6\%$
with it (\cref{tab:main-cf}). The remaining error has two sources, as in
\eqref{eq:residual-estimation-risk}: the interaction, which no predictor can
remove, and the error of the learned action contrast, to which the readout adds
its own. With exact simulator means, residual reuse explains $98.3$--$100\%$ of
the counterfactual variance on the primary coordinates of HandReach,
Ball-in-Cup, BlueROV2 and the Push-T pusher
(\cref{tab:stress-physical,tab:dial}). On these coordinates, most of the gap
between the learned and the ideal counterfactual therefore comes from learning,
consistent with the gains from more factual data (\cref{fig:scaling}).

\textbf{The operating window is set by the interaction, not by the clock.}~The
interaction ratio is the quantity to check. With exact factual predictions,
residual reuse beats the best predictor that ignores the outcome exactly when
this ratio is below one, which \cref{thm:interaction-bound} guarantees when the
interaction, scaled by the size of the action change, is small compared with how
strongly the disturbance moves the outcome. In our simulators, the ratio stays
well below one over one transition for every offset and external load of
\cref{tab:stress-physical,tab:plant-screen}, in coordinates without folds or
contacts, and it even decreases at longer durations once servos settle
(\cref{fig:interaction-duration}). Gain faults keep it near two at every tested
duration, and contacts push it above one. Damping faults show where the window
closes for a disturbance that does enter through its own channel: they move the
outcome so little over these durations that their ratio stays at $0.74$--$1.08$
(\cref{tab:plant-screen}); for such weak disturbances, \cref{cor:relative-rate}
guarantees a ratio below one only over durations proportional to $\kappa\sigma$,
a window that narrows with the disturbance. The ratio itself needs paired
outcomes, so it is an evaluation diagnostic, not a quantity one can compute
from factual logs before deployment. Factual data can flag a large interaction,
such as a gain fault, which spreads the outcome more under larger commands, but
cannot certify a small one (\cref{sec:factual-interaction}); certifying a small
interaction on a new system needs another source: knowledge of where the disturbance enters, a
simulator, or paired trials under a disturbance that persists long enough to try
two actions.

We now draw one consequence for each of the three fields this work connects:
world models, causal inference and control.

\textbf{For world models, a factual world model already contains a counterfactual
one.}~It takes two queries of the same predictor and the observed outcome. Its
accuracy is bounded by physics and approached by the factual world model: with
sixteen times more factual data, the same recipe moves the BlueROV2 underwater
vehicle from $73.6$ to $93.5\%$ ($94.1\%$ with seed ensembles), while the
trained predictor stays below the oracle predictor (no outcome). Accurate
counterfactuals need no paired data, latent disturbance model or extra
training, but they do need the observed outcome, coordinates in which the
response is near-additive, and features that determine the action response.
Generative likelihood, however good, does not supply the missing coupling
(\cref{thm:factual-gauge-nonidentification}).

\textbf{For causal inference, mechanics turns an identification assumption into
a physical one.}~For physical systems, the additive noise assumption that
identifies counterfactuals holds approximately when the action and the
disturbance enter through separate channels, with an explicit approximation rate
and explicit failure modes. The assumption concerns physical coordinates, such
as those read out from learned image features, and need not hold in the features
themselves: on the boat, MAE-B features give $\rho=1.92$
(\cref{sec:where-transport}). The interaction ratio is a scale-free diagnostic of how
far a system is from the additive noise model in given coordinates
(\cref{thm:interaction-bound}), measurable in simulation or from paired trials,
but not from factual logs alone.

\textbf{For control, classical disturbance estimators extend to learned image
world models.}~Disturbance observers, offset-free model predictive control and iterative learning
control are counterfactual estimators, usually analyzed with a known nominal
model, often a linear or linearized one
\citep{chen2016disturbance,pannocchia2003disturbance,bristow2006survey}. Our
analysis extends their reasoning to nonlinear models learned on frozen image
features, at the price of an interaction term that is second order for force-like disturbances
and first order for multiplicative faults. In planning, the same construction
lets a frozen image world model (i) choose in hindsight better than the true
simulator without the outcome, (ii) correct a retried event, as iterative
learning control does, and (iii) reject a persistent fault, as offset-free
control does; when nothing persists, the transported residual is independent
of the next error and costs success (\cref{sec:closed-loop}).

\textbf{Limitations.}~Residual reuse inherits four limits from its derivation.
First, it needs coordinates in which the effects of the action and the
disturbance add. A physical readout can provide them where raw pixels do not,
and even a readout fails where it folds, as $\cos(\mathrm{pitch})$ does
(\cref{sec:where-transport}); a world model without a compact readout must first
find such coordinates. Fitting the readout needs physical-state labels on
factual frames; without them, correction in feature space keeps most of the
gain on four environments and about half of it on BlueROV2
(\cref{sec:pinned-results}). Second, it needs the disturbance to enter through its own
channel, the transition to avoid contacts, and the duration to keep the
interaction small; the failures of \cref{sec:stress} require the additional
information listed in \cref{tab:map}. Third, it reuses one observed residual over
one transition. It does not chain counterfactuals over long horizons, and the
residual helps later decisions only when the disturbance persists
(\cref{sec:closed-loop}). Fourth, the predicted contrast must be causal: we
assume that the logged actions depended on nothing the world model does not see
(\cref{prop:shared-inputs}), and leave logs from operators who perceived more
than was recorded for future work.

\textbf{Scope of the evidence.}~All environments are simulated, and their
disturbances are controlled actuator faults with stated physical meaning, not
exhaustive models of real-world uncertainty. Our scores concern prespecified
physical coordinates decoded from frozen frame embeddings, not full simulator
state or Markov sufficiency. The planning tasks use one-transition search, at
most three steps and goals in those coordinates, and the one-shot and retry
tasks score a hindsight decision by replaying the event in simulation. Two of
seven environments fail our criteria for recovering the scored quantities or the
disturbance from images (\cref{app:pinned-protocol}); BlueROV2 passes only with a
four times larger factual collection for its disturbance readout. The oracle
references of HandReach, Ball-in-Cup, FetchReach and BlueROV2 are Monte Carlo
estimates that agree with independent numerical refinements but carry no formal
numerical-error bound, and PassWater's uses only four disturbance draws per
source state (\cref{app:independent-benchmark-design}).

\textbf{Open directions.}~Each limit suggests a direction. (i) The paired trials
above could tell, before deployment, where residual reuse is safe on a real
system. (ii) Since the error is set by the interaction in the chosen coordinates,
an encoder or readout could be trained, in simulation or from such trials, so
that actions and disturbances add, turning additivity from a property we check
into one we learn. (iii) Contacts and gain faults call for transport rules that
switch with the contact mode or rescale the residual with the command, combining
residual reuse with the other structures of \cref{tab:map}. (iv) Factual error
over the whole feature vector is an imperfect proxy for the action contrast:
selecting the factual predictor by held-out error improves or preserves the
counterfactual in five of six environments but not on Push-T, and a criterion
targeted at the action response could close part of the gap left by learning.
(v) Chaining counterfactuals over several transitions, each with its own observed
outcome, would extend the uses that motivated counterfactual reasoning in
reinforcement learning, such as credit assignment, data augmentation and
explanation
\citep{mesnard2021counterfactual,pitis2020counterfactual,tsirtsis2021counterfactual},
to learned world models of physical systems. Finally, the argument never uses
that the system is a robot: any process with smooth dynamics and a hidden
perturbation that enters through its own channel admits the same answer after a
surprise. We hope this study makes the
observed outcome a first-class input of world models, as offset-free control
has long made it for its own models.

\textbf{Acknowledgments.}~We thank Marcin Przewi\k{e}\'zlikowski and Jakob
Ambsdorf for their careful reading of this manuscript and for feedback that
improved its presentation.

\clearpage
\bibliographystyle{plainnat}
\bibliography{bibliography}

\begin{thebibliography}{64}
\providecommand{\natexlab}[1]{#1}
\providecommand{\url}[1]{\texttt{#1}}
\expandafter\ifx\csname urlstyle\endcsname\relax
  \providecommand{\doi}[1]{doi: #1}\else
  \providecommand{\doi}{doi: \begingroup \urlstyle{rm}\Url}\fi

\bibitem[Abbeel et~al.(2006)Abbeel, Quigley, and Ng]{abbeel2006using}
Pieter Abbeel, Morgan Quigley, and Andrew~Y. Ng.
\newblock Using inaccurate models in reinforcement learning.
\newblock In \emph{Proceedings of the 23rd International Conference on Machine
  Learning (ICML)}, pages 1--8, 2006.
\newblock \doi{10.1145/1143844.1143845}.

\bibitem[Arimoto et~al.(1984)Arimoto, Kawamura, and
  Miyazaki]{arimoto1984bettering}
Suguru Arimoto, Sadao Kawamura, and Fumio Miyazaki.
\newblock Bettering operation of robots by learning.
\newblock \emph{Journal of Robotic Systems}, 1\penalty0 (2):\penalty0 123--140,
  1984.
\newblock \doi{10.1002/rob.4620010203}.

\bibitem[Assran et~al.(2023)Assran, Duval, Misra, Bojanowski, Vincent, Rabbat,
  LeCun, and Ballas]{assran2023self}
Mahmoud Assran, Quentin Duval, Ishan Misra, Piotr Bojanowski, Pascal Vincent,
  Michael Rabbat, Yann LeCun, and Nicolas Ballas.
\newblock Self-supervised learning from images with a joint-embedding
  predictive architecture.
\newblock In \emph{Proceedings of the IEEE/CVF Conference on Computer Vision
  and Pattern Recognition}, pages 15619--15629, 2023.

\bibitem[Assran et~al.(2025)Assran, Bardes, Fan, Garrido, Howes, Komeili,
  Muckley, Rizvi, Roberts, Sinha, Zholus, Arnaud, Gejji, Martin, Hogan, Dugas,
  Bojanowski, Khalidov, Labatut, Massa, Szafraniec, Krishnakumar, Li, Ma,
  Chandar, Meier, LeCun, Rabbat, and Ballas]{assran2025vjepa2}
Mido Assran, Adrien Bardes, David Fan, Quentin Garrido, Russell Howes, Mojtaba
  Komeili, Matthew Muckley, Ammar Rizvi, Claire Roberts, Koustuv Sinha, Artem
  Zholus, Sergio Arnaud, Abha Gejji, Ada Martin, Francois~Robert Hogan, Daniel
  Dugas, Piotr Bojanowski, Vasil Khalidov, Patrick Labatut, Francisco Massa,
  Marc Szafraniec, Kapil Krishnakumar, Yong Li, Xiaodong Ma, Sarath Chandar,
  Franziska Meier, Yann LeCun, Michael Rabbat, and Nicolas Ballas.
\newblock {V-JEPA} 2: Self-supervised video models enable understanding,
  prediction and planning, 2025.
\newblock URL \url{https://arxiv.org/abs/2506.09985}.

\bibitem[Athey and Imbens(2006)]{athey2006identification}
Susan Athey and Guido~W. Imbens.
\newblock Identification and inference in nonlinear difference-in-differences
  models.
\newblock \emph{Econometrica}, 74\penalty0 (2):\penalty0 431--497, 2006.
\newblock \doi{10.1111/j.1468-0262.2006.00668.x}.

\bibitem[Baradel et~al.(2020)Baradel, Neverova, Mille, Mori, and
  Wolf]{baradel2020cophy}
Fabien Baradel, Natalia Neverova, Julien Mille, Greg Mori, and Christian Wolf.
\newblock {CoPhy}: Counterfactual learning of physical dynamics.
\newblock In \emph{International Conference on Learning Representations
  (ICLR)}, 2020.
\newblock URL \url{https://arxiv.org/abs/1909.12000}.

\bibitem[Bardes et~al.(2024)Bardes, Garrido, Ponce, Chen, Rabbat, LeCun,
  Assran, and Ballas]{bardes2024vjepa}
Adrien Bardes, Quentin Garrido, Jean Ponce, Xinlei Chen, Michael Rabbat, Yann
  LeCun, Mahmoud Assran, and Nicolas Ballas.
\newblock Revisiting feature prediction for learning visual representations
  from video.
\newblock \emph{Transactions on Machine Learning Research}, 2024.
\newblock URL \url{https://arxiv.org/abs/2404.08471}.

\bibitem[Bareinboim et~al.(2022)Bareinboim, Correa, Ibeling, and
  Icard]{bareinboim2022pearl}
Elias Bareinboim, Juan~D. Correa, Duligur Ibeling, and Thomas Icard.
\newblock On {Pearl's} hierarchy and the foundations of causal inference.
\newblock In Hector Geffner, Rina Dechter, and Joseph~Y. Halpern, editors,
  \emph{Probabilistic and Causal Inference: The Works of Judea Pearl},
  volume~36 of \emph{ACM Books}, pages 507--556. Association for Computing
  Machinery, 2022.
\newblock \doi{10.1145/3501714.3501743}.

\bibitem[Bear et~al.(2023)Bear, Feigelis, Chen, Lee, Venkatesh, Kotar, Durango,
  and Yamins]{bear2023cwm}
Daniel~M. Bear, Kevin Feigelis, Honglin Chen, Wanhee Lee, Rahul Venkatesh,
  Klemen Kotar, Alex Durango, and Daniel L.~K. Yamins.
\newblock Unifying (machine) vision via counterfactual world modeling.
\newblock \emph{arXiv preprint arXiv:2306.01828}, 2023.

\bibitem[Bica et~al.(2020)Bica, Alaa, Jordon, and van~der
  Schaar]{bica2020estimating}
Ioana Bica, Ahmed~M. Alaa, James Jordon, and Mihaela van~der Schaar.
\newblock Estimating counterfactual treatment outcomes over time through
  adversarially balanced representations.
\newblock In \emph{International Conference on Learning Representations
  (ICLR)}, 2020.
\newblock URL \url{https://arxiv.org/abs/2002.04083}.

\bibitem[Bristow et~al.(2006)Bristow, Tharayil, and Alleyne]{bristow2006survey}
Douglas~A. Bristow, Marina Tharayil, and Andrew~G. Alleyne.
\newblock A survey of iterative learning control.
\newblock \emph{IEEE Control Systems Magazine}, 26\penalty0 (3):\penalty0
  96--114, 2006.
\newblock \doi{10.1109/MCS.2006.1636313}.

\bibitem[Buesing et~al.(2019)Buesing, Weber, Zwols, Racani{\`e}re, Guez,
  Lespiau, and Heess]{buesing2019woulda}
Lars Buesing, Theophane Weber, Yori Zwols, Sebastien Racani{\`e}re, Arthur
  Guez, Jean-Baptiste Lespiau, and Nicolas Heess.
\newblock Woulda, coulda, shoulda: Counterfactually-guided policy search.
\newblock In \emph{International Conference on Learning Representations
  (ICLR)}, 2019.
\newblock URL \url{https://arxiv.org/abs/1811.06272}.

\bibitem[Chen et~al.(2016)Chen, Yang, Guo, and Li]{chen2016disturbance}
Wen-Hua Chen, Jun Yang, Lei Guo, and Shihua Li.
\newblock Disturbance-observer-based control and related methods---an overview.
\newblock \emph{IEEE Transactions on Industrial Electronics}, 63\penalty0
  (2):\penalty0 1083--1095, 2016.
\newblock \doi{10.1109/TIE.2015.2478397}.

\bibitem[Chernozhukov and Hansen(2005)]{chernozhukov2005iv}
Victor Chernozhukov and Christian Hansen.
\newblock An {IV} model of quantile treatment effects.
\newblock \emph{Econometrica}, 73\penalty0 (1):\penalty0 245--261, 2005.
\newblock \doi{10.1111/j.1468-0262.2005.00570.x}.

\bibitem[Chernozhukov et~al.(2018)Chernozhukov, Chetverikov, Demirer, Duflo,
  Hansen, Newey, and Robins]{chernozhukov2018}
Victor Chernozhukov, Denis Chetverikov, Mert Demirer, Esther Duflo, Christian
  Hansen, Whitney Newey, and James Robins.
\newblock Double/debiased machine learning for treatment and structural
  parameters.
\newblock \emph{The Econometrics Journal}, 21\penalty0 (1):\penalty0 C1--C68,
  2018.
\newblock \doi{10.1111/ectj.12097}.

\bibitem[Fang et~al.(2023)Fang, Sun, Wang, Huang, Wang, and Cao]{fang2023eva02}
Yuxin Fang, Quan Sun, Xinggang Wang, Tiejun Huang, Xinlong Wang, and Yue Cao.
\newblock {EVA-02}: A visual representation for neon genesis.
\newblock \emph{arXiv preprint arXiv:2303.11331}, 2023.
\newblock URL \url{https://arxiv.org/abs/2303.11331}.

\bibitem[Ha and Schmidhuber(2018)]{ha2018recurrent}
David Ha and J{\"u}rgen Schmidhuber.
\newblock Recurrent world models facilitate policy evolution.
\newblock In \emph{Advances in Neural Information Processing Systems
  (NeurIPS)}, volume~31, 2018.
\newblock URL
  \url{https://proceedings.neurips.cc/paper_files/paper/2018/hash/2de5d16682c3c35007e4e92982f1a2ba-Abstract.html}.

\bibitem[Hafner et~al.(2019)Hafner, Lillicrap, Fischer, Villegas, Ha, Lee, and
  Davidson]{hafner2019learning}
Danijar Hafner, Timothy Lillicrap, Ian Fischer, Ruben Villegas, David Ha,
  Honglak Lee, and James Davidson.
\newblock Learning latent dynamics for planning from pixels.
\newblock In \emph{International Conference on Machine Learning}, pages
  2555--2565, 2019.

\bibitem[Hafner et~al.(2020)Hafner, Lillicrap, Ba, and
  Norouzi]{hafner2020dream}
Danijar Hafner, Timothy Lillicrap, Jimmy Ba, and Mohammad Norouzi.
\newblock Dream to control: Learning behaviors by latent imagination.
\newblock In \emph{International Conference on Learning Representations
  (ICLR)}, 2020.

\bibitem[Hafner et~al.(2025)Hafner, Pasukonis, Ba, and
  Lillicrap]{hafner2025mastering}
Danijar Hafner, Jurgis Pasukonis, Jimmy Ba, and Timothy Lillicrap.
\newblock Mastering diverse control tasks through world models.
\newblock \emph{Nature}, 640\penalty0 (8059):\penalty0 647--653, 2025.
\newblock \doi{10.1038/s41586-025-08744-2}.

\bibitem[Hansen et~al.(2024)Hansen, Su, and Wang]{hansen2024tdmpc2}
Nicklas Hansen, Hao Su, and Xiaolong Wang.
\newblock {TD-MPC2}: Scalable, robust world models for continuous control.
\newblock In \emph{International Conference on Learning Representations
  (ICLR)}, 2024.
\newblock URL \url{https://openreview.net/forum?id=Oxh5CstDJU}.

\bibitem[Haugh and Singal(2023)]{haugh2023counterfactual}
Martin~B. Haugh and Raghav Singal.
\newblock Counterfactual analysis in dynamic latent state models.
\newblock In \emph{International Conference on Machine Learning (ICML)}, volume
  202 of \emph{Proceedings of Machine Learning Research}, pages 12647--12677,
  2023.
\newblock URL \url{https://proceedings.mlr.press/v202/haugh23a.html}.

\bibitem[He et~al.(2022)He, Chen, Xie, Li, Doll{\'a}r, and
  Girshick]{he2022masked}
Kaiming He, Xinlei Chen, Saining Xie, Yanghao Li, Piotr Doll{\'a}r, and Ross
  Girshick.
\newblock Masked autoencoders are scalable vision learners.
\newblock In \emph{Proceedings of the IEEE/CVF Conference on Computer Vision
  and Pattern Recognition}, 2022.
\newblock URL \url{https://arxiv.org/abs/2111.06377}.

\bibitem[Heckman et~al.(1997)Heckman, Smith, and
  Clements]{heckman1997heterogeneity}
James~J. Heckman, Jeffrey Smith, and Nancy Clements.
\newblock Making the most out of programme evaluations and social experiments:
  Accounting for heterogeneity in programme impacts.
\newblock \emph{The Review of Economic Studies}, 64\penalty0 (4):\penalty0
  487--535, 1997.
\newblock \doi{10.2307/2971729}.

\bibitem[Hern{\'a}n and Robins(2020)]{hernan2020causal}
Miguel~A. Hern{\'a}n and James~M. Robins.
\newblock \emph{Causal Inference: What If}.
\newblock Chapman \& Hall/CRC, Boca Raton, 2020.

\bibitem[Hoyer et~al.(2008)Hoyer, Janzing, Mooij, Peters, and
  Sch{\"o}lkopf]{hoyer2008nonlinear}
Patrik Hoyer, Dominik Janzing, Joris Mooij, Jonas Peters, and Bernhard
  Sch{\"o}lkopf.
\newblock Nonlinear causal discovery with additive noise models.
\newblock In \emph{Advances in Neural Information Processing Systems (NIPS)},
  volume~21, 2008.
\newblock URL
  \url{https://proceedings.neurips.cc/paper_files/paper/2008/hash/f7664060cc52bc6f3d620bcedc94a4b6-Abstract.html}.

\bibitem[Janny et~al.(2022)Janny, Baradel, Neverova, Nadri, Mori, and
  Wolf]{janny2022filtered}
Steeven Janny, Fabien Baradel, Natalia Neverova, Madiha Nadri, Greg Mori, and
  Christian Wolf.
\newblock Filtered-{CoPhy}: Unsupervised learning of counterfactual physics in
  pixel space.
\newblock In \emph{International Conference on Learning Representations
  (ICLR)}, 2022.
\newblock URL \url{https://arxiv.org/abs/2202.00368}.

\bibitem[LeCun(2022)]{lecun2022path}
Yann LeCun.
\newblock A path towards autonomous machine intelligence.
\newblock \emph{OpenReview preprint}, 2022.
\newblock URL \url{https://openreview.net/forum?id=BZ5a1r-kVsf}.

\bibitem[Lim et~al.(2018)Lim, Alaa, and van~der Schaar]{lim2018forecasting}
Bryan Lim, Ahmed Alaa, and Mihaela van~der Schaar.
\newblock Forecasting treatment responses over time using recurrent marginal
  structural networks.
\newblock In \emph{Advances in Neural Information Processing Systems
  (NeurIPS)}, volume~31, 2018.
\newblock URL
  \url{https://proceedings.neurips.cc/paper_files/paper/2018/hash/56e6a93212e4482d99c84a639d254b67-Abstract.html}.

\bibitem[Lu et~al.(2020)Lu, Huang, Wang, Hern{\'a}ndez-Lobato, Zhang, and
  Sch{\"o}lkopf]{lu2020sample}
Chaochao Lu, Biwei Huang, Ke~Wang, Jos{\'e}~Miguel Hern{\'a}ndez-Lobato, Kun
  Zhang, and Bernhard Sch{\"o}lkopf.
\newblock Sample-efficient reinforcement learning via counterfactual-based data
  augmentation, 2020.
\newblock URL \url{https://arxiv.org/abs/2012.09092}.
\newblock NeurIPS 2020 Workshop on Offline Reinforcement Learning.

\bibitem[Ma et~al.(2026)Ma, Shi, and Xu]{ma2026twin}
Yu~Ma, Hongli Shi, and Xinran Xu.
\newblock Twin rollouts: Noise-coupled counterfactual branching in interactive
  video world models, 2026.
\newblock URL \url{https://arxiv.org/abs/2608.08982}.

\bibitem[Melnychuk et~al.(2022)Melnychuk, Frauen, and
  Feuerriegel]{melnychuk2022causal}
Valentyn Melnychuk, Dennis Frauen, and Stefan Feuerriegel.
\newblock Causal transformer for estimating counterfactual outcomes.
\newblock In \emph{International Conference on Machine Learning (ICML)}, volume
  162 of \emph{Proceedings of Machine Learning Research}, pages 15293--15329,
  2022.
\newblock URL \url{https://proceedings.mlr.press/v162/melnychuk22a.html}.

\bibitem[Mesnard et~al.(2021)Mesnard, Weber, Viola, Thakoor, Saade,
  Harutyunyan, Dabney, Stepleton, Heess, Guez, Moulines, Hutter, Buesing, and
  Munos]{mesnard2021counterfactual}
Thomas Mesnard, Theophane Weber, Fabio Viola, Shantanu Thakoor, Alaa Saade,
  Anna Harutyunyan, Will Dabney, Thomas~S. Stepleton, Nicolas Heess, Arthur
  Guez, Eric Moulines, Marcus Hutter, Lars Buesing, and R{\'e}mi Munos.
\newblock Counterfactual credit assignment in model-free reinforcement
  learning.
\newblock In \emph{International Conference on Machine Learning (ICML)}, volume
  139 of \emph{Proceedings of Machine Learning Research}, pages 7654--7664,
  2021.
\newblock URL \url{https://proceedings.mlr.press/v139/mesnard21a.html}.

\bibitem[Mooij et~al.(2013)Mooij, Janzing, and Sch{\"o}lkopf]{mooij2013ode}
Joris~M. Mooij, Dominik Janzing, and Bernhard Sch{\"o}lkopf.
\newblock From ordinary differential equations to structural causal models: The
  deterministic case.
\newblock In \emph{Proceedings of the Twenty-Ninth Conference on Uncertainty in
  Artificial Intelligence (UAI)}, pages 440--448, 2013.
\newblock URL \url{https://arxiv.org/abs/1408.2063}.

\bibitem[Muske and Badgwell(2002)]{muske2002disturbance}
Kenneth~R. Muske and Thomas~A. Badgwell.
\newblock Disturbance modeling for offset-free linear model predictive control.
\newblock \emph{Journal of Process Control}, 12\penalty0 (5):\penalty0
  617--632, 2002.
\newblock \doi{10.1016/S0959-1524(01)00051-8}.

\bibitem[Nasr-Esfahany and Kiciman(2023)]{nasresfahany2023nonidentifiability}
Arash Nasr-Esfahany and Emre Kiciman.
\newblock Counterfactual (non-)identifiability of learned structural causal
  models, 2023.
\newblock URL \url{https://arxiv.org/abs/2301.09031}.

\bibitem[Nasr-Esfahany et~al.(2023)Nasr-Esfahany, Alizadeh, and
  Shah]{nasresfahany2023counterfactual}
Arash Nasr-Esfahany, Mohammad Alizadeh, and Devavrat Shah.
\newblock Counterfactual identifiability of bijective causal models.
\newblock In \emph{International Conference on Machine Learning (ICML)}, volume
  202 of \emph{Proceedings of Machine Learning Research}, pages 25733--25754,
  2023.
\newblock URL \url{https://proceedings.mlr.press/v202/nasr-esfahany23a.html}.

\bibitem[Oberst and Sontag(2019)]{oberst2019counterfactual}
Michael Oberst and David Sontag.
\newblock Counterfactual off-policy evaluation with {G}umbel-max structural
  causal models.
\newblock In \emph{International Conference on Machine Learning (ICML)}, pages
  4881--4890, 2019.
\newblock URL \url{https://arxiv.org/abs/1905.05824}.

\bibitem[Ohishi et~al.(1987)Ohishi, Nakao, Ohnishi, and
  Miyachi]{ohishi1987microprocessor}
Kiyoshi Ohishi, Masato Nakao, Kouhei Ohnishi, and Kunio Miyachi.
\newblock Microprocessor-controlled {DC} motor for load-insensitive position
  servo system.
\newblock \emph{IEEE Transactions on Industrial Electronics}, IE-34\penalty0
  (1):\penalty0 44--49, 1987.
\newblock \doi{10.1109/TIE.1987.350923}.

\bibitem[Oquab et~al.(2023)Oquab, Darcet, Moutakanni, Vo, Szafraniec, Khalidov,
  Fernandez, Haziza, Massa, El-Nouby, et~al.]{oquab2023dinov2}
Maxime Oquab, Timoth{\'e}e Darcet, Th{\'e}o Moutakanni, Huy Vo, Marc
  Szafraniec, Vasil Khalidov, Pierre Fernandez, Daniel Haziza, Francisco Massa,
  Alaaeldin El-Nouby, et~al.
\newblock Dinov2: Learning robust visual features without supervision.
\newblock \emph{arXiv preprint arXiv:2304.07193}, 2023.

\bibitem[Pannocchia and Rawlings(2003)]{pannocchia2003disturbance}
Gabriele Pannocchia and James~B. Rawlings.
\newblock Disturbance models for offset-free model-predictive control.
\newblock \emph{AIChE Journal}, 49\penalty0 (2):\penalty0 426--437, 2003.
\newblock \doi{10.1002/aic.690490213}.

\bibitem[Park et~al.(2025)Park, Frans, Eysenbach, and Levine]{ogbench_park2025}
Seohong Park, Kevin Frans, Benjamin Eysenbach, and Sergey Levine.
\newblock {OGBench}: Benchmarking offline goal-conditioned {RL}.
\newblock In \emph{International Conference on Learning Representations}, 2025.
\newblock URL
  \url{https://proceedings.iclr.cc/paper_files/paper/2025/hash/ecd92623ac899357312aaa8915853699-Abstract-Conference.html}.

\bibitem[Pawlowski et~al.(2020)Pawlowski, Castro, and
  Glocker]{pawlowski2020dscm}
Nick Pawlowski, Daniel~C. Castro, and Ben Glocker.
\newblock Deep structural causal models for tractable counterfactual inference.
\newblock In \emph{Advances in Neural Information Processing Systems
  (NeurIPS)}, volume~33, pages 857--869, 2020.
\newblock URL \url{https://arxiv.org/abs/2006.06485}.

\bibitem[Pearl(2009)]{pearl2009causality}
Judea Pearl.
\newblock \emph{Causality: Models, Reasoning, and Inference}.
\newblock Cambridge University Press, 2nd edition, 2009.

\bibitem[Peters et~al.(2017)Peters, Janzing, and
  Sch{\"o}lkopf]{peters2017elements}
Jonas Peters, Dominik Janzing, and Bernhard Sch{\"o}lkopf.
\newblock \emph{Elements of Causal Inference: Foundations and Learning
  Algorithms}.
\newblock Adaptive Computation and Machine Learning. MIT Press, Cambridge, MA,
  2017.

\bibitem[Peters et~al.(2022)Peters, Bauer, and Pfister]{peters2022causal}
Jonas Peters, Stefan Bauer, and Niklas Pfister.
\newblock Causal models for dynamical systems.
\newblock In Hector Geffner, Rina Dechter, and Joseph~Y. Halpern, editors,
  \emph{Probabilistic and Causal Inference: The Works of Judea Pearl},
  volume~36 of \emph{ACM Books}, pages 671--690. Association for Computing
  Machinery, 2022.
\newblock \doi{10.1145/3501714.3501752}.

\bibitem[Pitis et~al.(2020)Pitis, Creager, and Garg]{pitis2020counterfactual}
Silviu Pitis, Elliot Creager, and Animesh Garg.
\newblock Counterfactual data augmentation using locally factored dynamics.
\newblock In \emph{Advances in Neural Information Processing Systems
  (NeurIPS)}, volume~33, pages 3976--3990, 2020.
\newblock URL
  \url{https://proceedings.neurips.cc/paper_files/paper/2020/hash/294e09f267683c7ddc6cc5134a7e68a8-Abstract.html}.

\bibitem[Pitis et~al.(2022)Pitis, Creager, Mandlekar, and
  Garg]{pitis2022mocoda}
Silviu Pitis, Elliot Creager, Ajay Mandlekar, and Animesh Garg.
\newblock {MoCoDA}: Model-based counterfactual data augmentation.
\newblock In \emph{Advances in Neural Information Processing Systems
  (NeurIPS)}, volume~35, pages 18143--18156, 2022.
\newblock URL
  \url{https://proceedings.neurips.cc/paper_files/paper/2022/hash/7314e20a73542bbfff25030d1185ce88-Abstract-Conference.html}.

\bibitem[Qiu et~al.(2026)Qiu, Chen, Cao, Shi, Huo, and Gao]{qiu2026adwm}
Jiabin Qiu, Zixuan Chen, Hongye Cao, Jieqi Shi, Jing Huo, and Yang Gao.
\newblock {AD-WM}: Action-discriminative world models for counterfactual model
  predictive control, 2026.
\newblock URL \url{https://arxiv.org/abs/2609.30264}.

\bibitem[Radford et~al.(2021)Radford, Kim, Hallacy, Ramesh, Goh, Agarwal,
  Sastry, Askell, Mishkin, Clark, Krueger, and Sutskever]{radford2021learning}
Alec Radford, Jong~Wook Kim, Chris Hallacy, Aditya Ramesh, Gabriel Goh,
  Sandhini Agarwal, Girish Sastry, Amanda Askell, Pamela Mishkin, Jack Clark,
  Gretchen Krueger, and Ilya Sutskever.
\newblock Learning transferable visual models from natural language
  supervision.
\newblock \emph{arXiv preprint arXiv:2103.00020}, 2021.
\newblock URL \url{https://arxiv.org/abs/2103.00020}.

\bibitem[Rubenstein et~al.(2018)Rubenstein, Bongers, Sch{\"o}lkopf, and
  Mooij]{rubenstein2018deterministic}
Paul~K. Rubenstein, Stephan Bongers, Bernhard Sch{\"o}lkopf, and Joris~M.
  Mooij.
\newblock From deterministic {ODEs} to dynamic structural causal models.
\newblock In \emph{Conference on Uncertainty in Artificial Intelligence (UAI)},
  pages 114--123, 2018.
\newblock URL \url{https://www.auai.org/uai2018/proceedings/papers/43.pdf}.

\bibitem[Rubin(1974)]{rubin1974estimating}
Donald~B. Rubin.
\newblock Estimating causal effects of treatments in randomized and
  nonrandomized studies.
\newblock \emph{Journal of Educational Psychology}, 66\penalty0 (5):\penalty0
  688--701, 1974.
\newblock \doi{10.1037/h0037350}.

\bibitem[Schrittwieser et~al.(2020)Schrittwieser, Antonoglou, Hubert, Simonyan,
  Sifre, Schmitt, Guez, Lockhart, Hassabis, Graepel, Lillicrap, and
  Silver]{schrittwieser2020mastering}
Julian Schrittwieser, Ioannis Antonoglou, Thomas Hubert, Karen Simonyan,
  Laurent Sifre, Simon Schmitt, Arthur Guez, Edward Lockhart, Demis Hassabis,
  Thore Graepel, Timothy Lillicrap, and David Silver.
\newblock Mastering {Atari}, {Go}, chess and shogi by planning with a learned
  model.
\newblock \emph{Nature}, 588:\penalty0 604--609, 2020.
\newblock \doi{10.1038/s41586-020-03051-4}.

\bibitem[Shalit et~al.(2017)Shalit, Johansson, and
  Sontag]{shalit2017estimating}
Uri Shalit, Fredrik~D. Johansson, and David Sontag.
\newblock Estimating individual treatment effect: Generalization bounds and
  algorithms.
\newblock In \emph{International Conference on Machine Learning (ICML)},
  volume~70 of \emph{Proceedings of Machine Learning Research}, pages
  3076--3085, 2017.
\newblock URL \url{https://proceedings.mlr.press/v70/shalit17a.html}.

\bibitem[Shimodaira(2000)]{shimodaira2000improving}
Hidetoshi Shimodaira.
\newblock Improving predictive inference under covariate shift by weighting the
  log-likelihood function.
\newblock \emph{Journal of Statistical Planning and Inference}, 90\penalty0
  (2):\penalty0 227--244, 2000.
\newblock \doi{10.1016/S0378-3758(00)00115-4}.

\bibitem[Sim{\'e}oni et~al.(2025)Sim{\'e}oni, Vo, Seitzer, Baldassarre, Oquab,
  Jose, Khalidov, Szafraniec, Yi, Ramamonjisoa, et~al.]{simeoni2025dinov3}
Oriane Sim{\'e}oni, Huy~V Vo, Maximilian Seitzer, Federico Baldassarre, Maxime
  Oquab, Cijo Jose, Vasil Khalidov, Marc Szafraniec, Seungeun Yi, Micha{\"e}l
  Ramamonjisoa, et~al.
\newblock Dinov3.
\newblock \emph{arXiv preprint arXiv:2508.10104}, 2025.

\bibitem[Sobal et~al.(2025)Sobal, Zhang, Cho, Balestriero, Rudner, and
  LeCun]{sobal2025learning}
Vlad Sobal, Wancong Zhang, Kyunghyun Cho, Randall Balestriero, Tim G.~J.
  Rudner, and Yann LeCun.
\newblock Learning from reward-free offline data: A case for planning with
  latent dynamics models.
\newblock In \emph{Advances in Neural Information Processing Systems
  (NeurIPS)}, volume~38, pages 43905--43941, 2025.
\newblock \doi{10.52202/085713-1465}.
\newblock URL
  \url{https://proceedings.neurips.cc/paper_files/paper/2025/hash/3e7cf447f21cd11c846463affefce665-Abstract-Conference.html}.

\bibitem[Tschannen et~al.(2025)Tschannen, Gritsenko, Wang, Naeem,
  Alabdulmohsin, Parthasarathy, Evans, Beyer, Xia, Mustafa, H{\'e}naff,
  Harmsen, Steiner, and Zhai]{tschannen2025siglip2}
Michael Tschannen, Alexey Gritsenko, Xiao Wang, Muhammad~Ferjad Naeem, Ibrahim
  Alabdulmohsin, Nikhil Parthasarathy, Talfan Evans, Lucas Beyer, Ye~Xia, Basil
  Mustafa, Olivier H{\'e}naff, Jeremiah Harmsen, Andreas Steiner, and Xiaohua
  Zhai.
\newblock {SigLIP 2}: Multilingual vision-language encoders with improved
  semantic understanding, localization, and dense features.
\newblock \emph{arXiv preprint arXiv:2502.14786}, 2025.
\newblock URL \url{https://arxiv.org/abs/2502.14786}.

\bibitem[Tsirtsis and Gomez-Rodriguez(2023)]{tsirtsis2023finding}
Stratis Tsirtsis and Manuel Gomez-Rodriguez.
\newblock Finding counterfactually optimal action sequences in continuous state
  spaces.
\newblock In \emph{Advances in Neural Information Processing Systems
  (NeurIPS)}, volume~36, pages 3220--3247, 2023.
\newblock URL
  \url{https://proceedings.neurips.cc/paper_files/paper/2023/hash/09ae6beae5f1ff38f05c05979097ea0f-Abstract-Conference.html}.

\bibitem[Tsirtsis et~al.(2021)Tsirtsis, De, and
  Gomez-Rodriguez]{tsirtsis2021counterfactual}
Stratis Tsirtsis, Abir De, and Manuel Gomez-Rodriguez.
\newblock Counterfactual explanations in sequential decision making under
  uncertainty.
\newblock In \emph{Advances in Neural Information Processing Systems
  (NeurIPS)}, volume~34, pages 30127--30139, 2021.
\newblock URL
  \url{https://proceedings.neurips.cc/paper_files/paper/2021/hash/fd0a5a5e367a0955d81278062ef37429-Abstract.html}.

\bibitem[Wu et~al.(2025)Wu, Li, Zheng, Zeng, Chen, Liu, Guo, and
  Zhang]{wu2025learning}
Peng Wu, Haoxuan Li, Chunyuan Zheng, Yan Zeng, Jiawei Chen, Yang Liu, Ruocheng
  Guo, and Kun Zhang.
\newblock Learning counterfactual outcomes under rank preservation.
\newblock In \emph{Advances in Neural Information Processing Systems
  (NeurIPS)}, volume~38, pages 148262--148289, 2025.
\newblock \doi{10.52202/085713-4955}.
\newblock URL
  \url{https://proceedings.neurips.cc/paper_files/paper/2025/hash/d9e3587500916105767f0d4e05a29a09-Abstract-Conference.html}.

\bibitem[Xia et~al.(2023)Xia, Pan, and Bareinboim]{xia2023neural}
Kevin Xia, Yushu Pan, and Elias Bareinboim.
\newblock Neural causal models for counterfactual identification and
  estimation.
\newblock In \emph{International Conference on Learning Representations
  (ICLR)}, 2023.
\newblock URL \url{https://openreview.net/forum?id=vouQcZS8KfW}.

\bibitem[Zhang et~al.(2026)Zhang, Cui, Xu, Ye, Zhang, and
  Wang]{zhang2026driving}
Jiaru Zhang, Can Cui, Yi~Xu, Xin Ye, Ruqi Zhang, and Ziran Wang.
\newblock How can driving world models do counterfactual prediction?, 2026.
\newblock URL \url{https://arxiv.org/abs/2608.11601}.

\bibitem[Zhou et~al.(2025)Zhou, Pan, LeCun, and Pinto]{zhou2024dinowm}
Gaoyue Zhou, Hengkai Pan, Yann LeCun, and Lerrel Pinto.
\newblock {DINO-WM}: World models on pre-trained visual features enable
  zero-shot planning.
\newblock In \emph{International Conference on Machine Learning (ICML)}, volume
  267 of \emph{Proceedings of Machine Learning Research}, pages 79115--79135,
  2025.

\end{thebibliography}

\makeatletter
\let\fr@apx@subsection\@gobbletwo
\makeatother
\appendix

\section{Tables for the Main-Text Figures}
\label{app:complete-numbers}
All $R^2$ values are percentages. CF WM denotes counterfactual world model.
\begin{table}[htbp]
\centering\footnotesize
\setlength{\tabcolsep}{3.5pt}
\begin{tabular}{llllrrrr}
\toprule
Environment & Coordinates & Mechanism & Theory & Oracle & Copy & Ideal resid. & $\rho$\\
\midrule
HandReach & wrist angles$^\dagger$ & direct offset & additive & 49.1 & 7.3 & 98.3 & 0.033\\
HandReach & fingertips$^\ddagger$ & forward kinematics & smooth & 49.6--57.3 & $-9.6$--$-0.2$ & 94.6--98.9 & 0.025--0.116\\
BlueROV2 & body velocities & hydrodynamics & smooth & 50.4 & 0.5 & 96.9 & 0.062\\
BlueROV2 & $\sin(\mathrm{pitch})^\dagger$ & direct thrust & additive & 51.8 & 1.4 & 100.0 & 0.000\\
BlueROV2 & $\cos(\mathrm{pitch})$ & folded coordinate & fold & 16.9 & $-73.1$ & $-38.4$ & 1.666\\
Push-T & pusher $x,y^\dagger$ & kinematic & additive & 51.0 & 3.5 & 100.0 & 0.000\\
Push-T & block $v_x$ & contact & non-smooth & 36.4 & $-33.8$ & 5.6 & 1.485\\
Push-T & block $v_y$ & contact & non-smooth & 38.8 & $-64.4$ & $-15.5$ & 1.889\\
Push-T & block $\omega$ & contact & non-smooth & 19.9 & $-75.8$ & $-44.4$ & 1.802\\
\bottomrule
\end{tabular}
\caption{\textbf{Physical interaction diagnostics.}
Simulator-only screen on the benchmark evaluation population; $R^2$ in percent. \emph{Ideal residual} applies residual reuse
with within-parent Monte Carlo estimates of the conditional means in the listed physical coordinates;
$\rho$ is the interaction ratio of \cref{eq:interaction-ratio}, averaged over
parents and query pairs. \emph{Oracle} and \emph{Copy} are the oracle predictor (no outcome)
and factual copying; both below $60\%$ marks a non-trivial counterfactual problem.
$^\dagger$Primary benchmark coordinates. $^\ddagger$The seven fingertip
coordinates whose oracle and copy references are both below
$60\%$; ranges over coordinates. No image, readout or fitted model is used.
Per-coordinate values for every recorded coordinate of all seven
environments accompany the code release.}
\label{tab:stress-physical}
\end{table}

\begin{table}[htbp]
\centering\small
\setlength{\tabcolsep}{5pt}
\begin{tabular}{lrrrrrr}
\toprule
Environment & $\lambda$ & Oracle & Copy & Ideal residual & $\rho$ measured & $\rho$ predicted\\
\midrule
HandReach & 0 & 49.1 & 7.3 & 98.3 & 0.033 & 0\\
 & 0.5 & 53.5 & $-23.4$ & 76.9 & 0.497 & 0.5\\
 & 1 & 53.5 & $-87.2$ & 6.7 & 2.006 & 2\\
\addlinespace
Ball-in-Cup & 0 & 51.1 & 6.7 & 99.8 & 0.003 & 0\\
 & 0.5 & 52.1 & $-16.3$ & 77.0 & 0.479 & 0.5\\
 & 1 & 52.2 & $-86.9$ & 5.9 & 1.969 & 2\\
\bottomrule
\end{tabular}
\caption{\textbf{Actuator-offset to gain-fault sweep.}
Simulator-only, on the primary actuated coordinates; $R^2$ in percent. Rows
with $\lambda=0$ are the benchmark confirmation cohorts; rows with $\lambda>0$ are
independent cohorts of 256 parents $\times$ 64 disturbance draws $\times$ 3
queries, recalibrated analytically to the same oracle (no outcome) target before any model,
with all native control tests passing. Parent-bootstrap $95\%$ intervals for
$\rho$ are $[0.45,0.54]$ and $[0.44,0.52]$ at $\lambda=0.5$ and $[1.86,2.15]$ and
$[1.84,2.10]$ at $\lambda=1$. The predicted column is the equilibrium design
model; no image, readout or fitted model is used.}
\label{tab:dial}
\end{table}

\begin{table}[htbp]\centering\small
\setlength{\tabcolsep}{3.5pt}
\begin{tabular}{llrrrllr}\toprule
& & & & Oracle & \multicolumn{3}{c}{Counterfactual world model} \\ \cmidrule(lr){6-8}
Encoder & Environment & Copy & Predictor & (no outcome) & Feature & Readout & Readout, ens. \\ \midrule
MAE-B & HandReach & 7.5 & 38.1 & 49.1 & 82.4 {\scriptsize[80,85]} & 85.6 {\scriptsize[83,87]} & 86.7 \\
MAE-B & Ball-in-Cup & 6.7 & 41.6 & 51.1 & 92.0 {\scriptsize[91,93]} & 94.2 {\scriptsize[93,95]} & 94.7 \\
MAE-B & FetchReach & 43.3 & 44.8 & 50.0 & 94.0 {\scriptsize[93,95]} & 95.7 {\scriptsize[94,97]} & 96.6 \\
MAE-B & Liquid transport & 7.4 & 42.1 & 50.6 & 82.3 {\scriptsize[78,86]} & 94.8 {\scriptsize[92,96]} & 95.9 \\
MAE-B & Push-T$^\dagger$ & 5.1 & 18.3 & 51.0 & 57.3 {\scriptsize[54,61]} & 76.5 {\scriptsize[74,79]} & 79.2 \\
MAE-B & BlueROV2 (8k) & 2.3 & 31.1 & 51.8 & 47.7 {\scriptsize[44,51]} & 73.6 {\scriptsize[71,77]} & 76.9 \\
MAE-B & BlueROV2 (32k) & 2.3 & 43.0 & 51.8 & 65.9 {\scriptsize[64,68]} & 88.2 {\scriptsize[87,89]} & 89.7 \\
MAE-B & BlueROV2 (128k) & 2.3 & 47.4 & 51.8 & 72.2 {\scriptsize[71,74]} & 93.5 {\scriptsize[93,94]} & 94.1 \\
\addlinespace
DINOv2-S & HandReach & 8.5 & 39.9 & 49.1 & 82.3 {\scriptsize[81,84]} & 86.1 {\scriptsize[84,88]} & 87.2 \\
DINOv2-S & Ball-in-Cup & 7.5 & 42.4 & 51.1 & 89.5 {\scriptsize[88,91]} & 91.6 {\scriptsize[90,93]} & 92.4 \\
DINOv2-S & FetchReach & 43.2 & 45.6 & 50.0 & 94.1 {\scriptsize[94,94]} & 95.7 {\scriptsize[95,96]} & 96.3 \\
DINOv2-S & Liquid transport & 7.9 & 43.6 & 50.6 & 81.7 {\scriptsize[78,85]} & 93.8 {\scriptsize[92,95]} & 94.9 \\
DINOv2-S & Push-T$^\dagger$ & 4.9 & 29.9 & 51.0 & 64.5 {\scriptsize[62,67]} & 80.3 {\scriptsize[79,82]} & 82.4 \\
DINOv2-S & BlueROV2 (8k) & 2.8 & 33.5 & 51.8 & 53.0 {\scriptsize[49,57]} & 74.2 {\scriptsize[69,79]} & 76.5 \\
DINOv2-S & BlueROV2 (32k) & 2.8 & 42.5 & 51.8 & 64.2 {\scriptsize[62,67]} & 85.8 {\scriptsize[84,88]} & 86.9 \\
DINOv2-S & BlueROV2 (128k) & 2.8 & 46.4 & 51.8 & 69.1 {\scriptsize[67,71]} & 90.8 {\scriptsize[90,92]} & 91.4 \\
\addlinespace
DINOv2-B & HandReach & 9.0 & 38.1 & 49.1 & 83.3 {\scriptsize[82,85]} & 86.6 {\scriptsize[85,88]} & 87.4 \\
DINOv2-B & Ball-in-Cup & 7.5 & 39.1 & 51.1 & 83.3 {\scriptsize[82,84]} & 87.6 {\scriptsize[85,90]} & 88.4 \\
DINOv2-B & FetchReach & 43.5 & 44.3 & 50.0 & 94.0 {\scriptsize[93,95]} & 95.4 {\scriptsize[94,96]} & 96.6 \\
DINOv2-B & Liquid transport & 8.3 & 42.4 & 50.6 & 73.4 {\scriptsize[70,77]} & 91.1 {\scriptsize[89,93]} & 92.0 \\
DINOv2-B & Push-T$^\dagger$ & 4.2 & 25.3 & 51.0 & 58.3 {\scriptsize[54,62]} & 75.5 {\scriptsize[73,78]} & 77.7 \\
DINOv2-B & BlueROV2 (8k) & 2.7 & 30.7 & 51.8 & 46.1 {\scriptsize[43,50]} & 70.7 {\scriptsize[68,73]} & 73.5 \\
DINOv2-B & BlueROV2 (32k) & 2.7 & 42.1 & 51.8 & 59.9 {\scriptsize[57,63]} & 85.7 {\scriptsize[84,88]} & 86.9 \\
DINOv2-B & BlueROV2 (128k) & 2.7 & 45.7 & 51.8 & 65.9 {\scriptsize[64,68]} & 90.1 {\scriptsize[89,91]} & 90.6 \\
\bottomrule\end{tabular}
\caption{\textbf{Counterfactual prediction in physical coordinates.} Physical counterfactual $R^2$ (\%) on the primary coordinates, one frame, three seeds, $95\%$ whole-parent and seed bootstrap intervals. \emph{Copy}: factual outcome for every query; \emph{Predictor}: the same trained predictor without the outcome; \emph{Oracle (no outcome)}: the best possible prediction without the outcome, from the full simulator state. The counterfactual world model transports the observed residual in feature space and then decodes (\emph{Feature}) or after the readout (\emph{Readout}, \cref{eq:counterfactual-jepa-chart}); \emph{ens.} averages the predictors of the three seeds. Residual reuse with exact simulator means reaches $98.3\%$ on HandReach, $99.8\%$ on Ball-in-Cup and $100.0\%$ on the other four environments. Replacing the observed outcome by that of another disturbance draw of the same source state before correction in feature space (shuffle control) gives $-14.6$ to $6.3\%$ on HandReach, Ball-in-Cup, FetchReach and liquid transport with MAE-B and DINOv2-S. BlueROV2 meets the reported numerical thresholds with the disturbance head fitted on its $32$k-episode collection (\cref{app:bluerov-gate}); as for HandReach, Ball-in-Cup and FetchReach, its Monte Carlo oracle reference has no formal numerical-error bound (\cref{app:independent-benchmark-design}). For liquid transport, the oracle is a Monte Carlo estimate from four disturbance draws per source state; its analytic value is $50.2\%$ (\cref{tab:physical-design-confirmation}). Here (8k), (32k) and (128k) give the number of factual episodes collected for the predictor. $^\dagger$Disturbance recovery from $(z_t,a_t,z_{t+1})$ below the $95\%$ threshold ($93.7$--$94.6\%$); diagnostic.}
\label{tab:main-cf}
\end{table}

\begin{table}[htbp]\centering\small
\setlength{\tabcolsep}{3.5pt}
\begin{tabular}{llrrrllr}\toprule
& & & & & \multicolumn{3}{c}{Counterfactual world model} \\ \cmidrule(lr){6-8}
Encoder & Environment & $\varepsilon$ rec. & Copy & Predictor & Feature & Readout & Readout, ens. \\ \midrule
DINOv3-B & HandReach & 97.2 & 8.4 & 39.2 & 86.4 {\scriptsize[85,88]} & 88.8 {\scriptsize[87,90]} & 90.2 \\
DINOv3-B & Ball-in-Cup & 98.9 & 6.8 & 44.5 & 92.6 {\scriptsize[92,94]} & 95.0 {\scriptsize[94,96]} & 95.4 \\
DINOv3-B & FetchReach & 98.4 & 43.7 & 44.6 & 93.7 {\scriptsize[93,95]} & 94.9 {\scriptsize[94,96]} & 95.8 \\
DINOv3-B & Liquid transport & 98.2 & 7.6 & 42.8 & 82.3 {\scriptsize[76,87]} & 92.8 {\scriptsize[89,95]} & 93.3 \\
DINOv3-B & Push-T$^\dagger$ & 96.7 & 3.9 & 38.3 & 75.0 {\scriptsize[74,76]} & 89.1 {\scriptsize[88,90]} & 90.9 \\
DINOv3-B & BlueROV2 (8k) & 95.9 & 2.7 & 31.3 & 48.2 {\scriptsize[45,51]} & 75.5 {\scriptsize[73,78]} & 78.7 \\
\addlinespace
SigLIP 2-B & HandReach & 96.1 & 9.3 & 37.4 & 79.3 {\scriptsize[75,84]} & 83.6 {\scriptsize[80,87]} & 84.4 \\
SigLIP 2-B & Ball-in-Cup & 98.3 & 7.1 & 42.4 & 84.7 {\scriptsize[84,86]} & 91.3 {\scriptsize[90,92]} & 91.8 \\
SigLIP 2-B & FetchReach & 97.4 & 43.7 & 43.7 & 92.7 {\scriptsize[92,93]} & 95.0 {\scriptsize[94,96]} & 96.0 \\
SigLIP 2-B & Liquid transport & 97.4 & 7.9 & 42.7 & 75.8 {\scriptsize[72,79]} & 91.7 {\scriptsize[89,94]} & 92.2 \\
SigLIP 2-B & Push-T$^\dagger$ & 95.4 & 4.8 & 31.6 & 63.4 {\scriptsize[61,66]} & 81.2 {\scriptsize[80,83]} & 83.5 \\
SigLIP 2-B & BlueROV2 (8k)$^\dagger$ & 94.3 & 2.2 & 24.1 & 39.1 {\scriptsize[32,46]} & 66.6 {\scriptsize[59,73]} & 70.4 \\
\addlinespace
CLIP-B & HandReach & 95.3 & 9.2 & 36.4 & 79.3 {\scriptsize[77,81]} & 82.4 {\scriptsize[80,84]} & 82.9 \\
CLIP-B & Ball-in-Cup & 96.2 & 7.5 & 39.6 & 77.3 {\scriptsize[76,79]} & 86.0 {\scriptsize[85,87]} & 86.8 \\
CLIP-B & FetchReach & 97.4 & 43.7 & 45.4 & 93.5 {\scriptsize[93,94]} & 95.3 {\scriptsize[95,96]} & 96.1 \\
CLIP-B & Liquid transport$^\dagger$ & 94.4 & 7.6 & 39.4 & 69.1 {\scriptsize[65,73]} & 88.9 {\scriptsize[85,92]} & 90.8 \\
CLIP-B & Push-T$^\dagger$ & 93.2 & 4.2 & 28.1 & 56.1 {\scriptsize[52,60]} & 76.0 {\scriptsize[72,80]} & 77.8 \\
CLIP-B & BlueROV2 (8k)$^\dagger$ & 93.1 & 3.2 & 23.3 & 35.7 {\scriptsize[29,42]} & 62.1 {\scriptsize[56,68]} & 65.9 \\
\addlinespace
EVA-02-B & HandReach & 95.6 & 9.9 & 34.7 & 77.9 {\scriptsize[75,80]} & 81.0 {\scriptsize[78,84]} & 81.6 \\
EVA-02-B & Ball-in-Cup & 97.3 & 7.2 & 40.3 & 81.6 {\scriptsize[81,83]} & 88.2 {\scriptsize[87,90]} & 89.1 \\
EVA-02-B & FetchReach & 97.1 & 43.6 & 44.5 & 93.0 {\scriptsize[92,94]} & 95.3 {\scriptsize[95,96]} & 96.1 \\
EVA-02-B & Liquid transport & 97.2 & 7.4 & 40.9 & 71.0 {\scriptsize[67,75]} & 90.9 {\scriptsize[88,93]} & 91.9 \\
EVA-02-B & Push-T$^\dagger$ & 94.0 & 4.7 & 19.1 & 53.9 {\scriptsize[51,57]} & 71.2 {\scriptsize[68,74]} & 74.1 \\
EVA-02-B & BlueROV2 (8k)$^\dagger$ & 92.6 & 2.5 & 18.5 & 34.0 {\scriptsize[30,38]} & 59.7 {\scriptsize[56,63]} & 63.3 \\
\bottomrule\end{tabular}
\caption{\textbf{Four further frozen encoders.} Same environments, factual data, predictor recipe, readout recipe and evaluator as \cref{tab:main-cf}; only the encoder changes. Physical counterfactual $R^2$ (\%) on the primary coordinates, one frame, three seeds, $95\%$ whole-parent and seed bootstrap intervals. \emph{$\varepsilon$ rec.}: held-out disturbance recovery from frozen features on the evaluation cohort; for BlueROV2, on independent confirmation with the disturbance head fitted on its $32$k-episode collection, as in \cref{app:bluerov-gate}; DINOv3-B passes there on the point estimate, with lower confirmation bound $94.9\%$. Oracle (no outcome) references and residual reuse with exact simulator means are those of \cref{tab:main-cf}. $^\dagger$Diagnostic: Push-T as in \cref{tab:main-cf}, or disturbance recovery below the $95\%$ threshold.}
\label{tab:encoder-breadth}
\end{table}

\begin{table}[htbp]\centering\small
\begin{tabular}{lrrrrrrr}\toprule
& \multicolumn{2}{c}{Predictor} & \multicolumn{2}{c}{CF WM, feature} & \multicolumn{2}{c}{CF WM, readout} & Factual \\ \cmidrule(lr){2-3}\cmidrule(lr){4-5}\cmidrule(lr){6-7}
Environment & orig. & sel. & orig. & sel. & orig. & sel. & MSE gain \\ \midrule
HandReach & 27.5 & 38.1 & 77.3 & 82.4 & 78.6 & 85.6 & 12.4 \\
Ball-in-Cup & 42.0 & 41.6 & 93.1 & 92.0 & 94.4 & 94.2 & 10.4 \\
FetchReach & 41.0 & 44.8 & 91.9 & 94.0 & 92.6 & 95.7 & 10.4 \\
Liquid transport & 38.0 & 42.1 & 78.7 & 82.3 & 91.2 & 94.8 & 8.4 \\
Push-T$^\dagger$ & 30.4 & 18.3 & 70.9 & 57.3 & 83.8 & 76.5 & 8.4 \\
BlueROV2 (8k) & 21.7 & 31.1 & 42.7 & 47.7 & 64.7 & 73.6 & 17.3 \\
\bottomrule\end{tabular}
\caption{\textbf{Factual predictor selection.} MAE-B, one frame, three seeds; physical counterfactual $R^2$ (\%). \emph{orig.}: the original predictor recipe, whose one-frame checkpoints in this table were all selected at the first validation (step 500); \emph{sel.}: the recipe selected by held-out factual error alone (\cref{app:stress}). \emph{Factual MSE gain}: relative reduction of held-out factual mean squared error (MSE) in feature units. No counterfactual quantity enters the selection.}
\label{tab:wm-quality}
\end{table}

\begin{table}[htbp]\centering\small
\begin{tabular}{lrrrrrrrr}\toprule
& \multicolumn{4}{c}{CF WM, readout} & \multicolumn{4}{c}{Trained predictor} \\ \cmidrule(lr){2-5}\cmidrule(lr){6-9}
Fitting episodes & 819 & 1{,}638 & 3{,}276 & 6{,}553 & 819 & 1{,}638 & 3{,}276 & 6{,}553 \\ \midrule
HandReach & 69.8 & 71.0 & 80.9 & 85.6 & 22.6 & 29.7 & 31.5 & 38.1 \\
Ball-in-Cup & 80.8 & 80.4 & 92.6 & 94.2 & 23.7 & 26.9 & 40.0 & 41.6 \\
FetchReach & 90.7 & 90.6 & 93.3 & 95.7 & 33.6 & 35.5 & 41.2 & 44.8 \\
Liquid transport & 83.9 & 89.5 & 94.5 & 94.8 & 25.2 & 37.6 & 42.3 & 42.1 \\
\bottomrule\end{tabular}
\\[6pt]\begin{tabular}{llrrrrr}\toprule
\multicolumn{2}{l}{BlueROV2 fitting episodes} & 6{,}553 & 13{,}107 & 26{,}214 & 52{,}428 & 104{,}857 \\ \midrule
MAE-B & CF WM, readout & 73.7 & 83.6 & 88.2 & 91.5 & 93.5 \\
 & Trained predictor & 31.8 & 39.1 & 43.0 & 45.6 & 47.4 \\
DINOv2-S & CF WM, readout & 74.6 & 83.1 & 85.8 & 88.6 & 90.8 \\
 & Trained predictor & 35.1 & 39.8 & 42.5 & 44.4 & 46.4 \\
DINOv2-B & CF WM, readout & 71.1 & 81.3 & 85.7 & 88.5 & 90.1 \\
 & Trained predictor & 32.3 & 39.3 & 42.1 & 44.5 & 45.7 \\
\bottomrule\end{tabular}
\caption{\textbf{Factual-data scaling.} Selected recipe, one frame, three seeds; physical counterfactual $R^2$ (\%). Only the number of factual fitting episodes changes; validation episodes, evaluation population and readouts are fixed. Top: MAE-B. Bottom: BlueROV2 with all three encoders, from its $32$k-episode (up to $26{,}214$ fitting episodes) and $128$k-episode collections, whose reset seeds are nested; the oracle predictor (no outcome) reaches $51.8\%$ at every size, and the $8$k-collection results are in \cref{tab:main-cf}.}
\label{tab:data-size}
\end{table}

\begin{table}[htbp]\centering\footnotesize
\setlength{\tabcolsep}{3pt}
\begin{tabular}{lllrrrrrl}\toprule
Environment & Disturbance & Enters & $h$ (s) & Oracle & Copy & Ideal resid. & $\rho$ & Gates \\ \midrule
HandReach & actuator offset (benchmark) & command & 0.16 & 49.1 & 7.3 & 98.3 & 0.033 & pass \\
HandReach & external load on the palm & plant, force & 0.16 & 50.1 & 0.4 & 100.0 & 0.001 & pass \\
HandReach & joint damping $\times(1\pm0.9)$ & plant, state & 0.04 & 94.4 & 42.3 & 95.9 & 0.736 & fail \\
HandReach & joint damping $\times(1\pm0.9)$ & plant, state & 0.16 & 80.1 & $-$69.4 & 82.9 & 0.860 & fail \\
Ball-in-Cup & joint damping $\times(1\pm0.9)$ & plant, state & 0.08 & 92.7 & 38.3 & 92.1 & 1.081 & fail \\
Ball-in-Cup & joint damping $\times(1\pm0.9)$ & plant, state & 0.32 & 91.5 & $-$85.9 & 93.6 & 0.754 & fail \\
Ball-in-Cup & wind on the ball & plant, force & 0.32 & 96.0 & $-$58.2 & 99.7 & 0.076 & fail \\
\bottomrule\end{tabular}
\caption{\textbf{External-load and damping-fault screen.} Simulator-only screen on the primary actuated coordinates (HandReach wrist, Ball-in-Cup cup position), pooled over the two coordinates; $R^2$ in percent. One configuration switch selects the mechanism; actions, queries, resets and coordinates are those of the benchmark. \emph{Enters}: through the command, or through the plant as a force or as a coefficient that multiplies the state. \emph{Oracle}: oracle predictor (no outcome). \emph{Gates}: both per-coordinate oracle references within the $50\pm8\%$ design target and both copy references below $60\%$. The external load was calibrated on this development cohort before any model; a damping coefficient varied by up to $\pm90\%$ leaves the oracle reference at $80$--$94\%$. Cohorts of 128 parents $\times$ 32 draws (damping, wind), 256 $\times$ 64 (load) or 128 $\times$ 128 (offset); all native control and replay tests pass. No image, readout or fitted model is used.}
\label{tab:plant-screen}
\end{table}

\begin{table}[htbp]\centering\small
\setlength{\tabcolsep}{3.5pt}
\begin{tabular}{llrrrrllr}\toprule
Encoder & Disturbance & $\varepsilon$ rec. & Oracle & Copy & Predictor & CF WM, feature & CF WM, readout & Ideal resid. \\ \midrule
MAE-B & actuator offset & 97.1 & 49.1 & 7.5 & 38.1 & 82.4 {\scriptsize[80,85]} & 85.6 {\scriptsize[83,87]} & 98.3 \\
MAE-B & external load & 98.1 & 50.0 & 13.8 & 39.0 & 84.4 {\scriptsize[80,87]} & 88.7 {\scriptsize[85,91]} & 99.9 \\
\addlinespace
DINOv2-S & actuator offset & 96.6 & 49.1 & 8.5 & 39.9 & 82.3 {\scriptsize[81,84]} & 86.1 {\scriptsize[84,88]} & 98.3 \\
DINOv2-S & external load & 97.2 & 50.0 & 13.4 & 39.8 & 80.9 {\scriptsize[77,84]} & 86.9 {\scriptsize[83,90]} & 99.9 \\
\bottomrule\end{tabular}
\caption{\textbf{Prediction under external loads.} HandReach with the external load at the palm in place of the actuator offset; new factual data, frozen readouts and three factual-selected predictors, with the same pipeline, encoders and evaluator. Physical counterfactual $R^2$ (\%) on the wrist with $95\%$ whole-parent and seed bootstrap intervals; \emph{$\varepsilon$ rec.}: held-out disturbance recovery from frozen features on the evaluation cohort. \emph{Oracle}: the best possible prediction without the outcome, from the full simulator state. \emph{Ideal resid.}: residual reuse with simulator conditional means; a reference for this rule, not a universal upper bound. Oracle, copy and ideal-residual values come from the evaluation cohort and differ from the development screen of \cref{tab:plant-screen}.}
\label{tab:plant-learned}
\end{table}

\begin{table}[htbp]\centering\small
\setlength{\tabcolsep}{4pt}
\begin{tabular}{lrrrrllr}\toprule
Encoder & $\lambda$ & $\varepsilon$ rec. & Oracle & Predictor & CF WM, feature & CF WM, readout & Ideal resid. \\ \midrule
MAE-B & 0 & 97.1 & 49.1 & 38.1 & 82.4 {\scriptsize[80,85]} & 85.6 {\scriptsize[83,87]} & 98.3 \\
MAE-B & 0.5 & 91.0 & 52.9 & 41.0 & 70.5 {\scriptsize[67,73]} & 67.4 {\scriptsize[64,70]} & 78.9 \\
MAE-B & 1 & 62.9 & 52.4 & 41.4 & 6.4 {\scriptsize[$-$1,15]} & $-$11.8 {\scriptsize[$-$22,$-$1]} & $-$1.1 \\
\addlinespace
DINOv2-S & 0 & 96.6 & 49.1 & 39.9 & 82.3 {\scriptsize[81,84]} & 86.1 {\scriptsize[84,88]} & 98.3 \\
DINOv2-S & 0.5 & 91.5 & 52.9 & 43.3 & 70.6 {\scriptsize[67,74]} & 66.3 {\scriptsize[63,69]} & 78.9 \\
DINOv2-S & 1 & 66.7 & 52.4 & 41.6 & 11.0 {\scriptsize[2,19]} & $-$12.8 {\scriptsize[$-$23,$-$2]} & $-$1.1 \\
\bottomrule\end{tabular}
\caption{\textbf{Learned predictions across the interaction dial.} The disturbance moves from an actuator offset ($\lambda=0$) to a pure gain fault ($\lambda=1$); each condition has its own factual data, frozen readouts and factual-selected predictors (three seeds). Physical counterfactual $R^2$ (\%) on the wrist; \emph{$\varepsilon$ rec.}: held-out disturbance recovery from frozen features on the evaluation cohort, below the $95\%$ threshold for $\lambda>0$ (diagnostic). \emph{Oracle}: the best possible prediction without the outcome, from the full simulator state. \emph{Ideal resid.}: residual reuse with simulator conditional means, not a universal upper bound. Oracle and ideal-residual values come from the evaluation cohorts and differ slightly from the development screen of \cref{tab:dial}.}
\label{tab:dial-learned}
\end{table}

\clearpage

\clearpage

\section{Evaluation Protocol}
\label{app:pinned-protocol}
\label{sec:pinned-continuing}
\paragraph{Qualification.}\label{app:benchmark-selection-summary}
Each condition fixes the physics, query population, scored coordinates and
readouts before fitting. Qualification requires held-out disturbance and
retained-coordinate recovery of at least $95\%$, and oracle (no outcome) and
factual-outcome-copy $R^2$ strictly below $60\%$ on the same population;
simulation also checks true-state factual copying. A condition is qualified
only when every criterion has been measured and passes; an unmeasured criterion
leaves it unqualified. We call each criterion a \emph{gate}; conditions that
fail a gate are reported as diagnostics. Recovery is checked on held-out
factual episodes and on an independent confirmation cohort of new source states
collected after the physical setting was selected. A \emph{parent} is a source
state from which the factual and all query branches are simulated with the
same disturbance draw; the scored source is the $t=2$ anchor, reached after
two behavior-policy transitions. The numerical checks of the physical
references are reported separately in \cref{app:independent-benchmark-design}
(\cref{tab:physical-design-confirmation}).

\subsection{No-Outcome Information Sets}
Let $H_t^{(k)}=(z_{t-k+1},a_{t-k+1},\ldots,a_{t-1},z_t)$ denote a
causal $k$-frame observation history. Past commands are included, the current
command is replaced by the query $a'$, and the current endpoint $z_{t+1}$ is
absent. Write $Y=S_{t+1}^{\star}(a')$ for the retained physical
counterfactual coordinates and $I_k=(H_t^{(k)},a')$.
The ideal no-outcome predictor, the oracle (no outcome) given $I_k$, and its risk are
\begin{align}
 b_k^*(I_k)&=\E[Y\mid I_k],\\
 R_{\mathrm{Bayes},k}^2
 &=1-\frac{\E\operatorname{tr}\operatorname{Var}(Y\mid I_k)}
                 {\E\|Y-\E Y\|^2}.\label{eq:history-bayes-risk}
\end{align}
This is a population squared-error ceiling for predictors with information
$I_k$; it is neither a ceiling for a predictor that observes the current
factual endpoint nor a universal constant independent of history or
population. The fully fresh simulator reference used in the main text conditions on the full
physical source and query action, a stronger information set than a single
image, and estimates conditional noise variance from independent same-source
simulator branches. All branching is
evaluation-only; the plotted reference includes Monte Carlo uncertainty.

The practical baseline is a fitted embedding predictor,
\begin{equation}
 \widehat z_{t+1}^{\mathrm{det}}=\widehat\mu_k(H_t^{(k)},a'),\qquad
 \widehat Y_{\mathrm{det}}=\pi(\widehat z_{t+1}^{\mathrm{det}}),
 \label{eq:practical-history-mean}
\end{equation}
where $\pi$ is the frozen nonlinear physical readout. Factual squared loss has
the population target $\E[z_{t+1}\mid H_t^{(k)},A_t=a]$, which sufficient data,
capacity and successful optimization can approximate. They do not generally
make $\pi(\E[z_{t+1}\mid I_k])$ equal
$\E[\pi(z_{t+1})\mid I_k]$, and cannot supply physical variables absent from
the observation. The
practical decoded predictor need not reach the physical oracle even in an
idealized infinite-data limit, so the oracle and learned bars must remain
separate.

\paragraph{History and counterfactual risk.}
\begin{proposition}[History helps prediction; exact additive cancellation is already complete]
\label{prop:history-risk-order}
On a fixed population, let $\mathcal I_1\subseteq\mathcal I_2$ be nested
no-current-outcome information sets. Their optimal squared risks satisfy
\begin{equation}
 \mathcal R_1^*-\mathcal R_2^*
 =\E\|\E[Y\mid\mathcal I_2]-\E[Y\mid\mathcal I_1]\|^2\ge0.
 \label{eq:history-projection-identity}
\end{equation}
If the factual means identify the potential-outcome means given each history,
the population residual risk is
$\mathcal R_{R,k}=\E\operatorname{tr}\operatorname{Var}(\Delta\mid H^{(k)})$ and
decreases under nested histories by
$\E\|\E[\Delta\mid H^{(2)}]-\E[\Delta\mid H^{(1)}]\|^2$. If the action contrast
is determined by the shorter history, both residual risks are zero: history
has no error left to remove.
(Proof in \cref{proof:history-risk-order}.)
\end{proposition}

\paragraph{Persistent disturbances.}
For the Gaussian driver in \cref{eq:pinned-driver}, write
$g_t=\sqrt{1-\alpha}\,b_t+\sqrt\alpha\,\eta_t$ with $b_t=b(t/\ell)^\top\xi$ and a
unit-variance IID innovation $\eta_t$ independent of pre-transition
information. Then
\begin{equation}
 \operatorname{Var}(g_t\mid H_t^{(k)})
 =(1-\alpha)\operatorname{Var}(b_t\mid H_t^{(k)})+\alpha.
 \label{eq:history-iid-variance}
\end{equation}
This identity concerns driver variance, not physical prediction error after
nonlinear dynamics. Even at $\alpha=1$, history may help recover hidden state.
The smooth-noise Monte Carlo reference conditions on the native reset, past
commands and realized past drivers, and integrates the current driver under
its conditional Gaussian law. It excludes the current outcome and driver.
Its privileged past contains each compared image window, making its
$R^2$ an upper bound on the image-window optimum, so a value above $60\%$
does not by itself show that the best image-history predictor exceeds that
threshold. A frozen image predictor
provides a lower bound for its actual information set. Pointwise bounds are
not combined into a joint confidence interval.

\subsection{Encoder Specifications}
\label{app:encoders}
The counterfactual experiments use frozen DINOv2-S and DINOv2-B
\citep{oquab2023dinov2} and MAE-B \citep{he2022masked}, retaining their CLS
tokens (384, 768 and 768 coordinates, respectively). All receive bilinearly
resized full-frame $224\times224$ images, without center cropping, followed
by ImageNet channel normalization. This explicitly overrides the DINOv2
checkpoint's default input size. No physical coordinate is concatenated with
the features. The encoder comparison of \cref{tab:encoder-breadth} adds
DINOv3-B, SigLIP~2-B, CLIP-B (trained on LAION-2B) and EVA-02-B, through the
timm checkpoints \path{vit_base_patch16_dinov3.lvd1689m},
\path{vit_base_patch16_siglip_224.v2_webli},
\path{vit_base_patch16_clip_224.laion2b} and
\path{eva02_base_patch14_224.mim_in22k}, each with 768 coordinates: the mean
patch token for DINOv3-B, EVA-02-B and SigLIP~2-B, whose attention-pooling head
is not used, and the CLS token for CLIP-B. They receive the same resized frames,
each with its own channel normalization. The code release records model identifiers, checkpoint hashes,
normalization and pooling, and checks them at extraction and inference.

Feature displacement is scaled by
\[
 r=\max\left\{10^{-4},\left[d_z^{-1}\sum_j
 \widehat{\operatorname{Var}}_{\rm fit}(z_{t+1,j}-z_{t,j})\right]^{1/2}\right\}.
\]
Conditioning means and standard deviations use fitting episodes only, with
a $10^{-4}$ standard-deviation floor. A scalar output scale preserves the
relative feature weights in the squared loss. Predictions are returned to
the original feature coordinates before decoding or residual correction.

\subsection{Factual Fitting and Evaluation}
\label{app:factual-fitting}
There are 8,192 independent factual episodes per environment, each with three
transitions. Whole-episode splitting gives 6,553 fitting, 820 validation and
819 reserved episodes per seed. The selected recipe uses all recorded fitting
transitions, two width-1,024 attention blocks, four heads per block and multilayer perceptron (MLP)
ratio one. History lengths $k=1,2,3$ share capacity and initialization policy.
Fits use seeds $0,1,2$, 20,000 updates, batch size 256, AdamW learning rate
$3\times10^{-4}$, weight decay $0.3$ and cosine scheduling. Checkpoints
minimize factual validation loss at the $t=2$ anchor, checked every 100
updates. The selection grid is specified in \cref{app:stress}.
No counterfactual outcome selects a seed, encoder or checkpoint.

Frozen DINOv2-S, DINOv2-B and MAE-B encode each frame separately
(\cref{app:encoders}); the predictor receives no physical-state labels.
The state and disturbance heads have three hidden layers of width 1,024,
30,000 updates, AdamW learning rate $3\times10^{-4}$, weight decay $10^{-4}$,
and factual-validation checkpoint selection. Physical labels supervise these
evaluation heads only.

The predictor outputs a normalized
feature displacement $d_\theta$ and minimizes the factual loss of
\eqref{eq:factual-loss},
\begin{equation}
 \mathcal L_{\mathrm{factual}}
 =\E_{\mathrm{train}}\left\|
 d_\theta(H_t^{(k)},a_t)-\frac{z_{t+1}-z_t}{r}\right\|^2,
 \qquad \widehat\mu_k(H,a)=z_t+r\,d_\theta(H,a),
\end{equation}
where $r$ is the fitting-only root mean square of the feature-displacement standard
deviations; input state/action standardization is also fitted only on training
episodes. This differs from physical-state supervision by a frozen head: no
physical target is used in this loss. Residual inference adds the observed
endpoint to the difference of two queries of the same fitted predictor, as in
\eqref{eq:history-residual-rule}, and requires no second training stage. Both
arms therefore share the encoder, predictor weights, factual training budget
and checkpoint: residual reuse changes only inference and adds no learned
parameters. Same-disturbance alternative outcomes enter exclusively in
evaluation; the supervised diagnostic heads neither train nor select the
factual predictor. All training, validation and checkpoint selection use the
stable-pretraining library\footnote{\url{https://github.com/rbalestr-lab/stable-pretraining},
revision \texttt{9aa93f8b6153}.}.
\paragraph{Coordinates and recovery.}The masks are two wrist angles (HandReach), cup $x,z$
(Ball-in-Cup), end-effector $x,y,z$ (FetchReach), heading and lane offset
(Racetrack), $\sin(\mathrm{pitch})$ (BlueROV2), glass $x$ (PassWater), and
pusher $x,y$ (Push-T). These are selected actuator or pose coordinates, not
full task state: they do not establish finger, ball, fluid-particle or
block-contact counterfactual accuracy. Every retained coordinate must exceed
the image recovery threshold; no coordinate is dropped after inspecting CF
performance. HandReach, Ball-in-Cup, FetchReach, PassWater and BlueROV2 (with
the disturbance readout of \cref{app:bluerov-gate}) pass both held-factual and
confirmation image gates with all three encoders. With DINOv2-B on HandReach,
Ball-in-Cup, FetchReach and PassWater, disturbance recovery is
$96.0$--$98.3\%$ held-out and $96.4$--$98.4\%$ on confirmation, state recovery
$98.6$--$99.9\%$ and $98.4$--$99.8\%$, and the upper $95\%$ bounds of the
oracle predictor (no outcome) and factual copying are at most $55.5$ and
$53.2\%$. Racetrack fails heading and disturbance recovery; Push-T fails
disturbance recovery. A failed fitted recovery head
does not prove that the representation contains no such information. State
heads read one frame, while the shared disturbance head reads
$(z_t,a_t,z_{t+1})$ at every history length; its failure blocks the
common-screen comparison but does not establish that a separately fitted
history-aware disturbance head would also fail. Disturbance recovery is a
conservative benchmark eligibility requirement, not a necessary condition of
the residual-cancellation theorem.

\paragraph{Numerical checks and uncertainty.}PassWater's native confirmation has only 128 parents with four disturbance
draws each, below the declared minimum of 64 draws; its intervals are
therefore especially limited. The other
scored native cohorts use 128 parents $\times$ 128 draws (HandReach), 512
$\times$ 128 (Ball-in-Cup), and 256 $\times$ 128 (FetchReach), with three
unequal query actions per parent. All rows and all predeclared seeds are
retained. The shuffle control of \cref{sec:pinned-results} replaces the
observed outcome by that of another disturbance draw of the same parent before
applying the correction. Reported aggregate physical $R^2$ uses the summed squared errors and
summed target variances of the fixed coordinates. Pointwise intervals use 512
paired whole-parent and fitting-seed bootstrap draws; the point estimate is
average single-fit risk, not the risk of an ensemble. These intervals
condition on the fixed readouts and collected training data; three fitting
seeds do not characterize every source of pipeline variability. The oracle
predictor (no outcome) is a Monte Carlo reference conditioned on the full
current simulator state and query action, without the current outcome or
disturbance; fresh disturbances make its conditional law independent of the
past. It uses more information than the image predictor, it is not the optimum
over image histories, and its intervals do not include numerical integration
error.

\subsection{History and Fresh-Share Ablations}
\label{sec:pinned-ablations}
\begin{figure}[tbp]\centering
\includegraphics[width=.92\linewidth]{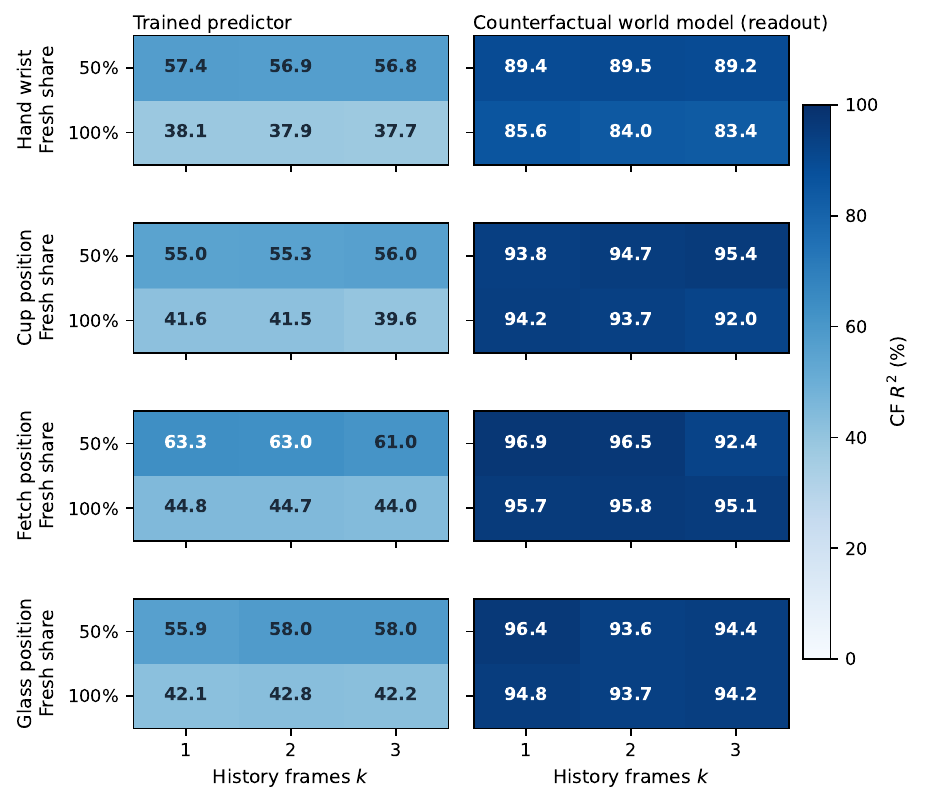}
\caption{\textbf{History and fresh-share ablations, MAE-B.} Trained predictor ({\bf left}) and counterfactual world model with
readout transport ({\bf right}) of the same factual-selected predictor;
physical counterfactual $R^2$ in percent, three seeds per cell. The $100\%$
row at $k=1$ is the benchmark of \cref{tab:main-cf}; physics is not retuned
between rows. Intervals, feature transport, copy and oracle (no outcome)
references are in \cref{tab:pinned-half-iid-mae-cls,tab:main-cf}.}
\label{fig:pinned-iid-history-mae}
\end{figure}

\begin{figure}[tbp]\centering
\includegraphics[width=.92\linewidth]{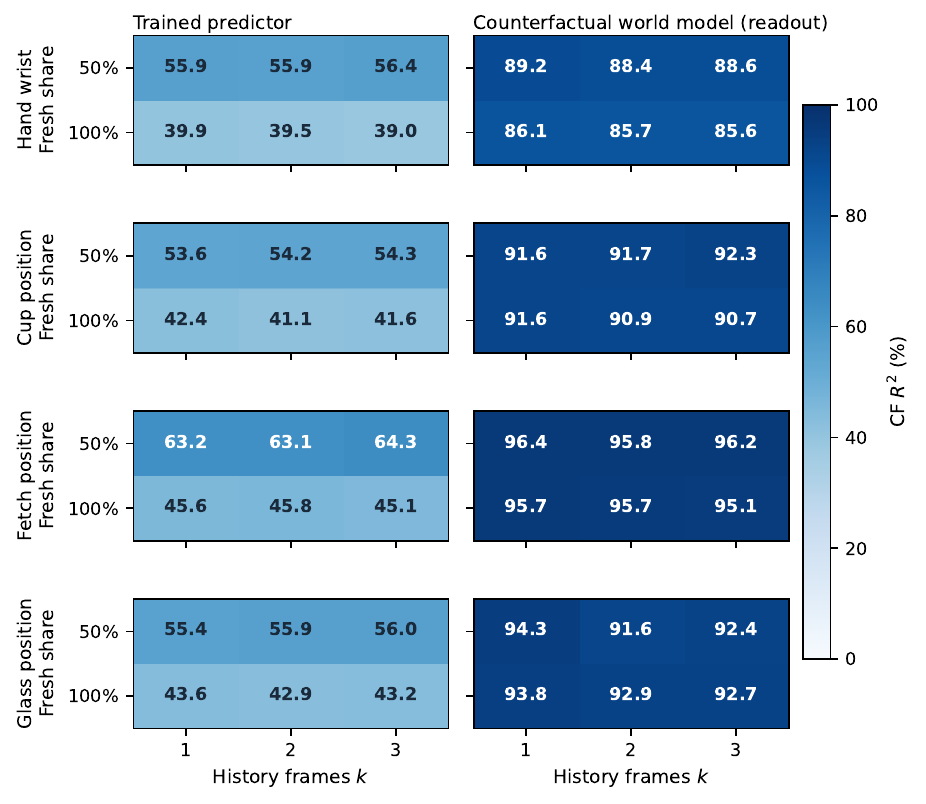}
\caption{\textbf{History and fresh-share ablations, DINOv2-S.} Same
sampling and fitting protocol as \cref{fig:pinned-iid-history-mae}, with the
frozen DINOv2-S readouts; all cells have three seeds. Intervals for the $50\%$
rows are in \cref{tab:pinned-half-iid-dinov2s-cls}; the $100\%$ row at $k=1$ is
the benchmark of \cref{tab:main-cf}.}
\label{fig:pinned-iid-history-dino}
\end{figure}

The matched MAE-B and DINOv2-S grids
(\cref{fig:pinned-iid-history-mae,fig:pinned-iid-history-dino}) use fresh
shares $\alpha=.5$ and $\alpha=1$ (the benchmark), with temporal scale $\ell=4$
(\cref{eq:pinned-driver}), on HandReach, Ball-in-Cup, FetchReach and PassWater.
HandReach/MAE-B additionally uses $\alpha=.25$ and $.75$
(\cref{tab:pinned-hand-iid}). Each history/seed cell
fits a predictor with the selected recipe above. Gains, durations, policy,
query law, coordinate masks and frozen readouts stay fixed; readout recovery
is rechecked on the new factual population. The same $t=2$ anchors and queries
are used across history lengths within each disturbance setting.
Across fresh shares, the visited-state population and target variance change.
These are ablations of the data-generating process, not same-state effects of
persistence alone. Cells that exceed the no-outcome threshold remain labeled
diagnostics; their physics is not recalibrated.

\begin{table}[tbp]\centering\small
\setlength{\tabcolsep}{4pt}
\begin{tabular}{lrlllrr}\toprule
Environment & $k$ & Trained predictor & CF WM, feature & CF WM, readout & Copy & Oracle \\ \midrule
Hand wrist & 1 & 57.4 {\scriptsize[53.0, 61.1]} & 88.3 {\scriptsize[86.6, 89.8]} & 89.4 {\scriptsize[87.5, 90.9]} & 18.9 & 67.8 \\
 & 2 & 56.9 {\scriptsize[53.0, 60.3]} & 88.3 {\scriptsize[86.7, 89.6]} & 89.5 {\scriptsize[87.7, 91.2]} & 18.9 & 67.8 \\
 & 3 & 56.8 {\scriptsize[52.7, 60.4]} & 88.1 {\scriptsize[86.2, 89.5]} & 89.2 {\scriptsize[87.3, 90.9]} & 18.9 & 67.8 \\
Cup position & 1 & 55.0 {\scriptsize[52.0, 57.4]} & 92.1 {\scriptsize[91.1, 93.0]} & 93.8 {\scriptsize[92.6, 94.8]} & 12.5 & 66.4 \\
 & 2 & 55.3 {\scriptsize[51.4, 58.3]} & 92.8 {\scriptsize[91.3, 94.0]} & 94.7 {\scriptsize[93.8, 96.0]} & 12.5 & 66.4 \\
 & 3 & 56.0 {\scriptsize[53.5, 58.1]} & 94.2 {\scriptsize[93.4, 95.0]} & 95.4 {\scriptsize[94.8, 96.0]} & 12.5 & 66.4 \\
Fetch position & 1 & 63.3 {\scriptsize[60.5, 65.7]} & 95.7 {\scriptsize[95.1, 96.3]} & 96.9 {\scriptsize[96.4, 97.4]} & 54.3 & 70.3 \\
 & 2 & 63.0 {\scriptsize[60.2, 65.4]} & 95.5 {\scriptsize[95.0, 95.9]} & 96.5 {\scriptsize[96.1, 96.9]} & 54.3 & 70.3 \\
 & 3 & 61.0 {\scriptsize[57.3, 63.8]} & 91.9 {\scriptsize[89.9, 93.9]} & 92.4 {\scriptsize[89.9, 94.7]} & 54.3 & 70.3 \\
Glass position & 1 & 55.9 {\scriptsize[48.9, 61.4]} & 88.3 {\scriptsize[85.9, 90.3]} & 96.4 {\scriptsize[95.3, 97.4]} & 11.3 & 65.0 \\
 & 2 & 58.0 {\scriptsize[51.9, 63.1]} & 85.9 {\scriptsize[82.8, 88.4]} & 93.6 {\scriptsize[90.7, 95.8]} & 11.3 & 65.0 \\
 & 3 & 58.0 {\scriptsize[51.5, 63.3]} & 86.4 {\scriptsize[83.6, 88.7]} & 94.4 {\scriptsize[92.0, 96.1]} & 11.3 & 65.0 \\
\bottomrule\end{tabular}
\caption{\textbf{50\% fresh-share ablation, MAE-B.} Physical counterfactual $R^2$ in percent with paired seed/parent $95\%$ intervals; three factual-selected predictors per row (all-anchor training, width 1024, weight decay 0.3; selection on held factual error only), with the frozen readouts, physics and query law of the benchmark. \emph{Oracle}: the parent-conditional Monte Carlo reference. With a persistent component, it uses privileged past information and is an upper bound on the best $R^2$ achievable from the image window, not that window's exact optimum. These rows remain diagnostics.}
\label{tab:pinned-half-iid-mae-cls}
\end{table}

\begin{table}[tbp]\centering\small
\setlength{\tabcolsep}{4pt}
\begin{tabular}{lrlllrr}\toprule
Environment & $k$ & Trained predictor & CF WM, feature & CF WM, readout & Copy & Oracle \\ \midrule
Hand wrist & 1 & 55.9 {\scriptsize[51.6, 60.0]} & 86.6 {\scriptsize[85.0, 88.1]} & 89.2 {\scriptsize[87.9, 90.4]} & 19.2 & 67.8 \\
 & 2 & 55.9 {\scriptsize[51.7, 59.7]} & 86.1 {\scriptsize[84.6, 87.7]} & 88.4 {\scriptsize[87.1, 89.7]} & 19.2 & 67.8 \\
 & 3 & 56.4 {\scriptsize[51.8, 60.2]} & 86.2 {\scriptsize[84.5, 87.7]} & 88.6 {\scriptsize[87.3, 90.0]} & 19.2 & 67.8 \\
Cup position & 1 & 53.6 {\scriptsize[51.3, 55.3]} & 90.3 {\scriptsize[89.3, 91.1]} & 91.6 {\scriptsize[90.2, 93.0]} & 13.1 & 66.4 \\
 & 2 & 54.2 {\scriptsize[51.8, 56.1]} & 91.1 {\scriptsize[90.2, 91.7]} & 91.7 {\scriptsize[90.6, 92.8]} & 13.1 & 66.4 \\
 & 3 & 54.3 {\scriptsize[51.7, 56.3]} & 91.2 {\scriptsize[90.7, 91.6]} & 92.3 {\scriptsize[91.5, 92.9]} & 13.1 & 66.4 \\
Fetch position & 1 & 63.2 {\scriptsize[60.3, 65.6]} & 95.4 {\scriptsize[94.9, 95.8]} & 96.4 {\scriptsize[96.0, 96.9]} & 54.5 & 70.3 \\
 & 2 & 63.1 {\scriptsize[60.1, 65.3]} & 94.8 {\scriptsize[94.0, 95.5]} & 95.8 {\scriptsize[95.1, 96.6]} & 54.5 & 70.3 \\
 & 3 & 64.3 {\scriptsize[61.5, 66.6]} & 95.3 {\scriptsize[94.7, 95.8]} & 96.2 {\scriptsize[95.7, 96.7]} & 54.5 & 70.3 \\
Glass position & 1 & 55.4 {\scriptsize[48.8, 61.0]} & 84.8 {\scriptsize[81.6, 87.5]} & 94.3 {\scriptsize[91.9, 95.9]} & 12.3 & 65.0 \\
 & 2 & 55.9 {\scriptsize[49.4, 61.6]} & 81.9 {\scriptsize[78.2, 85.1]} & 91.6 {\scriptsize[89.0, 93.7]} & 12.3 & 65.0 \\
 & 3 & 56.0 {\scriptsize[49.4, 61.5]} & 82.2 {\scriptsize[78.6, 85.2]} & 92.4 {\scriptsize[90.2, 94.1]} & 12.3 & 65.0 \\
\bottomrule\end{tabular}
\caption{\textbf{50\% fresh-share ablation, DINOv2-S.} Physical counterfactual $R^2$ in percent with paired seed/parent $95\%$ intervals; three factual-selected predictors per row (all-anchor training, width 1024, weight decay 0.3; selection on held factual error only), with the frozen readouts, physics and query law of the benchmark. \emph{Oracle}: the parent-conditional Monte Carlo reference. With a persistent component, it uses privileged past information and is an upper bound on the best $R^2$ achievable from the image window, not that window's exact optimum. These rows remain diagnostics.}
\label{tab:pinned-half-iid-dinov2s-cls}
\end{table}

\begin{table}[tbp]\centering\small
\setlength{\tabcolsep}{4pt}
\begin{tabular}{lrlllrr}\toprule
Fresh share & $k$ & Trained predictor & CF WM, feature & CF WM, readout & Copy & Oracle \\ \midrule
25\% & 1 & 71.7 {\scriptsize[68.1, 74.7]} & 90.7 {\scriptsize[89.0, 92.0]} & 91.4 {\scriptsize[89.8, 92.7]} & 24.3 & 81.1 \\
 & 2 & 71.7 {\scriptsize[67.6, 74.8]} & 91.0 {\scriptsize[89.3, 92.3]} & 91.7 {\scriptsize[90.1, 93.1]} & 24.3 & 81.1 \\
 & 3 & 71.8 {\scriptsize[67.7, 74.8]} & 90.6 {\scriptsize[88.9, 92.0]} & 91.1 {\scriptsize[89.3, 92.5]} & 24.3 & 81.1 \\
50\% & 1 & 57.4 {\scriptsize[53.0, 61.1]} & 88.3 {\scriptsize[86.6, 89.8]} & 89.4 {\scriptsize[87.5, 90.9]} & 18.9 & 67.8 \\
 & 2 & 56.9 {\scriptsize[53.0, 60.3]} & 88.3 {\scriptsize[86.7, 89.6]} & 89.5 {\scriptsize[87.7, 91.2]} & 18.9 & 67.8 \\
 & 3 & 56.8 {\scriptsize[52.7, 60.4]} & 88.1 {\scriptsize[86.2, 89.5]} & 89.2 {\scriptsize[87.3, 90.9]} & 18.9 & 67.8 \\
75\% & 1 & 45.8 {\scriptsize[42.4, 49.0]} & 85.9 {\scriptsize[83.8, 87.6]} & 87.9 {\scriptsize[85.8, 89.6]} & 13.2 & 56.4 \\
 & 2 & 45.6 {\scriptsize[42.3, 48.8]} & 85.2 {\scriptsize[82.9, 87.1]} & 87.3 {\scriptsize[85.1, 89.1]} & 13.2 & 56.4 \\
 & 3 & 45.1 {\scriptsize[41.0, 48.4]} & 85.0 {\scriptsize[82.3, 87.0]} & 86.8 {\scriptsize[84.2, 88.8]} & 13.2 & 56.4 \\
100\% & 1 & 38.1 {\scriptsize[35.3, 40.7]} & 82.4 {\scriptsize[79.9, 84.5]} & 85.6 {\scriptsize[83.4, 87.4]} & 7.5 & 49.1 \\
 & 2 & 37.9 {\scriptsize[35.0, 40.7]} & 80.4 {\scriptsize[75.6, 84.3]} & 84.0 {\scriptsize[79.6, 87.6]} & 7.5 & 49.1 \\
 & 3 & 37.7 {\scriptsize[35.0, 40.3]} & 80.0 {\scriptsize[75.4, 83.7]} & 83.4 {\scriptsize[79.6, 86.5]} & 7.5 & 49.1 \\
\bottomrule\end{tabular}
\caption{\textbf{Fresh-share ablation on HandReach, MAE-B.} Physical counterfactual $R^2$ in percent with paired seed/parent $95\%$ intervals; three factual-selected predictors per row (all-anchor training, width 1024, weight decay 0.3; selection on held factual error only), with the frozen readouts, physics and query law of the benchmark. \emph{Oracle}: the parent-conditional Monte Carlo reference. With a persistent component, it uses privileged past information and is an upper bound on the best $R^2$ achievable from the image window, not that window's exact optimum. These rows remain diagnostics.}
\label{tab:pinned-hand-iid}
\end{table}

\clearpage

\section{Planning Protocols and Results}
\label{app:closed-loop}
\begin{figure}[htbp]\centering
\includegraphics[width=\linewidth]{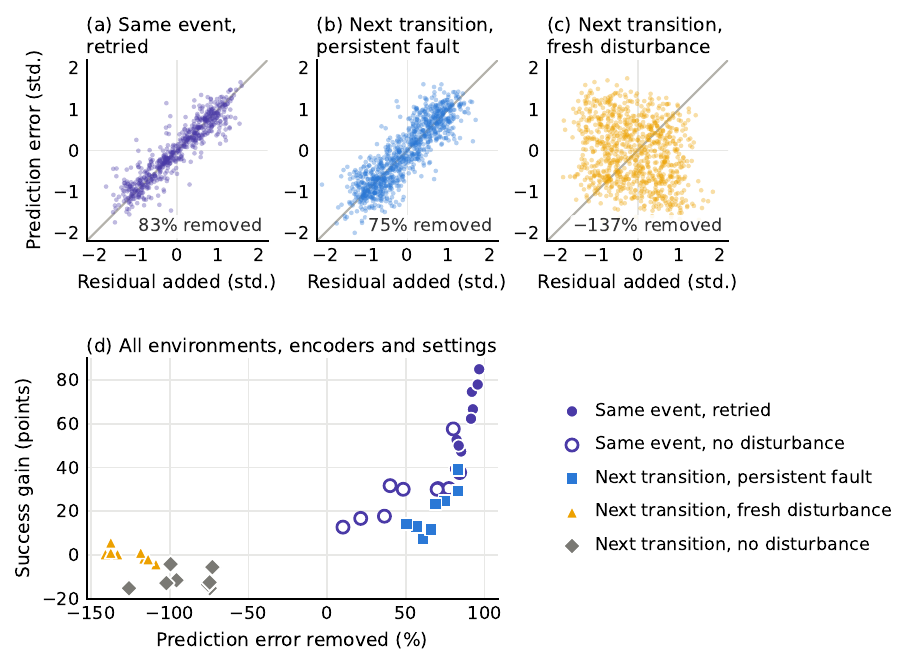}
\caption{\textbf{Command-specific prediction errors and residual corrections.} ({\bf a})--({\bf c}) BlueROV2, MAE-B: for each command the counterfactual world model
executed, the error of the trained prediction for that command against the
residual it added, in units of the coordinate's standard deviation. Points on
the diagonal are corrected exactly. Within the same event the residual
predicts the error; for the next transition it does so under a persistent
fault and not under a fresh disturbance. ({\bf d}) For every environment, encoder
and setting, the success gain against the fraction of the trained
prediction's error that the residual removes.}
\label{fig:closed-loop-why}
\end{figure}

\paragraph{Frozen models and simulator replay.}Each run loads one frozen
encoder, one factual-selected one-frame predictor (seeds $0$--$2$) and the frozen state readout of \cref{tab:main-cf}; BlueROV2 uses the
$32$k-episode predictors. Nothing is fitted. The simulator runs as a separate
process in the runtime that produced the evaluation cohorts, driven through its
native reset, step and render calls with the evaluation configuration. In-loop
encoding reproduces the stored evaluation features to within one float16
rounding step ($0.008$), and in-loop decoded predictions reproduce the scored
ones to $10^{-7}$. For liquid transport, simulated in the SoftGym benchmark through PyFleX, the
Python interface to the NVIDIA FleX particle simulator, replaying the
evaluation cohort reproduces the full native state, the sources and the
endpoints exactly ($128$ parents, maximum absolute difference $0$). Its native
episode ends after $75$ native steps ($22$ per transition), too early for the
two prefix and three replanning transitions. Liquid transport therefore enters
the one-shot and retry tasks, but not consecutive replanning; each event
takes $66$ native steps.

\begin{figure}[htbp]\centering
\includegraphics[width=\linewidth]{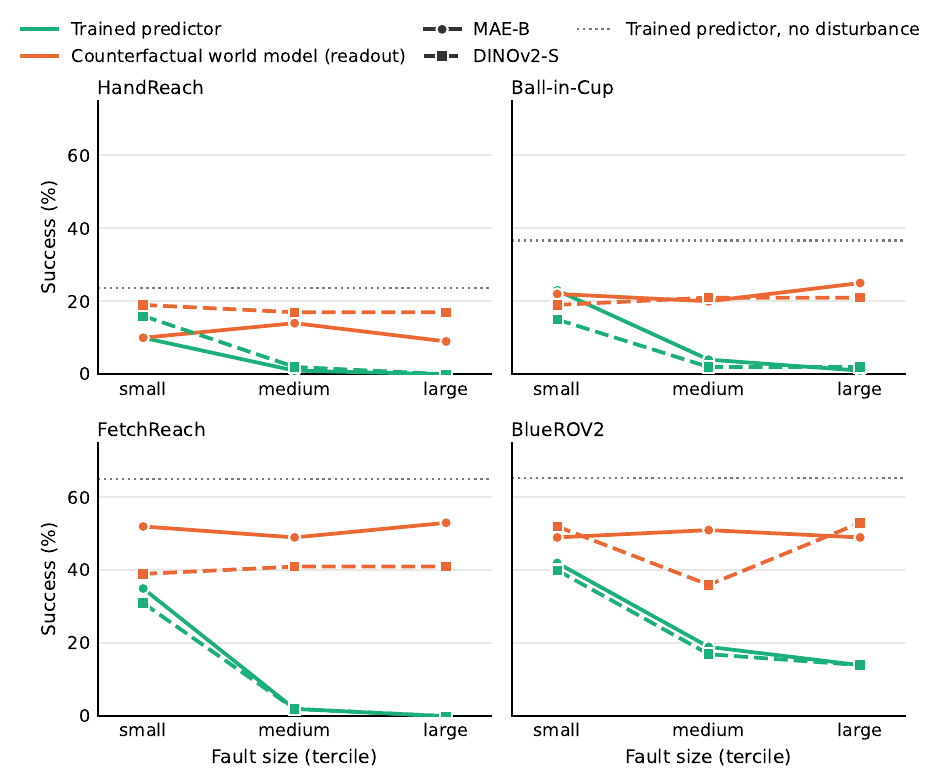}
\caption{\textbf{Replanning success under the persistent fault, by fault magnitude.}
Success rate after three replanning steps by tercile of $|\varepsilon|$
($100$ episodes per point, three predictor seeds pooled). Green: trained
predictor; orange: counterfactual world model; dotted: MAE-B trained predictor
without disturbance; solid MAE-B, dashed DINOv2-S.}
\label{fig:closed-loop-fault}
\end{figure}

\paragraph{Replanning.}Episode
$i\in\{0,\dots,299\}$ resets with seed $3{,}109{,}000{,}000+i$, disjoint from
every collection, confirmation and evaluation range; episodes $100s$ to
$100s+99$ use predictor seed $s$. After the native reset, the benchmark's
behavior policy executes the two prefix transitions, then the planner acts for
three transitions. Three disturbance conditions use the benchmark disturbance
$\sigma\tanh(g)$ of \cref{eq:pinned-driver}, $g\sim\mathcal N(0,1)$, with each
environment's bound $\sigma$:
\emph{persistent}, one $g$ per episode held for all five transitions;
\emph{fresh}, an independent $g$ per transition; and \emph{none}, $g=0$. The
goal is the vector of scored coordinates reached from the warmed-up state by
one command drawn uniformly from the command support, under the persistent
fault, or under zero disturbance in the other two conditions. The horizon of
three planning transitions keeps every episode inside the native episode
limits (BlueROV2: $100$ native steps, five transitions; FetchReach: $50$ native
steps); it was fixed from these limits before any full run. Results are in
\cref{tab:closed-loop}; under the persistent fault, the counterfactual world
model's success barely depends on the fault size, whereas the trained
predictor's falls sharply (\cref{fig:closed-loop-fault}).

\paragraph{One-shot planning.}Episode
$i\in\{0,\dots,299\}$ resets with seed $3{,}110{,}000{,}000+i$ (predictor seed
$\lfloor i/100\rfloor$), runs the two behavior-policy prefix transitions and one
event transition, all three with independent drivers from the benchmark law,
and draws the factual and goal commands independently and uniformly from the
command support. The learned and true-simulator planners rebuild the same event
from these seeds, and the report checks that their goals agree exactly. The
oracle planner (no outcome) minimizes the expected squared standardized
distance over the driver, computed with six-node Gauss--Hermite quadrature, and
uses the same cross-entropy search as the learned planners ($256$ candidates,
$32$ elites, four rounds); for the one-dimensional liquid-transport command,
both true-simulator planners use $16$ candidates per round and run on episodes
$0$--$99$, since each simulator evaluation takes about $1.5$ seconds.

\begin{figure}[htbp]\centering
\includegraphics[width=\linewidth]{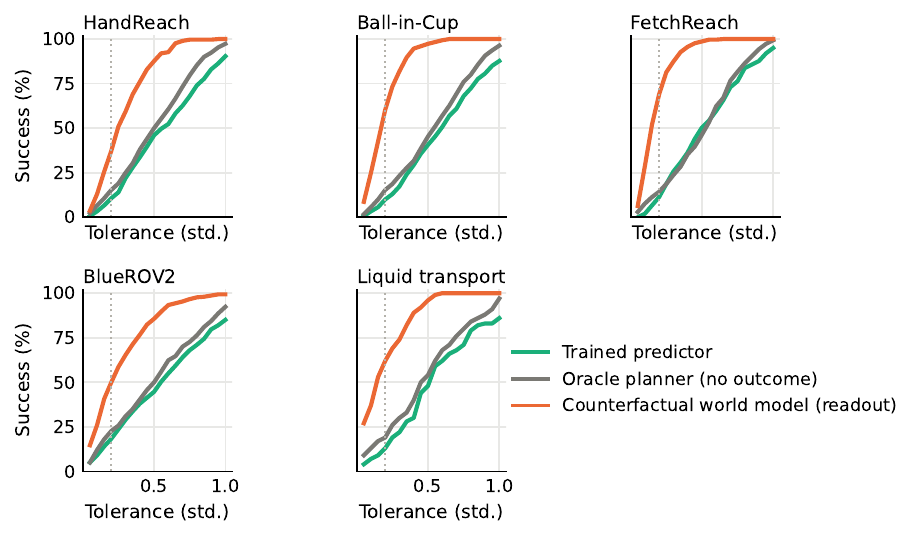}
\caption{\textbf{One-shot success across goal tolerances.} Success rate of each one-shot planner
(\cref{tab:oneshot}) against the standardized-distance tolerance, MAE-B; the
dotted vertical line marks the predeclared tolerance $0.2$.}
\label{fig:oneshot-tolerance}
\end{figure}
\begin{table}[htbp]\centering\small
\setlength{\tabcolsep}{2.5pt}
\begin{tabular}{llrrrrll}\toprule
& & & \multicolumn{2}{c}{Oracle planner} & & \multicolumn{2}{c}{Counterfactual world model minus} \\ \cmidrule(lr){4-5}\cmidrule(lr){7-8}
Encoder & Environment & Predictor & No outcome & True dist. & CF WM & Predictor & Oracle (no outcome) \\ \midrule
MAE-B & HandReach & 11 & 15 & 100 & 37 & +27 {\scriptsize[+20,+33]} & +22 {\scriptsize[+16,+28]} \\
 & Ball-in-Cup & 10 & 15 & 100 & 60 & +50 {\scriptsize[+45,+56]} & +45 {\scriptsize[+38,+51]} \\
 & FetchReach & 11 & 14 & 100 & 69 & +58 {\scriptsize[+51,+64]} & +55 {\scriptsize[+48,+61]} \\
 & BlueROV2 & 18 & 23 & 100 & 50 & +32 {\scriptsize[+25,+38]} & +27 {\scriptsize[+21,+34]} \\
 & Liquid transport & 13 & 19 & 100 & 62 & +49 {\scriptsize[+39,+59]} & +43 {\scriptsize[+32,+55]} \\
\addlinespace
DINOv2-S & HandReach & 13 & 15 & 100 & 35 & +23 {\scriptsize[+16,+29]} & +20 {\scriptsize[+13,+27]} \\
 & Ball-in-Cup & 9 & 15 & 100 & 46 & +37 {\scriptsize[+31,+43]} & +31 {\scriptsize[+24,+37]} \\
 & FetchReach & 12 & 14 & 100 & 61 & +49 {\scriptsize[+42,+55]} & +46 {\scriptsize[+39,+53]} \\
 & BlueROV2 & 23 & 23 & 100 & 48 & +25 {\scriptsize[+19,+31]} & +25 {\scriptsize[+19,+32]} \\
 & Liquid transport & 25 & 19 & 100 & 64 & +39 {\scriptsize[+25,+52]} & +45 {\scriptsize[+33,+57]} \\
\bottomrule\end{tabular}
\caption{\textbf{One-shot planning results.} Success rate (\%) of the command each planner chooses after observing one factual transition of a benchmark event, executed in that same event (standardized distance to the goal below 0.2). \emph{Predictor} and \emph{CF WM} plan from frames with the same frozen predictor and readout; the oracle planners use the true simulator and full state, \emph{No outcome} minimizing the expected cost over the disturbance law and \emph{True dist.} knowing the realized disturbance. Differences are paired, with $95\%$ episode-bootstrap intervals. Episodes: 300 HandReach, 300 Ball-in-Cup, 300 FetchReach, 300 BlueROV2, 100 Liquid transport (recall \cref{fig:oneshot}).}
\label{tab:oneshot}
\end{table}

\paragraph{Same-event retries.}The replanning episodes and
goal commands also define one event: the native reset, the two prefix
transitions and the event transition, with drivers drawn independently per
transition from the benchmark law (or zero). The goal is the vector of scored
coordinates that the goal command reaches in the event. The first attempt plans
with the trained predictor in both arms; each further attempt resets and
replays the prefix, verifies that the scored coordinates match the event start
exactly, and plans again with a new search seed, the counterfactual world model adding
the readout residual of its previous attempt. The retry task was specified
after the replanning results and before any retry was run. Results are in
\cref{tab:retry}.

\begin{figure}[htbp]\centering
\includegraphics[width=\linewidth]{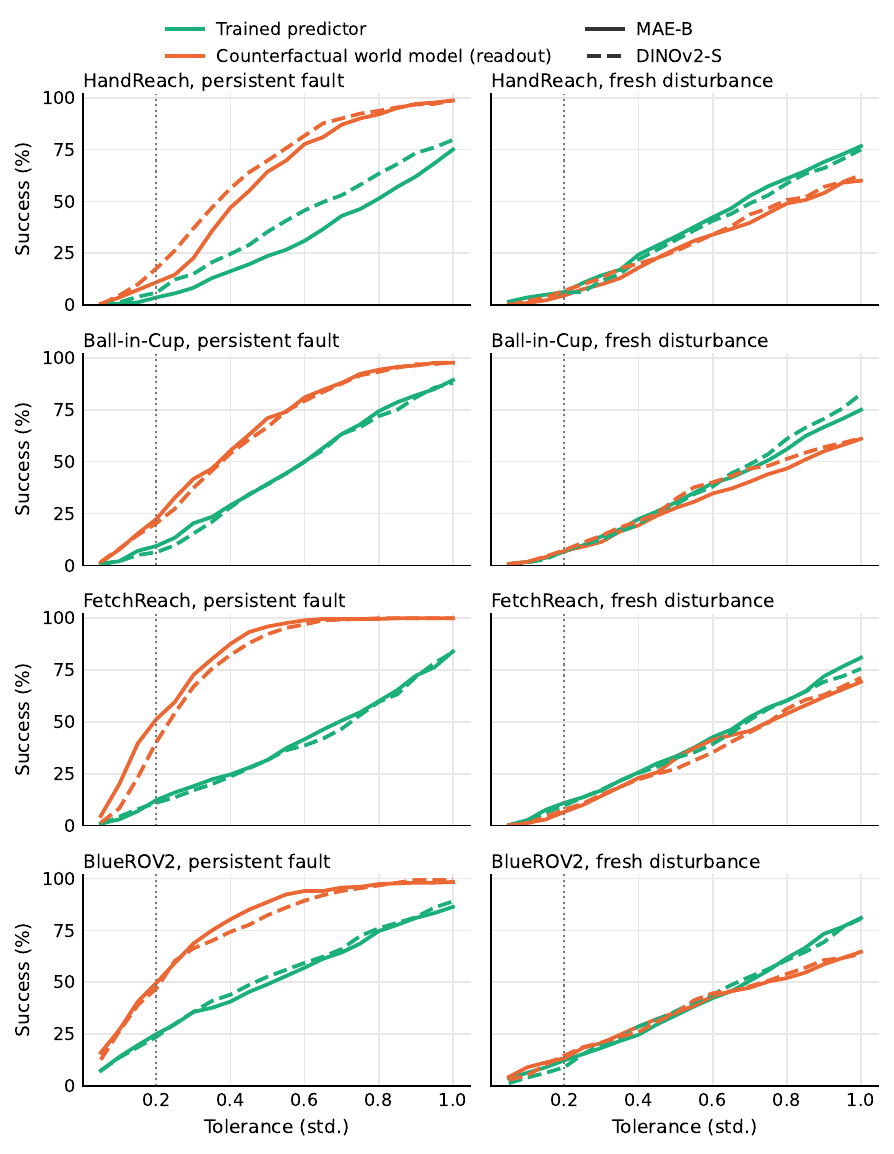}
\caption{\textbf{Replanning success across goal tolerances.}
Success rate after three replanning steps against the standardized-distance
tolerance, for the persistent fault ({\bf left}) and the fresh disturbance
({\bf right}); solid MAE-B, dashed DINOv2-S, $300$ episodes per curve. The
dotted line marks the predeclared tolerance $0.2$.}
\label{fig:closed-loop-tolerance}
\end{figure}

\paragraph{Planner and success criterion.}The cross-entropy search
draws $256$ commands uniformly over the command support, keeps the $32$ with
the lowest cost, refits a diagonal Gaussian (standard deviation at least $1\%$
of the bound), and repeats for four rounds with clipped Gaussian draws; it
executes the best command seen. The cost sums squared standardized differences between the decoded
prediction and the goal, using each coordinate's standard deviation over
factual outcomes of the evaluation cohort. Candidate draws are seeded by episode and step and are identical for
the two arms. Success uses the true simulator coordinates: their
root-mean-square standardized distance to the goal after the last step must be
below $0.2$. Intervals are $95\%$ bootstraps over the episodes of each cell,
paired across arms. The one-shot ordering of \cref{tab:oneshot} holds across
tolerances (\cref{fig:oneshot-tolerance}); in replanning, the persistent-fault
gain holds across tolerances, whereas with fresh disturbances the correction
lowers success at loose tolerances (\cref{fig:closed-loop-tolerance}).

\begin{table}[htbp]\centering\footnotesize
\setlength{\tabcolsep}{3.5pt}
\begin{tabular}{llrrrrrrrrr}\toprule
 & & \multicolumn{3}{c}{Disturbance, 2nd attempt} & \multicolumn{3}{c}{Disturbance, 3rd attempt} & \multicolumn{3}{c}{No disturbance, 3rd attempt} \\
\cmidrule(lr){3-5}\cmidrule(lr){6-8}\cmidrule(lr){9-11}
Encoder & Environment & Pred. & CF & $\Delta$ & Pred. & CF & $\Delta$ & Pred. & CF & $\Delta$ \\ \midrule
MAE-B & HandReach & 12 & 50 & +38 {\scriptsize[+32,+44]} & 13 & 60 & +47 {\scriptsize[+41,+53]} & 34 & 64 & +30 {\scriptsize[+23,+37]} \\
 & Ball-in-Cup & 11 & 63 & +52 {\scriptsize[+46,+58]} & 11 & 86 & +75 {\scriptsize[+70,+80]} & 33 & 91 & +58 {\scriptsize[+52,+63]} \\
 & FetchReach & 13 & 80 & +68 {\scriptsize[+62,+73]} & 12 & 97 & +85 {\scriptsize[+81,+89]} & 62 & 99 & +38 {\scriptsize[+32,+43]} \\
 & BlueROV2 & 23 & 60 & +37 {\scriptsize[+31,+43]} & 21 & 71 & +50 {\scriptsize[+44,+56]} & 57 & 75 & +18 {\scriptsize[+11,+24]} \\
 & Liquid transport & 21 & 69 & +48 {\scriptsize[+43,+54]} & 21 & 87 & +67 {\scriptsize[+61,+72]} & 63 & 93 & +30 {\scriptsize[+25,+36]} \\
\addlinespace
DINOv2-S & HandReach & 11 & 41 & +29 {\scriptsize[+24,+35]} & 12 & 51 & +39 {\scriptsize[+33,+45]} & 41 & 58 & +17 {\scriptsize[+9,+24]} \\
 & Ball-in-Cup & 9 & 46 & +37 {\scriptsize[+31,+42]} & 9 & 62 & +53 {\scriptsize[+47,+58]} & 33 & 64 & +32 {\scriptsize[+25,+38]} \\
 & FetchReach & 15 & 78 & +63 {\scriptsize[+57,+68]} & 14 & 92 & +78 {\scriptsize[+73,+83]} & 65 & 95 & +30 {\scriptsize[+25,+36]} \\
 & BlueROV2 & 22 & 56 & +34 {\scriptsize[+28,+41]} & 24 & 63 & +39 {\scriptsize[+33,+46]} & 55 & 68 & +13 {\scriptsize[+6,+19]} \\
 & Liquid transport & 22 & 70 & +48 {\scriptsize[+42,+53]} & 22 & 84 & +62 {\scriptsize[+57,+68]} & 61 & 91 & +30 {\scriptsize[+25,+35]} \\
\bottomrule\end{tabular}
\caption{\textbf{Same-event retry results.} Success rate (\%) after the second and third attempts; the first attempt is shared by both arms (HandReach 13, Ball-in-Cup 11, FetchReach 13, BlueROV2 20, Liquid transport 21\% with MAE-B under the disturbance). Each retry replays the event: same state, same disturbance drawn from the benchmark law. \emph{Pred.}: the trained predictor plans alone; \emph{CF}: the counterfactual world model adds the readout residual of its last observation; $\Delta$: paired difference with $95\%$ episode-bootstrap interval. $300$ episodes per cell ($100$ for each of three predictor seeds); both arms share predictor, readout, search draws, goals and disturbances.}
\label{tab:retry}
\end{table}
\begin{table}[htbp]\centering\small
\setlength{\tabcolsep}{4pt}
\begin{tabular}{llrrrrrrrrr}\toprule
 & & \multicolumn{3}{c}{Persistent fault} & \multicolumn{3}{c}{Fresh disturbance} & \multicolumn{3}{c}{No disturbance} \\
\cmidrule(lr){3-5}\cmidrule(lr){6-8}\cmidrule(lr){9-11}
Encoder & Environment & Pred. & CF & $\Delta$ & Pred. & CF & $\Delta$ & Pred. & CF & $\Delta$ \\ \midrule
MAE-B & HandReach & 4 & 11 & +7 {\scriptsize[+4,+11]} & 6 & 5 & $-$2 {\scriptsize[$-$6,+2]} & 24 & 12 & $-$12 {\scriptsize[$-$18,$-$6]} \\
 & Ball-in-Cup & 9 & 22 & +13 {\scriptsize[+8,+19]} & 7 & 7 & 0 {\scriptsize[$-$4,+4]} & 37 & 24 & $-$13 {\scriptsize[$-$20,$-$6]} \\
 & FetchReach & 12 & 51 & +39 {\scriptsize[+33,+46]} & 11 & 7 & $-$4 {\scriptsize[$-$9,0]} & 65 & 59 & $-$6 {\scriptsize[$-$14,+2]} \\
 & BlueROV2 & 25 & 50 & +25 {\scriptsize[+18,+31]} & 12 & 13 & +1 {\scriptsize[$-$4,+6]} & 65 & 52 & $-$14 {\scriptsize[$-$21,$-$7]} \\
\addlinespace
DINOv2-S & HandReach & 6 & 18 & +12 {\scriptsize[+7,+17]} & 6 & 7 & +1 {\scriptsize[$-$3,+5]} & 30 & 25 & $-$4 {\scriptsize[$-$11,+2]} \\
 & Ball-in-Cup & 6 & 20 & +14 {\scriptsize[+9,+19]} & 7 & 7 & 0 {\scriptsize[$-$4,+4]} & 35 & 19 & $-$15 {\scriptsize[$-$22,$-$9]} \\
 & FetchReach & 11 & 40 & +29 {\scriptsize[+23,+36]} & 10 & 8 & $-$2 {\scriptsize[$-$6,+3]} & 56 & 41 & $-$15 {\scriptsize[$-$24,$-$7]} \\
 & BlueROV2 & 24 & 47 & +23 {\scriptsize[+17,+30]} & 9 & 15 & +6 {\scriptsize[+1,+10]} & 63 & 50 & $-$13 {\scriptsize[$-$20,$-$6]} \\
\bottomrule\end{tabular}
\caption{\textbf{Three-step replanning results.} Success rate (\%) after three replanning steps: the standardized distance of the true simulator coordinates to the goal is below 0.2. \emph{Pred.}: the trained predictor plans alone; \emph{CF}: the counterfactual world model adds the readout residual of its last observation; $\Delta$: paired difference with $95\%$ episode-bootstrap interval. $300$ episodes per cell ($100$ for each of three predictor seeds); both arms share predictor, readout, search draws, goals and disturbances.}
\label{tab:closed-loop}
\end{table}

\clearpage

\section{Stress-Test and Model-Selection Protocols}
\label{app:stress}
\paragraph{Equilibrium prediction for the gain-fault interaction.}
\label{app:gain-fault-analysis}
The analytic comparison in \cref{sec:theory-checks} uses the equilibrium response below. Here $T$ is the centered disturbance draw, $c$ is the equilibrium response scale and $g$ is the command gain.
For an equilibrium response
\[
    Y
    =
    c\left\{
    ga+
    \sigma T[(1-\lambda)+\lambda u]
    \right\},
    \qquad
    u=a/a_{\max},
\]
with uniformly sampled normalized actions, the no-outcome disturbance variance is proportional to
\[
    m(\lambda)
    =
    (1-\lambda)^2+\frac{\lambda^2}{3},
\]
while the residual-reuse error variance is proportional to $2\lambda^2/3$. Hence
\begin{equation}
    \rho(\lambda)
    =
    \frac{2\lambda^2}{3m(\lambda)}.
    \label{eq:gain-fault-interaction-ratio}
\end{equation}
This gives $\rho=0$, $0.5$ and $2$ at $\lambda=0$, $0.5$ and $1$, respectively.

\paragraph{Physical interaction screen.}For every coordinate
recorded in the evaluation branches (parents $\times$ disturbance draws
$\times$ three unequal queries, \cref{app:pinned-protocol}), we compute the
oracle (no outcome) $R^2$ from within-parent conditional variances, the
factual-copy $R^2$, the ideal residual $R^2$ obtained with within-parent
Monte Carlo estimates of the conditional means, and the interaction ratio $\rho$ (\cref{eq:interaction-ratio}),
with whole-parent bootstrap intervals; \cref{tab:stress-physical} lists the
point estimates. The screen
uses no image, readout or fitted model; the per-coordinate values for all seven
environments accompany the code release.

\paragraph{Fingertip readout.}The HandReach
predictors of \cref{tab:main-cf} are rescored through a second frozen readout trained on factual frames for
all 39 recorded coordinates (three hidden layers of width 1{,}024, 30{,}000
updates, factual-validation checkpoint selection, and the same diagnostic
cohort and episode split as the state readouts of \cref{tab:main-cf}). Two criteria fix the scored mask
before scoring: simulator-only oracle (no outcome) and copy references both below $60\%$, and
factual validation recovery at least $95\%$ for both encoders. Seven fingertip
coordinates meet these selection criteria, with validation recovery $95.1$--$97.6\%$ (DINOv2-S) and
$96.0$--$98.8\%$ (MAE-B). On the evaluation population, MAE-B recovers all
seven at $95.1$--$97.7\%$, whereas DINOv2-S recovers five of them at only
$93.1$--$94.0\%$, so the DINOv2-S result is reported as a diagnostic. Pooled
$R^2$ uses summed squared errors and summed variances over the seven
coordinates, averages single-fit risk over three seeds, and resamples parents
and seeds jointly. With MAE-B, correction in feature space reaches $83.6\%$ on
these seven coordinates, against $39.8\%$ for the trained predictor, $-1.3\%$
for factual copying and $53.0\%$ for the oracle predictor (no outcome);
residual reuse with exact simulator means reaches $96.5\%$.

\paragraph{Interaction-dial calibration.}The interaction dial perturbs the normalized
command $a$ by $\sigma\varepsilon\,d\,[(1-\lambda)+\lambda\,a/a_{\max}]$, where
$d$ is the declared actuator direction pattern and $a_{\max}=g\,a_{\mathrm{lim}}$
the declared command bound; the fault magnitude never exceeds $\sigma$, so the
native no-clipping bound $g\,a_{\mathrm{lim}}+\sigma\le1$ is unchanged. For the
equilibrium model of \cref{app:gain-fault-analysis}, matching the oracle (no outcome) target
of $50\%$ with a $2\%$ actuator margin gives
$\sigma=[1+\sqrt{3v\,m(\lambda)}/0.98]^{-1}$ and $g\,a_{\mathrm{lim}}=0.98(1-\sigma)$,
with $v=\operatorname{Var}\tanh(G)=0.394$ for $G\sim\mathcal N(0,1)$. For
$\lambda\in\{0.5,1\}$ this gives $\sigma=0.609$ and action gains $0.850$
(HandReach) and $1.093$ (Ball-in-Cup). Development cohorts use 256 parents
$\times$ 64 draws with fresh reset seeds, and all native control and replay
tests pass. The $\lambda=0$ rows of \cref{tab:dial} reuse the benchmark
confirmation cohorts.

\paragraph{BlueROV2 disturbance recovery.}\label{app:bluerov-gate}
BlueROV2 is rendered from a level camera with elevation $-10$ degrees, distance $1.6$ and azimuth
$90$, selected by held-factual recovery. It changes no physics, action law,
reset state or coordinate mask. The disturbance readout uses 32k factual
episodes (26,214 fitting anchors), with the architecture, budget and
checkpoint rule above. Held-out recovery is $97.9$, $96.0$ and $96.2\%$
for MAE-B, DINOv2-S and DINOv2-B; on independent confirmation it is
$97.2\%$ $[96.3,97.8]$, $95.6\%$ $[94.7,96.2]$ and
$95.5\%$ $[94.7,96.1]$. State recovery is respectively $99.3/99.2$,
$98.4/98.3$ and $98.6/98.4\%$ (held-out/confirmation). The no-outcome oracle
has upper $95\%$ bound $53.0\%$; decoded factual-copy upper bounds are
$8.1$, $8.4$ and $8.3\%$. For both DINOv2 encoders, the disturbance pass
uses the point estimate: the lower confirmation bound is $94.7\%$.
The disturbance head enters
neither counterfactual prediction nor its score.

\paragraph{Factual model selection.}
The reference arm in \cref{tab:wm-quality} uses width 512, the $t=2$ fitting
anchor and weight decay $10^{-4}$. The selected recipe minimizes the mean
held-out factual error over the seven data cells with MAE-B.
The development grid
uses MAE-B, seed zero and one frame on all seven data cells (BlueROV2 with its
level camera) and varies the training anchors (the $t=2$ anchor only, or all
three recorded factual transitions of each fitting episode), AdamW weight decay
($10^{-4}$, $0.05$, $0.3$) and predictor width ($256$, $512$, $1{,}024$; two
blocks), with validation every 100 updates. Validation episodes and anchors are
identical across the grid. The criterion is the mean over cells of the
held-out factual mean squared error in raw feature units, relative to the
reference recipe. Fitting all recorded transitions dominates every other factor
(held-out error $4$--$10\%$ lower than with the evaluation anchor alone, in
every completed cell), width $1{,}024$ is best at every training scope, and
weight decay changes the criterion by less than $2\%$. Over all seven cells,
the recipe with all transitions, width $1{,}024$ and weight decay $0.3$ has the
lowest mean relative error ($0.931$), ahead of weight decays $0.05$ ($0.933$)
and $10^{-4}$ ($0.934$) at the same width, with the largest gains on BlueROV2
and Racetrack ($0.89$ and $0.90$). This grid is a separate development run:
its reference fits use seed zero and validate every 100 updates, whereas the
\emph{orig.} arm of \cref{tab:wm-quality} reports the original three-seed fits,
so its factual gains differ from these relative errors. The final refits were launched with this
recipe once it led on every completed cell and were confirmed when the
remaining width-$1{,}024$ and reference cells finished; $30$ of the $126$ grid
fits (narrower widths and the anchor-only scope on four cells, worse on every
cell where they completed) were not run to completion. Every arm validates on
held-out factual episodes at the $t=2$ anchor, the anchor at which
counterfactuals are evaluated, so selection sees source states drawn like the
evaluation ones but only their factual outcomes.

\paragraph{Ensembling and readout correction.}Both are computed from the same stored predictions as the
single-seed arms of \cref{tab:main-cf}. The ensemble averages the embedding
predictions of the three fitted seeds before decoding; readout transport
applies \cref{eq:counterfactual-jepa-chart} to the decoded predictions of the
factual endpoint and of the trained mean at the factual and query actions.

\paragraph{Pixel, feature and readout transport.}
\label{app:pixel-transport}
The comparison in \cref{sec:where-transport} uses the title-page boat with a uniform current, the east-west mean flow without eddies, and the frozen MAE-B encoder and state readout fitted on that collection. For $16$ fresh events and eight alternative headings each, conditional means over the current average $32$ draws shared by all headings, and a duration $h$ scales both the commanded move and the current. In their own coordinates, residual reuse gives interaction ratios of $1.74$--$2.18$ in pixels and $1.92$--$2.30$ in the $768$ features for $h$ from $1$ to $0.05$. Against the true position, the relative risks at $h=1$, $0.5$ and $0.25$ are $5.86$, $3.70$ and $4.06$ for pixel correction read back through the encoder, $0.31$, $0.32$ and $0.47$ for correction in feature space, and $0.002$, $0.014$ and $0.064$ for correction after the readout. For shorter transitions, the readout's own error approaches the displacement caused by the current, and every variant degrades.

\clearpage

\section{Short-Time Bounds and Assumptions}
\label{app:short-time-residual}

\subsection{Baseline Feasibility}
The benchmark thresholds of \cref{app:pinned-protocol} restrict which dynamics
can serve as benchmarks: they fail for pure integrators
(\cref{prop:integrator-benchmark-limit}) and very short transitions
(\cref{prop:small-step-baselines}), but can hold for contracting systems such
as servoed actuators (\cref{prop:contracting-benchmark}).

\begin{proposition}[A continuing-integrator benchmark limit]
\label{prop:integrator-benchmark-limit}
Consider $S_{t+1}=S_t+A_t+\varepsilon_t$, with independent fresh innovations
of mean zero and variance $N>0$. Past actions may use arbitrary feedback from
the available past, but cannot observe their current innovation. At a
continuing anchor $t\ge1$, let the factual and query actions be independent
draws from the same law with variance $A>0$, independent of $S_t,\varepsilon_t$.
With $V=\operatorname{Var}(S_t)+A+N$, the state-information Bayes reference
and same-disturbance factual-copy scores satisfy
\begin{equation}
 R^2_B=1-\frac{N}{V},\qquad
 R^2_{\mathrm{copy}}=1-\frac{2A}{V},\qquad V\ge A+2N.
\end{equation}
They cannot both be strictly below $0.60$, although population residual reuse
has zero counterfactual error.
\end{proposition}
\begin{proof}
The latest independent innovation gives
$\operatorname{Var}(S_t)=\operatorname{Var}(S_{t-1}+A_{t-1})+N\ge N$.
The query conditional mean is $S_t+A_t'$, with risk $N$.
Factual copying errs by $A_t-A_t'$, with risk $2A$.
Both strict score thresholds would imply $N>0.4V$ and $A>0.2V$,
contradicting $V\ge A+2N>V$.
The residual prediction is exactly
$S_{t+1}(A_t)+(A_t'-A_t)=S_{t+1}(A_t')$.
\end{proof}
\begin{proposition}[A contracting system can meet both baseline gates]
\label{prop:contracting-benchmark}
Let $S_{t+1}=\rho S_t+(1-\rho)(A_t+\varepsilon_t)$ with
$0\le\rho<1$, mutually independent zero-mean IID actions and innovations,
and variances $A>0$ and $N>0$. On its stationary population, with an
independent query action from the same law and the same current innovation,
\begin{equation}
 R_B^2=1-(1-\rho^2)\frac{N}{A+N},\qquad
 R_{\mathrm{copy}}^2=1-2(1-\rho^2)\frac{A}{A+N}.
\end{equation}
For $\rho=1/2$ and $N=2A$, both scores equal $1/2$, while population
residual reuse remains exact.
\end{proposition}
\begin{proof}
Stationarity gives $V=\rho^2V+(1-\rho)^2(A+N)$, hence
$V=(1-\rho)(A+N)/(1+\rho)$. The conditional-mean risk is
$(1-\rho)^2N$, and same-innovation factual copying has risk
$2(1-\rho)^2A$. Dividing by $V$ gives the scores. Subtracting the factual
conditional mean leaves $(1-\rho)\varepsilon_t$, which cancels exactly
when reused with the query conditional mean.
\end{proof}
\begin{proposition}[Small-step absolute-state baselines]
\label{prop:small-step-baselines}
For a scalar scored coordinate $S$ with $0<\operatorname{Var}(S)<\infty$,
let $X_h$ and $Y_h$ be factual and same-disturbance query endpoints.
Suppose $\E|X_h-S|^2+\E|Y_h-S|^2=O(h^2)$ as $h\downarrow0$,
on the same source population and uniformly bounded action/noise scaling.
If the no-outcome information includes $S$, then
\begin{equation}
 R^2_{B,h}=1-O(h^2),\qquad R^2_{\mathrm{copy},h}=1-O(h^2).
\end{equation}
Thus both strict low-baseline gates fail for sufficiently small $h$, even
when residual counterfactual error is of a higher order or identically zero.
\end{proposition}
\begin{proof}
The assumed $L^2$ convergence implies
$\operatorname{Var}(Y_h)\to\operatorname{Var}(S)>0$.
Conditional expectation has risk at most the valid no-outcome predictor $S$,
namely $\E|Y_h-S|^2=O(h^2)$. Factual copying has risk
$\E|Y_h-X_h|^2\le2\E|Y_h-S|^2+2\E|X_h-S|^2=O(h^2)$.
Divide by the common endpoint variance.
\end{proof}
\subsection{ODE Remainder and Inertial Extension}
Write $V_{a,\varepsilon}(s)=v_s(s,a)+b_s(s,\varepsilon)$.
Fix a source state $s$ and a bounded action domain $\mathcal A$. For
$0\le t\le h_0$, suppose the solutions of \eqref{eq:short-time-physical-ode}
exist in a region on which $V_{a,\varepsilon}$ is continuously differentiable
and $\mathcal E$ is twice continuously differentiable. We require that source
features determine the relevant physical source, e.g., through a measurable
inverse $s=s(z)$ on the source support; a weaker sufficient condition is that
the encoded transitions and the two first-order terms below factor through $z$
with the common disturbance law assumed in the main theorem. Accurate recovery
of a few proprioceptive coordinates proves neither condition.

Suppose the following bound holds along every relevant solution, uniformly
over $s,a,t$, with $M(\varepsilon)\in L^2(\nu_z)$ and a uniform bound on its
conditional $L^2$ norm:
\begin{equation}
 \|D^2\mathcal E(S_t)\|_{\mathrm{op}}\,
       \|V_{a,\varepsilon}(S_t)\|^2
 +\|D\mathcal E(S_t)\|_{\mathrm{op}}\,
       \|DV_{a,\varepsilon}(S_t)\|_{\mathrm{op}}
       \|V_{a,\varepsilon}(S_t)\|
 \le M(\varepsilon).
 \label{eq:ode-encoded-acceleration-bound}
\end{equation}
Then, with $Z(t)=\mathcal E(S_t)$,
\begin{align}
 Z'(0)&=D\mathcal E(s)v_s(s,a)+D\mathcal E(s)b_s(s,\varepsilon),\\
 Z''(t)&=D^2\mathcal E(S_t)[V_{a,\varepsilon}(S_t),V_{a,\varepsilon}(S_t)]
       +D\mathcal E(S_t)DV_{a,\varepsilon}(S_t)V_{a,\varepsilon}(S_t).
\end{align}
Taylor's integral remainder is
\begin{equation}
 R_h(z,a,\varepsilon)=\int_0^h(h-t)Z''(t)\,dt,
 \qquad
 \|R_h(z,a,\varepsilon)\|_{L^2(\nu_z)}
 \le \tfrac12 h^2\|M\|_{L^2(\nu_z)}.
 \label{eq:ode-encoded-remainder}
\end{equation}
This proves \eqref{eq:short-time-expansion} with
$C=\tfrac12\sup_z\|M\|_{L^2(\nu_z)}$. On a common compact region, bounded
disturbances and uniform bounds on the displayed derivatives are one simple
sufficient set of conditions. Unbounded disturbances are allowed when the
stated existence and moment conditions hold.

The noise is held constant \emph{within} each transition. It may be
independent between transitions or drawn from a persistent field; the result
is conditional on the declared source and common action-conditional law. A
visited-state distribution coupled to a persistent field must satisfy that
conditional-law assumption and cannot inherit it merely from an initial-reset
marginal. An action-independent draw passed through a nonlinear actuator or
contact map need not have the vector-field form
\eqref{eq:short-time-physical-ode}, as in the saturation and contact
exceptions measured in \cref{sec:stress}.

\paragraph{Remainder scale.}Let
$A_{\mathcal E}=\sup_z\|M\|_{L^2(\nu_z)}$ in
\eqref{eq:ode-encoded-acceleration-bound}, on a declared common time interval
and source/action support, and fix an outcome scale $\sigma_{\rm ref}>0$.
A dimensionless sufficient small-remainder parameter is
\[
 \eta_h=\frac{h^2 A_{\mathcal E}}{\sigma_{\rm ref}},\qquad
 \frac{\|T_R-Y_h(a')\|_{L^2}}{\sigma_{\rm ref}}\le\eta_h.
\]
The constant includes both encoder curvature and physical dynamics,
$\|D^2\mathcal E\|\|V\|^2+\|D\mathcal E\|\|DV\|\|V\|$, so an encoder Lipschitz
constant alone is insufficient. Slow dynamics can permit larger physical
intervals; large speed with constant encoded velocity need not introduce a
Taylor remainder. This is a conservative bound, not an invariant or uniquely
defined measure of counterfactual difficulty.

\paragraph{Risk decomposition.}Condition
on an independently fitted factual predictor and the source/action pair, and
write $d_h=e_h(z,a')-e_h(z,a)$, the estimation term of
\eqref{eq:short-time-learned-mean}. Because the population residual error in
\eqref{eq:residual-contrast-identity} is centered under $\nu_z$,
\begin{equation}
 \E_{\nu_z}\|\widehat T_R-Y_h(a')\|^2
 =\operatorname{tr}\operatorname{Var}_{\nu_z}(\Delta_h)+\|d_h\|^2.
 \label{eq:duration-data-risk-decomposition}
\end{equation}
The cross term vanishes exactly; averaging over independent training sets,
sources and action pairs preserves this identity. The first term, structural
error at a finite duration, is at most $h^4 A_{\mathcal E}^2$ under the ODE
assumptions. More factual data can reduce the second term, the estimation of
the factual mean contrast, but no rate in sample size follows without
assumptions on the predictor class, optimization and action coverage. Neither
term, hence no duration or sample-count sweep, is guaranteed to vary
monotonically in a finite experiment, and an $O(h^2)$ upper bound on the root-mean-squared error is not
a claim that the error equals a nonzero constant times $h^2$. A sample-count
sweep holds policy, action support, representation and validation population
fixed, and counts independent episodes rather than correlated frames. Very
short durations can fail the benchmark's requirement that copy and oracle (no outcome) $R^2$ stay below $60\%$
(\cref{prop:small-step-baselines}). Finite-duration comparisons alone do not test an asymptotic
$h\downarrow0$ scaling law.

\paragraph{Action-distance refinement.}The
general bound is uniform and does not assert a monotone empirical error curve.
If, in addition,
\[
 \|R_h(z,a,\varepsilon)-R_h(z,a',\varepsilon)\|_{L^2(\nu_z)}
 \le C_a h^2\|a'-a\|,
\]
then centering the remainder \emph{difference} in the proof of
\cref{thm:short-time-residual} gives the sharper bound
$\|T_R-Y_h(a')\|_{L^2}\le C_a h^2\|a'-a\|$. This is an additional uniform
Lipschitz condition on the action-dependent remainder, not a bound valid for
arbitrary out-of-support actions.

\begin{corollary}[Pose observations with hidden velocity]
\label{cor:inertial-residual}
Consider held-action, held-disturbance inertial dynamics
\[
 \ddot q=f(q,a)+b(q,\dot q,\varepsilon),\qquad z=\mathcal E(q_0).
\]
The actuation term $f$ does not depend on velocity, and $b$ does not depend
on action. Suppose $z$ determines $q_0$ on the source support, while
$V_0=\dot q_0$ may be hidden. The joint conditional law of
$\xi=(V_0,\varepsilon)$ given $z$ is common across factual and query actions;
both variables are held fixed when defining the counterfactual.
Let $\mathcal E$ be $C^3$, let $f,b$ be continuously differentiable in their
state arguments, and assume all terms in the expansion below are square
integrable. Suppose the encoded trajectories have a uniform
third-order Taylor remainder bound
$\|R_h(z,a,\xi)\|_{L^2(\xi\mid z)}\le C_3h^3$.
Then residual reuse of the conditional embedding mean satisfies
\[
 \|T_R-Y_h(a')\|_{L^2(\xi\mid z)}\le2C_3h^3,
 \qquad \E[\|T_R-Y_h(a')\|^2\mid z]\le4C_3^2h^6
\]
for each supported action pair. No velocity or disturbance label is an input
to the residual predictor.
\end{corollary}
\begin{proof}
Write $q=q_0$ and evaluate derivatives of $\mathcal E$ at $q$. Taylor's formula
gives
\begin{align}
 Y_h(a)={}&\mathcal E(q)+hD\mathcal E(q)V_0\nonumber\\
 &+\frac{h^2}{2}\Bigl(
 D\mathcal E(q)f(q,a)+D\mathcal E(q)b(q,V_0,\varepsilon)
 +D^2\mathcal E(q)[V_0,V_0]\Bigr)+R_h(z,a,\xi).
 \label{eq:inertial-encoded-expansion}
\end{align}
Every displayed term except $\frac{h^2}{2}D\mathcal E(q)f(q,a)$ is shared
across actions for the same event. That remaining term is deterministic given
$z,a$. Thus the centered same-event action contrast contains only the centered
remainder difference. Applying \eqref{eq:residual-contrast-identity} with
the nuisance variable $\xi$ proves the bounds by the same $L^2$ projection
argument as before.
\end{proof}
A sufficient remainder condition is a bound on
$\sup_{0\le t\le h_0}\|\frac{d^3}{dt^3}\mathcal E(q_t)\|$ by a random
envelope with uniformly bounded conditional second moment; then
$C_3$ can be its uniform $L^2$ bound divided by six. Action-independent drag
may depend on the hidden velocity and is allowed in $b$. Hidden velocity that
changes the leading actuation response, missing pose coordinates needed for
that response, or a policy dependent on unobserved velocity can invalidate the
corollary (see the aliasing counterexample below). It does not establish
continuing-transition performance of a benchmark evaluated only at resets.

For an explicit check, let $\ddot q=a+\varepsilon$ and $\mathcal E(q)=q^2$
on a positive-position region. If $V_0$ and $\varepsilon$ are independent,
bounded, centered, unit-variance random variables, then for sufficiently small $h$,
\[
 Y_h(a)=\bigl[q+hV_0+\tfrac12h^2(a+\varepsilon)\bigr]^2,
 \qquad
 \E|T_R-Y_h(a')|^2=(a'-a)^2\bigl(h^6+\tfrac14h^8\bigr).
\]
Hidden velocity and a nonlinear pose encoder can thus attain the $h^6$
order, while with the identity pose encoder residual reuse is exact for this
constant-acceleration example despite the hidden velocity.

\subsection{Estimation Error and Physical Readouts}
\label{app:readout-error}
\paragraph{Population baseline risks.}Fix $z,a,a'$ and take expectations under the common same-event nuisance
law. Let $\mathcal R_R$, $\mathcal R_{\mathrm{copy}}$, and
$\mathcal R_{\mathrm{mean}}$ denote embedding-space squared-error risks of
the population residual rule, factual-outcome copying $Y_h(a)$, and the
no-outcome conditional mean $\mu_h(z,a')$, respectively. With
$\Delta_h=Y_h(a')-Y_h(a)$, the bias--variance identity gives
\begin{align}
 \mathcal R_{\mathrm{copy}}-\mathcal R_R
 &=\|\E\Delta_h\|^2,\label{eq:residual-copy-gain}\\
 \mathcal R_{\mathrm{mean}}-\mathcal R_R
 &=2\operatorname{tr}\operatorname{Cov}(Y_h(a),Y_h(a'))
   -\operatorname{tr}\operatorname{Var}(Y_h(a)).
 \label{eq:residual-mean-gain}
\end{align}
For the second line, expand
$\operatorname{Var}(Y_h(a')-Y_h(a))$ in
\eqref{eq:residual-contrast-risk}. Thus the population residual rule always
weakly improves on factual copying in this space, by exactly the squared
conditional mean action effect.
It improves on the no-outcome mean when the shared-event covariance is large
enough. In an additive response family, the latter gain is the disturbance
variance and residual risk is zero. In general the second gain can be
negative: reusing a negatively correlated or greatly amplified factual
residual can be worse than ignoring it. These are population identities in the
modeled coordinates, not guarantees for a fitted mean or after a nonlinear
physical readout. For a fitted mean, the nonnegative contrast-estimation term
below reduces the gain over factual copying, and a comparison against its own
fitted no-outcome prediction must also account for that prediction's
mean-estimation error.

\paragraph{Fitted-mean error.}Let a trained mean
be fixed independently of the scored events, and write
$e_h=\widehat\mu_h-\mu_h$ and
$\widetilde R_h(a)=R_h(z,a,\varepsilon)-\E_{\nu_z}R_h(z,a,\varepsilon)$.
Conditional on its training data, the exact error is
\begin{equation}
 \widehat T_R-Y_h(a')
 =\widetilde R_h(a)-\widetilde R_h(a')+e_h(z,a')-e_h(z,a).
 \label{eq:residual-estimation-decomposition}
\end{equation}
For fixed $z,a,a'$ the first term is centered, which gives the exact
bias--variance decomposition already used in
\eqref{eq:duration-data-risk-decomposition}:
\begin{equation}
 \E_{\nu_z}\|\widehat T_R-Y_h(a')\|^2
 =\operatorname{tr}\operatorname{Var}_{\nu_z}(\Delta_h)
  +\|e_h(z,a')-e_h(z,a)\|^2.
 \label{eq:residual-estimation-risk-appendix}
\end{equation}
Common mean errors cancel; unrelated errors at the two actions do not.
To claim $O(h^4)$ squared error for a learned predictor rather than the
population rule, its mean-contrast error must itself be $O(h^2)$ in root mean
square. Consistency at a fixed $h$ does not supply a uniform joint limit in
sample size and $h$. This distinction matters when feature
normalization depends on the time step; the stated rate is in fixed encoder coordinates.

\paragraph{Fixed disturbance scaling.}The short-time limit also
fixes the disturbance law and action bounds. Increasing disturbance amplitude
as $h$ shrinks can make the remainder constant depend on $h$ and remove the
claimed rate. For example, replacing $\varepsilon$ by $\xi/h$ in the
squared-encoder example of \cref{app:counterexamples} changes its residual MSE from order $h^4$ to
order $h^2$ for fixed nondegenerate $\xi$. Strong-noise benchmark
qualification and short-time approximation are separate requirements; neither
supplies the other (recall
\cref{prop:integrator-benchmark-limit,prop:small-step-baselines}).

\paragraph{Readout error bound.}Let
$Q$ denote the query population and
$\eta_h=\|e_h(Z,A')-e_h(Z,A)\|_{L^2(Q)}$, and suppose the frozen readout
$\pi$ is $L_\pi$-Lipschitz on a region containing the true and predicted
embeddings. If
$\kappa_h=\E_Q\|\pi(Y_h(A'))-s^{\star\prime}\|^2$, where $s^{\star\prime}$
denotes the true scored coordinates under the query action, then
\begin{equation}
 \|\pi(\widehat T_R)-s^{\star\prime}\|_{L^2(Q)}
 \le L_\pi\sqrt{4C^2h^4+\eta_h^2}+\sqrt{\kappa_h}.
 \label{eq:residual-proprio-bound}
\end{equation}
The bound follows from \eqref{eq:residual-estimation-risk}, Lipschitzness and
Minkowski's inequality. For a pooled $R^2$ denominator
$V_Q=\E_Q\|s^{\star\prime}-\E_Qs^{\star\prime}\|^2>0$, it implies
\begin{equation}
 R^2_{\mathrm{CF}}\ge
 1-\frac{\bigl(L_\pi\sqrt{4C^2h^4+\eta_h^2}+\sqrt{\kappa_h}\bigr)^2}{V_Q}.
\end{equation}
Apply this separately to each coordinate for a macro-average of coordinate
$R^2$ values. A high factual readout score does not establish Lipschitzness
off the image manifold or a small $\kappa_h$ on a shifted CF population.
Nor does the absolute $O(h^4)$ squared-error rate directly imply an $R^2$
rate: $V_Q$ may also shrink with $h$. With fixed-source disturbance variation
of order $h^2$, the corresponding normalized error can be only $O(h^2)$.
Subtracting the same observed factual endpoint from both prediction and truth
does not change squared prediction error; outcome-error and ITE-error MSE
are then algebraically identical, though differently normalized scores need
not agree.

\paragraph{Nonlinear readouts.}
$\pi(\mu_h(z,a))$ need not equal $\E[\pi(Y_h(a))\mid z]$, even if the
embedding mean is known exactly. The squared-loss theorem for embeddings
(\cref{prop:irreducible}) therefore does not make the decoded deterministic
predictor the squared-error optimum in physical coordinates, and we report the
oracle predictor (no outcome) separately (\cref{tab:main-cf}). For a simple algebraic
example, let $Z(a)=m(a)+\varepsilon$ with
$\varepsilon\in\{-1,+1\}$ equiprobable and let $\pi(z)=z^2$.
Then $\pi(\E Z(a))=m(a)^2$, while $\E\pi(Z(a))=m(a)^2+1$.
An exact embedding mean therefore retains a decoded mean error, although
residual reuse recovers the same-noise $Z(a')$ and hence its decoded outcome
exactly. This example concerns the loss/readout distinction, not an observed
benchmark result. Likewise, a readout can discard non-additive feature
variation, so accurate decoded counterfactuals do not establish additivity of
the full embeddings, and near-additive physical coordinates alone do not
verify \eqref{eq:short-time-expansion} in the chosen feature representation.

Correction after the readout (\cref{eq:counterfactual-jepa-chart}) instead
uses $\pi\circ\widehat\mu$ as its physical predictor. Its action-contrast
error includes the mismatch between this decoded predictor and the physical
conditional mean. The feature-space bound above does not directly bound this
different construction.

\paragraph{Feature-space counterexample.}Let the physical outcome be $x(a)=m(a)+\varepsilon$ with bounded
centered noise, and use the fixed smooth encoder $\mathcal E(x)=(x,Kx^2)$ and
readout $\pi(z)=z_1$. Residual reuse with the exact conditional embedding means
has encoded error
\[
 T_R-\mathcal E(x(a'))
 =\bigl(0,\,2K[m(a)-m(a')]\varepsilon\bigr).
\]
The physical readout is exact for every realization, while feature MSE is
$4K^2[m(a)-m(a')]^2\operatorname{Var}(\varepsilon)$ and can be arbitrarily
large. This establishes that the two observations are mathematically
compatible; it does not establish that the pretrained representations in our
experiments implement this construction.

\subsection{Counterexamples}
\label{app:counterexamples}
The first two are the multiplicative-fault and aliasing exceptions of
\cref{subsec:short-time-residual} (the gain-fault stress tests of
\cref{sec:stress} measure the former); the third shows that a smooth encoder
attains the stated order even for exactly additive physics.

\paragraph{Multiplicative disturbance.}In one
dimension, $\dot s=a\varepsilon$ is smooth and can use an IID,
action-independent, centered disturbance draw. Nevertheless
$Y_h(a)=s+ha\varepsilon$, $\mu_h(s,a)=s$, and residual reuse errs by
$h(a-a')\varepsilon$. Its MSE is
$h^2(a-a')^2\operatorname{Var}(\varepsilon)$, rather than $O(h^4)$.
This model violates the action-independent instantaneous-effect assumption,
not the exogeneity of the random draw.

\paragraph{State aliasing.}Let the physical
state be $(x,v)$ with hidden $v\in\{-1,+1\}$ equally likely,
let $\dot x=av+\varepsilon$, $\dot v=0$, and observe only $z=x$.
All maps are smooth and the disturbance effect is additive and action-independent.
With centered noise and actions independent of $v$, the encoded conditional
mean is $\mu_h(z,a)=z$. Residual reuse then errs by $h(a-a')v$: the source
features fail the required sufficiency for the action response. Recovering
some visible coordinates with a probe does not exclude this case.

\paragraph{Encoder curvature.}For
$\dot s=a+\varepsilon$, $\E\varepsilon=0$, and $\mathcal E(s)=s^2$ on a
region where the source is determined by its features, direct calculation gives
\[
 Y_h(a)=[s+h(a+\varepsilon)]^2,\qquad
 T_R-Y_h(a')=2h^2(a-a')\varepsilon.
\]
Thus the $O(h^4)$ MSE order is attained even with exactly additive physical
increments. A smooth encoder need not make the finite-step response exactly
additive. Neural encoders and rasterized observations must satisfy the local
regularity assumptions; their pretraining objective is not a substitute.

\subsection{Minimax Lower Bound}
\label{app:minimax-rate}
\paragraph{Model class and information.}Fix $C>0$, $h_0>0$, the action domain
$\mathcal A=[-1,1]$ and a source domain $\mathcal Z\subset\R^3$ containing
$z_0=0$. Let $\mathcal F_C$ contain every transition family with, uniformly over
supported sources and actions and $0<h\le h_0$,
\begin{equation}
 F^M_h(z,a,\varepsilon)=z+h\,v_M(z,a)+h\,w_M(z,\varepsilon)+R^M_h(z,a,\varepsilon),
 \qquad \|R^M_h(z,a,\varepsilon)\|_{L^2(\nu^M_z)}\le Ch^2,
 \label{eq:minimax-class}
\end{equation}
with square-integrable terms, first-order maps $v_M,w_M$ that do not depend on
$h$, a disturbance law $\nu^M_z$ common across actions and durations, and the
same disturbance held across the potential outcomes of one event. Let
$\mathcal O_C\subset\mathcal F_C$ be its smooth autonomous ODE subclass,
$\dot z=v_M(z,a)+w_M(z,\varepsilon)$ with action and disturbance held within the
transition, which is \eqref{eq:short-time-physical-ode} with an identity
encoder. An estimator receives the indexed family of exact factual laws
$\mathcal P_M=\{\mathrm{Law}_M(F^M_\tau(z,a,\varepsilon)):z\in\mathcal Z,\
a\in\mathcal A,\ 0<\tau\le h_0\}$, which is more than any finite unpaired dataset
provides, the observed outcome $Y^M_-=F^M_h(0,-1,\varepsilon)$ of the query
event, and independent randomization $W$; it must predict
$Y^M_+=F^M_h(0,1,\varepsilon)$. These are separate transition laws, not joint
observations of one disturbance at several actions or durations. For
$\mathcal G\in\{\mathcal F_C,\mathcal O_C\}$ let
$\mathcal R^*_h(\mathcal G)=\inf_\delta\sup_{M\in\mathcal G}
\E_M\|\delta(\mathcal P_M,Y^M_-,W)-Y^M_+\|^2$. We prove \eqref{eq:minimax-rate}
for both classes. It implies $\lim_{h\downarrow0}\mathcal R^*_h(\mathcal G)/h^4=4C^2$
and $1\le\sup_M\E_M\|T^M_R-Y^M_+\|^2/\mathcal R^*_h(\mathcal G)\le1+C^2h^2$, and
it holds for the individual treatment effect as well, since subtracting the
observed $Y^M_-$ from the estimate and the target leaves the error unchanged.

\begin{proof}
\emph{Upper bound.} With $\widetilde R_h(a)=R_h(0,a,\varepsilon)-\E R_h(0,a,\varepsilon)$,
population residual reuse errs by exactly $T_R-Y_+=\widetilde R_h(-1)-\widetilde R_h(1)$;
centering is an $L^2$ contraction, so $\E\|T_R-Y_+\|^2\le4C^2h^4$, as in
\cref{thm:short-time-residual}.

\emph{Two smooth ODEs.} Let $J=\left(\begin{smallmatrix}0&-1\\1&0\end{smallmatrix}\right)$,
so $J^\top=-J$, $J^2=-I_2$ and $J^\top J=I_2$, and let $\varepsilon$ be uniform on
the disk of radius $\sqrt2$ in $\R^2$, which is bounded, rotation invariant,
centered and has $\E\|\varepsilon\|^2=1$. For a sign $\theta\in\{-1,+1\}$ unknown
to the estimator, take the observed state $(q,x)\in\R\times\R^2$ and
\begin{equation}
 \dot q=a,\qquad \dot x=(I_2+2\theta CqJ)\varepsilon,
 \label{eq:minimax-odes}
\end{equation}
smooth autonomous ODEs with separate channels $v((q,x),a)=(a,0)$ and
$w_\theta((q,x),\varepsilon)=(0,(I_2+2\theta CqJ)\varepsilon)$. From any source
$(q_0,x_0)$ the exact time-$h$ solution is
$F^\theta_h((q_0,x_0),a,\varepsilon)=\bigl(q_0+ha,\ x_0+h[I_2+\theta C(2q_0+ah)J]\varepsilon\bigr)$,
whose Taylor remainder is exactly $R^\theta_h=(0,\theta Cah^2J\varepsilon)$ with
$\|R^\theta_h\|_{L^2}=C|a|h^2\le Ch^2$. Both worlds therefore belong to
$\mathcal O_C\subset\mathcal F_C$, for all durations, without hidden state,
observation noise, contact or aliasing.

\emph{All factual laws coincide.} For every real $u$,
$(I_2+\theta uJ)(I_2+\theta uJ)^\top=(1+u^2)I_2$, so
$(I_2+\theta uJ)/\sqrt{1+u^2}$ is a rotation, and rotation invariance gives
$(I_2+uJ)\varepsilon\overset{d}{=}(I_2-uJ)\varepsilon\overset{d}{=}\sqrt{1+u^2}\,\varepsilon$.
With $u=C(2q_0+ah)$, both signs give the same factual law for every source,
action and duration, so even the full family $\mathcal P_M$ cannot reveal
$\theta$.

\emph{The counterfactuals disagree.} At source $0$ with factual action $-1$, let
$k=Ch$ and write the observed outcome as $Y^\theta_-=(-h,X)$ with
$X=h(I_2-\theta kJ)\varepsilon$, whose law is common to both signs and which has
$\E\|X\|^2=h^2(1+k^2)$. Since $(I_2-\theta kJ)^{-1}=(I_2+\theta kJ)/(1+k^2)$, the
same disturbance under the alternative action $1$ gives
$Y^\theta_+=(h,\ \alpha X+\theta\beta JX)$ with $\alpha=(1-k^2)/(1+k^2)$ and
$\beta=2k/(1+k^2)$. Knowing $\theta$ would identify the counterfactual exactly;
the obstruction is the unknown coupling.

\emph{Two-point lower bound.} Let $g_\theta(X)=(h,\alpha X+\theta\beta JX)$ and
$\bar g(X)=(h,\alpha X)$. For every output $d\in\R^3$,
$\tfrac12\bigl(\|d-g_{+1}(X)\|^2+\|d-g_{-1}(X)\|^2\bigr)
=\|d-\bar g(X)\|^2+\beta^2\|X\|^2$. All information given to the estimator,
including any randomization or further unpaired data, has a common law under the
two worlds; applying the identity to its output and integrating, the larger of
its two risks is at least $\beta^2\E\|X\|^2=4C^2h^4/(1+C^2h^2)$, which the
midpoint $\bar g$ attains. Both worlds lie in each class, which gives the lower
bound for $\mathcal F_C$ and $\mathcal O_C$.

\emph{Residual reuse attains the upper bound.} The common means give
$T_R=(h,X)$, hence $T_R-Y^\theta_+=(0,-2\theta Ch^2J\varepsilon)$ and
$\E_\theta\|T_R-Y^\theta_+\|^2=4C^2h^4$.
\end{proof}

\paragraph{Interior actions.}If the query pair is instead
drawn from any law $\eta$ on $[-1,1]^2$ independent of the current disturbance,
with $b_\eta=\E_\eta(A'-A)^2>0$, and both actions are given to the estimator, the
same two ODEs give conditional counterfactuals
$\bigl[\tfrac{1+C^2h^2aa'}{1+C^2h^2a^2}I_2+\theta\tfrac{Ch(a'-a)}{1+C^2h^2a^2}J\bigr]X$
with $\E[\|X\|^2\mid A=a]=h^2(1+C^2h^2a^2)$, and the same argument bounds the
action-averaged minimax risk below by $C^2b_\eta h^4/(1+C^2h^2)$, while residual
reuse stays below $4C^2h^4$; the rate is again $\Theta(h^4)$, without a claim on
the sharp constant for this averaged criterion.

\begin{corollary}[Relative minimax rate]
\label{cor:minimax-relative}
Restrict either class to the worlds whose first-order disturbance effect at
source $0$ has unit variance, $\E\|w_M(0,\varepsilon)-\E w_M(0,\varepsilon)\|^2=1$,
and normalize each world's risk by $V_h(M)=\operatorname{tr}\operatorname{Var}_M Y^M_+$.
For $Ch<1$ the normalized minimax risk $\mathcal Q^*_h$ satisfies
\[
 \frac{4C^2h^2}{(1+C^2h^2)^2}\le\mathcal Q^*_h
 \le\sup_M\frac{\E_M\|T^M_R-Y^M_+\|^2}{V_h(M)}\le\frac{4C^2h^2}{(1-Ch)^2},
\]
so the relative $h^2$ rate of \cref{cor:relative-rate} cannot be uniformly
improved.
\end{corollary}
\begin{proof}
Centering the expansion and using the reverse triangle inequality and the $L^2$
contraction of centering gives $\sqrt{V_h(M)}\ge h-Ch^2$, hence the upper bound.
The hard pair lies in the normalized class and has common variance
$h^2(1+C^2h^2)$; dividing its absolute lower bound by that variance gives the
lower bound.
\end{proof}

\paragraph{Scope.}The theorem concerns the population rule. For a trained
predictor $\widehat\mu_h=\mu_h+e_h$, \eqref{eq:residual-estimation-risk} adds
$\E\|e_h(0,1)-e_h(0,-1)\|^2$, so a uniform $O(h^4)$ bound on that term preserves
the rate and a uniform $o(h^4)$ bound the sharp constant; no sample-size,
optimization or readout guarantee follows. The optimality is asymptotic and
worst case: on the hard pair the midpoint beats residual reuse at every finite
$h$, by $4C^4h^6/(1+C^2h^2)$, so other transports can be better on particular
systems; what is excluded is a uniformly smaller leading constant or a uniformly
$o(h^4)$ error over the class. The information model allows any amount of
unpaired data generated with fresh disturbances, but not repeated observations of
the same disturbance within the query event, paired alternatives or a known
coupling, each of which can supply the missing information. Finally, the bound
covers the smooth additive-channel class of \eqref{eq:short-time-physical-ode};
multiplicative faults, contacts and aliasing remain outside it, and so does the
pose-only rate of \cref{cor:inertial-residual}. More restricted scalar, monotone
or known-mechanism classes can be counterfactually identifiable.

\clearpage

\section{Physical Calibration and Numerical Checks}
\label{app:independent-benchmark-design}
\begin{table}[htbp]\centering\small
\begin{tabular}{llrr}\toprule
Environment & Coordinate & Oracle (no outcome) $R^2$ & Factual-copy $R^2$ \\ \midrule
HandReach & wrist WRJ1 & 52.3 [49.9, 54.5] & 7.2 [-3.3, 17.0] \\
HandReach & wrist WRJ0 & 47.8 [45.1, 50.4] & 7.4 [-7.3, 21.7] \\
Ball-in-Cup & cup $x$ & 51.6 [50.4, 52.9] & 5.2 [-1.0, 11.2] \\
Ball-in-Cup & cup $z$ & 50.6 [49.2, 52.0] & 8.2 [2.3, 13.6] \\
FetchReach & end-effector $x$ & 51.8 [49.3, 54.0] & 48.1 [42.5, 53.2] \\
FetchReach & end-effector $y$ & 49.0 [46.5, 51.3] & 40.2 [34.9, 46.2] \\
FetchReach & end-effector $z$ & 49.2 [46.6, 51.4] & 42.6 [36.6, 47.8] \\
Racetrack & heading & 54.5 [51.8, 56.8] & 28.2 [21.4, 35.4] \\
Racetrack & lane offset & 46.8 [44.8, 48.6] & 16.0 [9.2, 23.6] \\
BlueROV2 & $\sin$(pitch) & 51.8 [50.5, 53.0] & 1.4 [-4.9, 7.2] \\
PassWater & glass $x$ & 50.2 [47.3, 52.6] & 8.1 [-5.3, 19.1] \\
Push-T & pusher $x$ & 50.8 [49.7, 51.9] & 2.0 [-3.4, 8.3] \\
Push-T & pusher $y$ & 51.4 [50.2, 52.5] & 5.4 [-0.6, 11.0] \\
\bottomrule\end{tabular}
\caption{\textbf{Physical confirmation results.}
Percent $R^2$ at the fixed continuing $t=2$ anchor, with marginal 95\% whole-parent
intervals; no learned model or image readout. Racetrack, Push-T and PassWater use
independently checked analytic conditional moments; the other rows use native
conditional Monte Carlo estimates with separate native quadrature checks but
no formal numerical-error bound. PassWater has only four native disturbance
draws per source state, below our design minimum. WRJ1 and WRJ0 are the two
wrist joints of the Shadow hand, as named in its MuJoCo model. These simulator checks test
neither image recovery (\cref{app:pinned-protocol}) nor the unlisted state
coordinates.}
\label{tab:physical-design-confirmation}
\end{table}

\paragraph{Physical selection.}No encoder,
probe, residual predictor or transport score selects the benchmark physics. A finite
physical search declares the reset law, task-directed behavior policy,
coordinate groups and uniformly random factual/query commands before scoring,
and new parents then confirm the selected condition, with every terminal case
retained. The scored source is the same $t=2$ anchor after two genuine native
transitions; changing a future model's history length must not change this
population.

\paragraph{Fresh-disturbance reference.}The environment
configuration exposes the fresh (IID) share of the Gaussian driver's variance,
its temporal correlation length, action gain, bounded physical disturbance scale and number of native
controls held per transition; the native integration step is unchanged. The
benchmark uses a fresh share of one, so its current driver is
independent of the complete physical source state and history. Conditional
integration over the fresh driver is therefore an information-matched
full-state $k=1$ oracle (no outcome) reference, neither a trained predictor nor a prediction
at zero disturbance. With persistent disturbances this full-state reference
no longer matches the information of an image-history predictor; the reference
used there is described in \cref{app:pinned-protocol}.

\paragraph{Calibration targets and replay.}For
every prospectively retained coordinate, the target is a marginal 95\%
parent-bootstrap interval of the oracle (no outcome) $R^2$ inside $[.42,.58]$ and a factual-outcome-copy
upper interval strictly below $.60$. Numerical error enlarges both intervals,
which are marginal, not a simultaneous all-environment confidence claim. We also
measure same-disturbance action influence directly as
$\E[(s^{\rm cf}_j-s^f_j)^2]/\operatorname{Var}(s^{\rm cf}_j)$, which equals
one minus factual-copy $R^2$ on the same population. Independent native
replays check complete simulator state, intervening query steps, branch order,
command/gain equivalence, held-control duration and episode stopping, with zero
measured discrepancy for the selected confirmation cohorts.

\paragraph{Status of the numerical checks.}The references of Racetrack
(heading and lane offset) and Push-T (pusher position) are analytic, with
formal numerical-error bounds (below). HandReach, Ball-in-Cup, FetchReach and
PassWater pass the image recovery criteria of \cref{app:pinned-protocol} with
DINOv2-S, DINOv2-B and MAE-B, as does BlueROV2 with the disturbance readout of
\cref{app:bluerov-gate}; Racetrack and Push-T fail them and serve as
diagnostics. For HandReach, Ball-in-Cup, FetchReach and BlueROV2, independent
numerical refinements of the Monte Carlo reference agree, but we have no formal
bound on its numerical error. PassWater has a bounded glass-position reference
and full native replay, but only four native disturbance draws per source
state, below our design minimum of 64.

\paragraph{Scope.}
Calibration uses simulator outcomes and optimizes neither a learned model nor
a difference between methods. The suite uses bounded actuator and position-goal
disturbances and scores observable actuated coordinates. These choices can
favor approximately affine responses; they do not represent arbitrary contact
or fluid dynamics. The $60\%$ baseline thresholds define benchmark eligibility.

\paragraph{Disturbance process.}The
declared Gaussian driver and bounded physical offset are
\begin{equation}
 g_t=\sqrt{1-\alpha}\,b(t/\ell)^\top\xi+
       \sqrt{\alpha}\,\eta_t,\qquad
 \varepsilon_t=\sigma\tanh(g_t),
 \label{eq:pinned-driver}
\end{equation}
where $\xi$ is an episode-specific standard Gaussian coefficient vector,
$\eta_t$ are independent standard Gaussian innovations, and $b$ is a fixed
unit-norm Fourier basis. Its covariance is the actual finite-basis
covariance $(1-\alpha)b(t/\ell)^\top b(t'/\ell)+\alpha\mathbf1_{t=t'}$;
we do not replace it by an ideal radial basis function kernel. The fresh share $\alpha$ refers
to the Gaussian driver's variance, not the variance of its nonlinear bounded
image, and at $\alpha=1$ the temporal scale $\ell$ is inactive. A realized
physical offset is held fixed throughout a transition and across all
its factual/query branches. This models actuator/control disturbance, not
hydrodynamic wind or current.

The action gain, offset bound $\sigma$, and number of held native steps are
fixed independently of the later fresh-share and history sweeps. The selected
triples (gain, bound, held steps) are HandReach $(1.145558,\allowbreak .473978,\allowbreak 4)$,
Ball-in-Cup $(1.472860,\allowbreak .473978,\allowbreak 16)$, FetchReach $(.939167,\allowbreak .54,\allowbreak 8)$,
Racetrack $(.308,\allowbreak .112,\allowbreak 1)$, BlueROV2 $(1.472860,\allowbreak .473978,\allowbreak 20)$,
PassWater $(1.088889,\allowbreak .5,\allowbreak 22)$, and Push-T $(1.472860,\allowbreak .473978,\allowbreak 4)$,
and the configuration files of the code release retain full precision. Native integration time
steps are unchanged; these held durations are not equal physical times across
environments and are not evidence of the asymptotic small-time regime of
\cref{thm:short-time-residual}. FetchReach and PassWater use position-goal
offsets, Racetrack a lane-goal offset and Push-T its pusher goal offset;
HandReach, Ball-in-Cup and BlueROV2 use their declared actuator/thruster interfaces.

\paragraph{Collection and branching.}Each
independent episode starts from the declared native reset law and runs two
actual disturbed policy transitions before the anchor. Policies are the
HandReach inverse-kinematics controller, a Ball-in-Cup policy pretrained with deep deterministic policy gradient (DDPG),
the FetchReach reach controller, a Racetrack lane follower, the BlueROV2 goal
controller, the PassWater water-transport controller and a pretrained Push-T
diffusion policy, with the declared exploration. At the
anchor the factual command is independently uniform over the fixed support, and
evaluation uses the same factual-command law and three independent uniform query
commands. No equal-action reconstruction row enters a CF score, and no difficult
contact, terminal or small-action case is removed to improve a score; terminal
handling follows the declared absorbing convention.

Training collection writes only the single factual continuation; branching from
the identical complete simulator snapshot and reusing the realized offset occur
only in evaluation. Rendering is isolated by native replay, so refreshing
renderer kinematics cannot change a subsequent behavior command. Recorded prefix
states, commands, disturbances and endpoints are checked against the independent
evaluation sampler. State labels remain private to calibration, replay checks and
evaluation-head fitting; the world predictor sees only frozen frame embeddings
and observed commands. The BlueROV2 camera change (\cref{app:bluerov-gate}) affects observation geometry only,
and verifies that its factual state/action/disturbance arrays are unchanged.

\paragraph{Numerical references.}Push-T's
kinematic pusher obeys an affine position/velocity/goal recurrence; the
independent formula matches all 262,144 confirmation endpoints within
$3.42\times10^{-13}$ pixels, and its $R^2$ numerical bound is $1.84\times10^{-5}$.
This concerns the pusher, not the contacted block. The frozen behavior policy
improves block-goal distance on confirmation, while its block-overlap interval on
that subset includes zero. Racetrack composes the native lane controller and
bicycle map on the initial straight. Support checks cover controller saturation,
each lateral integration substep, lane boundaries, and separation from the
retained traffic vehicle over all current disturbance values. Monotone
Gaussian-bin enclosures give a confirmation $R^2$ error bound of $.002697$, and
native endpoint agreement is within $2.85\times10^{-14}$. These bounds include
conservative floating-point padding, not formal verification of numerical
libraries. The PassWater coordinate is glass position; native fluid motion is
present but not the checked output. Its task controller improves transport
distance while retaining $99.8845\%$ of the water, versus $99.9784\%$ for uniform
commands. None of these restricted results establishes additive block-contact,
fluid-particle or image-embedding dynamics.

\paragraph{Restricted-coordinate oracle.}Let
$I_k$ contain exactly the last $k$ retained physical-coordinate vectors, the
intervening observed commands and the query command; this differs from fixing
the complete native simulator state. The recovery mask defines the scored
outputs, not all information contained in an image embedding: features may retain
useful information about unselected physical coordinates. Thus the exact
$I_k$-conditional optimum is not automatically an upper bound for an
image-history predictor unless the required information inclusion is
established. Likewise, a complete-state reference upper-bounds achievable $R^2$
only when its information, possibly augmented by the observed history, is
sufficient for the queried outcome; in particular, IID current noise does not by
itself make omitted source velocities irrelevant. For a sufficient larger
information set $J$ and any fixed $I_k$-measurable predictor $g$,
conditional-expectation projection yields
\begin{equation}
 \E\|Y-\E[Y\mid J]\|^2
 \le \inf_{f\;I_k\text{-measurable}}\E\|Y-f(I_k)\|^2
 \le \E\|Y-g(I_k)\|^2.
 \label{eq:retained-oracle-bracket}
\end{equation}

For the fully fresh Push-T pusher, the independently checked recurrence of the native
proportional--derivative (PD) controller is $Y=q p+c_v v+c_g(256+216\,\gamma a')+216c_g\epsilon(1,-1)$.
The admissible position-only predictor drops $c_vv$, and its excess risk relative
to the complete-state conditional mean is exactly $c_v^2\E\|v\|^2$; current
velocity is used to evaluate this bound, never as an input to that predictor.
This encloses the optimum for all three position-history lengths. On the fixed
confirmation parents the coordinatewise $R^2$ brackets, including the
Gaussian-moment enclosure and conservative arithmetic padding, are
$[50.7973,50.8008]\%$ for pusher x and $[51.3517,51.3552]\%$ for pusher y.
These narrow numerical/information brackets are separate from the wider
whole-parent sampling intervals, and say nothing about the block dynamics or
the failed disturbance recovery from images.

For fully fresh PassWater, current glass position is sufficient for the retained
coordinate: the audited scalar recurrence gives the conditional mean
$\rho x+(1-\rho)(.575+.825\,\gamma a')$ exactly, and extra history cannot improve
that conditional mean under IID. Its confirmation oracle (no outcome) $R^2$, given the retained state,
lies in the numerical bracket $[50.1968,50.2258]\%$, obtained from the
independently checked monotone Gaussian-bin enclosure of $\E\tanh(G)^2$
and a propagated ratio-error bound; the separate outer-parent 95\% envelope
is $[47.32,52.65]\%$. This assumes the stated conservative floating-point
padding and is not machine-verified interval arithmetic. PassWater still has
fewer native disturbance draws than our design minimum, so its sampling
intervals remain wide.
Neither bound transfers to temporally correlated disturbances,
which require integrating the disturbance law conditional on the permitted
history. Restricted-coordinate oracles for HandReach, Ball-in-Cup, FetchReach
and BlueROV2 are not computed; for these environments the reported oracle uses
the complete simulator state and is therefore a stronger-information reference.

\clearpage

\section{Proofs of the Main-Text Statements}
\label{app:main-proofs}
\subsection[Proof that Marginals Do Not Identify Matchings]{Proof of \cref{prop:marginals-not-counterfactual}}
\label{proof:marginals-not-counterfactual}
\begin{proof}
Let $U\sim\mathrm{Unif}(0,1)$. The worlds
$(Y(0),Y(1))=(U,U)$ and $(Y(0),Y(1))=(U,1-U)$ have identical uniform
marginals. Given $Y(0)=y$, their alternatives are $y$ and $1-y$.
\end{proof}

\subsection[Proof that Residual Reuse is the Only Exact Additive Rule]{Proof of \cref{prop:residual-unique}}
\label{proof:residual-unique}
\begin{proof}
In an additive world the counterfactual of the observed outcome $y=Y_h(a)$ is
$Y_h(a')=y+m_h(z,a')-m_h(z,a)$. Because the disturbance law does not depend on
the action, $\mu_h(z,a)=m_h(z,a)+\E_{\nu_z}q_h(z,\varepsilon)$, so
$m_h(z,a')-m_h(z,a)=\mu_h(z,a')-\mu_h(z,a)$. A rule that is exact in this world
must return $Y_h(a')$ for every observed $y$ in the support, that is,
$y+\mu_h(z,a')-\mu_h(z,a)$.
\end{proof}

\subsection[Proof of the Interaction Bound]{Proof of \cref{thm:interaction-bound}}
\label{proof:interaction-bound}
\begin{proof}
Let $\bar\varepsilon=\E_{\nu_z}\varepsilon\in\mathcal U$ (closed and convex),
$\varepsilon_t=\bar\varepsilon+t(\varepsilon-\bar\varepsilon)$ and
$a_s=a+s(a'-a)$. By the fundamental theorem of calculus, first in
$\varepsilon$ and then in $a$,
\[
 \Delta(\varepsilon)-\Delta(\bar\varepsilon)
 =\int_0^1\!\!\int_0^1\partial_a\partial_\varepsilon
 F(z,a_s,\varepsilon_t)\,[a'-a,\ \varepsilon-\bar\varepsilon]\,ds\,dt,
\]
so $\|\Delta(\varepsilon)-\Delta(\bar\varepsilon)\|\le K_z\|a'-a\|\,
\|\varepsilon-\bar\varepsilon\|$. A variance is at most the second moment
about any constant, which gives \eqref{eq:interaction-bound} by
\eqref{eq:residual-contrast-risk}. For an independent copy $\varepsilon'$,
$\operatorname{tr}\operatorname{Var}Y(a')=\tfrac12\E\|F(z,a',\varepsilon)
-F(z,a',\varepsilon')\|^2\ge\tfrac12\kappa_z^2\E\|\varepsilon-\varepsilon'\|^2
=\kappa_z^2\operatorname{tr}\operatorname{Cov}(\varepsilon)$.
\end{proof}

\subsection[Proof of the Short-Time Rate]{Proof of \cref{thm:short-time-residual}}
\label{proof:short-time-residual}
\begin{proof}
Let $\widetilde R_h(a)=R_h(z,a,\varepsilon)-\E_{\nu_z}R_h(z,a,\varepsilon)$.
Substituting \eqref{eq:short-time-expansion} into \eqref{eq:residual-transport}
cancels both $v$ and the shared $w-\E_{\nu_z}w$, leaving
$T_R-Y_h(a')=\widetilde R_h(a)-\widetilde R_h(a')$. Centering is an orthogonal
projection in $L^2$, so each centered remainder has norm at most $Ch^2$; the
triangle inequality proves \eqref{eq:short-time-residual-rate}. Replacing
$\mu_h$ by $\widehat\mu_h$ adds exactly $e_h(z,a')-e_h(z,a)$.
\end{proof}

\subsection[Proof of the Relative Rate]{Proof of \cref{cor:relative-rate}}
\label{proof:relative-rate}
\begin{proof}
For an independent copy $\varepsilon'$, the terms $z+hv(z,a')$ cancel in
$Y_h(a')-Y_h'(a')$, so in $L^2$,
$\|Y_h(a')-Y_h'(a')\|\ge h\kappa\|\varepsilon-\varepsilon'\|-\|R_h-R_h'\|
\ge\sqrt2\,h\kappa\sigma-2Ch^2$. Hence
$\operatorname{tr}\operatorname{Var}Y_h(a')\ge(h\kappa\sigma-\sqrt2\,Ch^2)^2$,
and \eqref{eq:short-time-residual-rate} bounds the numerator by $4C^2h^4$.
\end{proof}

\subsection[Conditional-mean projection proof]{Proof of \cref{prop:irreducible} and
\cref{cor:frozen-encoder-mean}}
\label{proof:irreducible}

\begin{proof}
Write $X=X_t$, $Y=Y_t$, and $\mu=\mathbb{E}[Y\mid X]$. For any admissible
$g$, add and subtract $\mu(X)$ and expand:
\begin{align}
\mathbb{E}\|Y-g(X)\|^2
&=\mathbb{E}\|Y-\mu(X)\|^2
  +\mathbb{E}\|\mu(X)-g(X)\|^2 \notag\\
&\quad+2\,\mathbb{E}\!\left[
  \langle Y-\mu(X),\mu(X)-g(X)\rangle\right].
\end{align}
The cross term is zero by the tower property because $\mu(X)-g(X)$ is
$\sigma(X)$-measurable and
\[
    \mathbb{E}[Y-\mu(X)\mid X]=0.
\]
This proves \eqref{eq:mse-pythagorean}. Its second term is nonnegative and
vanishes exactly when $g(X)=\mu(X)$ almost surely, proving uniqueness on the
training support. Differentiating the pointwise squared loss with respect to its
prediction gives $2(g(X)-Y)$; substituting $g=\mu$ and conditioning on $X$
proves \eqref{eq:mse-unbiased-score}. Finally,
\[
    \mathbb{E}\|Y-\mu(X)\|^2
    =\mathbb{E}\!\left[\operatorname{tr}\operatorname{Var}(Y\mid X)\right],
\]
which proves \eqref{eq:irreducible-error}.

For the pure-covariate-shift claim, keep the joint law of $(X,Y)$ fixed and
multiply the same pointwise expansion by the $\sigma(X)$-measurable $w(X)$.
The cross term remains zero:
\[
\mathbb E\!\left[w(X)
  \langle Y-\mu(X),\mu(X)-g(X)\rangle\right]
=\mathbb E\!\left[w(X)
  \langle\mathbb E[Y-\mu(X)\mid X],\mu(X)-g(X)\rangle\right]=0.
\]
The excess weighted risk is therefore
$\mathbb E[w(X)\|g(X)-\mu(X)\|^2]$, which vanishes only when
$g(X)=\mu(X)$ almost surely because $w(X)>0$.  This proves
\eqref{eq:mse-reweighting}.  If the reweighting is zero at some input,
uniqueness is of course lost there; squared loss cannot identify an unsupported
action.  Nor does this calculation apply when a distribution change modifies
$P(Y\mid X)$, as an arbitrary policy change can do when $X$ aliases latent
state.

Nothing in the argument requires the context and target maps to be equal:
take $X=(H_t,\va_t)$ and $Y=f(\vo_{t+1})$, with $H_t$ any fixed measurable
context of observations and past actions. For a released encoder $f$ and a
fixed context construction, the population target of a newly trained
squared-loss predictor is therefore
$\mathbb{E}[f(\vo_{t+1})\mid H_t,\va_t]$, proving
\cref{cor:frozen-encoder-mean}.
\end{proof}

\paragraph{Policy-shift counterexample.} Let a latent state
$S\in\{-1,+1\}$ be uniform, let the frozen history feature $H$ be constant, and
set $Y=S$. Under one behavior policy, $A=S$ with probability $.9$ and $A=-S$
otherwise, so both actions have positive support and
$\mathbb E[Y\mid H,A=a]=.8a$. Under a second policy that chooses $A$
independently of $S$, $\mathbb E[Y\mid H,A=a]=0$. This is not an
$X=(H,A)$-measurable reweighting of one fixed joint law: the policy changes the
latent-state mixture at fixed $X$. The reweighting result therefore requires
pure covariate shift, invariant $P(Y\mid X)$, or a transition-sufficient
representation.

\subsection[Proof that Shared Inputs Rule Out Confounding]{Proof of \cref{prop:shared-inputs}}
\label{proof:shared-inputs}
\begin{proof}
Conditionally on $H_t=h$, the action $\va_t=\pi_b(g(h),\xi_t)$ is a measurable
function of $\xi_t$ alone, and $\xi_t$ is conditionally independent of
$(Y_t(a))_a$ given $H_t$; hence so is $\va_t$, which is conditional
exchangeability. For a supported action $a$, consistency then gives
$\mu(h,a)=\E[Y_t\mid H_t=h,\va_t=a]=\E[Y_t(a)\mid H_t=h,\va_t=a]
=\E[Y_t(a)\mid H_t=h]$.
\end{proof}

\subsection[Proof of the History Risk Order]{Proof of \cref{prop:history-risk-order}}
\label{proof:history-risk-order}
\begin{proof}
Conditional expectation is an orthogonal projection in $L^2$; apply the
Pythagorean identity to the nested projections of $Y$ and of $\Delta$, using
\eqref{eq:residual-contrast-identity}.
\end{proof}

\clearpage

\section{Nonidentification Proofs}
\label{sec:factual-gauge-revision}

Fix a state
$s$ and write $x=(s,a)$. Let $\nu$ be a reference law on an exogenous space
$\mathcal U$, and let $T_x:\mathcal U\to\mathcal Y$ be a bimeasurable bijection,
modulo null sets, such that $(T_x)_\#\nu=P_x$, where $P_x$ is the factual
conditional law of the next outcome. The model-internal counterfactual coupling
from $a$ to $a'$ is
\begin{equation}
  C_{a'\leftarrow a}(y)
  =T_{s,a'}\circ T_{s,a}^{-1}(y).
  \label{eq:model-cf-coupling}
\end{equation}

\begin{theorem}[Factual gauge nonidentification, full statement]
\label{thm:factual-gauge-nonidentification-full}
For every measurable family of $\nu$-preserving bijections
$G_x:\mathcal U\to\mathcal U$, the alternative mechanisms
\begin{equation}
  \widetilde T_x=T_x\circ G_x
  \label{eq:gauge-twin-map}
\end{equation}
induce exactly the same factual conditional laws $P_x$.  Their counterfactual
coupling is instead
\begin{equation}
  \widetilde C_{a'\leftarrow a}(y)
  =T_{s,a'}\circ G_{s,a'}\circ G_{s,a}^{-1}
    \circ T_{s,a}^{-1}(y).
  \label{eq:gauge-twin-cf}
\end{equation}
Consequently, factual conditional likelihood, calibration, or any other
criterion depending only on the family $\{P_x\}_x$ cannot identify the physical
same-shock counterfactual.  If $G_{s,a}=G_s$ is shared across actions, the gauge
cancels from \eqref{eq:gauge-twin-cf}; if it depends on the action, it generally
does not.  Conversely, every bimeasurable $\widetilde T_x$ with
$(\widetilde T_x)_\#\nu=P_x$ is of this form modulo null sets.
\end{theorem}

\begin{proof}
Because $(G_x)_\#\nu=\nu$,
$(\widetilde T_x)_\#\nu=(T_x)_\#(G_x)_\#\nu=(T_x)_\#\nu=P_x$.
Moreover,
$\widetilde T_x^{-1}=G_x^{-1}\circ T_x^{-1}$ modulo null sets.  Substitution in
$\widetilde T_{s,a'}\circ\widetilde T_{s,a}^{-1}$ gives
\eqref{eq:gauge-twin-cf}.  When $G_{s,a'}=G_{s,a}$ the middle maps cancel.  For
an action-dependent family they need not; for example, with
$\nu=\mathcal N(0,I_2)$, $T_{s,a}=\operatorname{Id}$, and
$G_{s,a}=R(ka)$, every factual arm remains $\mathcal N(0,I_2)$, whereas the
counterfactual is $y\mapsto R(k(a'-a))y$ rather than $y\mapsto y$.
For the converse, define $G_x=T_x^{-1}\circ\widetilde T_x$.  Then
$(G_x)_\#\nu=(T_x^{-1})_\#P_x=\nu$ and
$\widetilde T_x=T_x\circ G_x$ modulo null sets.
\end{proof}

\subsection{Individual Counterfactual Lower Bound}
\label{app:individual-cf-minimax-proof}
\begin{theorem}[Sharp minimax impossibility for one individual counterfactual]
\label{thm:individual-cf-minimax}
Let $U\sim\mathrm{Unif}(0,1)$ and consider the two structural models
$\mathcal M_+:(Y(0),Y(1))=(U,U)$ and $\mathcal M_-:(Y(0),Y(1))=(U,1-U)$.
Let $D_n=\{(A_i,Y_i)\}_{i=1}^n$ be an unpaired factual sample under any known
randomized design, including $n=\infty$, where the exact interventional laws
are given. For a query unit with $Y(0)=y$, let $\widehat Y_1=\delta(D_n,y,W)$ be
any possibly randomized estimate of $Y(1)$, and
$R_m(\delta;y)=\E_m[(\widehat Y_1-Y_m(1))^2\mid Y(0)=y]$. Then
\begin{align}
 \inf_\delta\max_m R_m(\delta;y)
   &=\left(y-\tfrac12\right)^2
   &&\text{for a.e. }y\in[0,1],
   \label{eq:individual-cf-pointwise-lower-bound}\\
 \inf_\delta\max_m\E_m[(\widehat Y_1-Y_m(1))^2]
   &=\frac1{12},
   \label{eq:individual-cf-integrated-lower-bound}\\
 \inf_\delta\max_m\operatorname*{ess\,sup}_{y\in[0,1]}R_m(\delta;y)
   &=\frac14,
   \label{eq:individual-cf-uniform-lower-bound}
\end{align}
all attained by $\widehat Y_1\equiv\tfrac12$.
\end{theorem}

\begin{proof}[Proof of \cref{thm:individual-cf-minimax}]
Under both models $\mathcal L(Y\mid A=a)=\mathrm{Unif}(0,1)$ for
$a\in\{0,1\}$.  Thus the complete observable experiment, including $D_n$ and
the factual query, has the same law in both models for every $n$.  Conditional
on $Y(0)=y$, the estimator output $Z=\delta(D_n,y,W)$ consequently has the same
distribution, whereas the two individual counterfactual targets are $y$ and
$1-y$.  Therefore
\begin{align*}
 \frac{R_+(\delta;y)+R_-(\delta;y)}2
 &=\E[(Z-\tfrac12)^2\mid Y(0)=y]
   +\left(y-\tfrac12\right)^2\\
 &\geq\left(y-\tfrac12\right)^2.
\end{align*}
The larger risk is no smaller than their average, and $Z\equiv\tfrac12$
attains equality, proving \eqref{eq:individual-cf-pointwise-lower-bound}.
Integrating over $Y(0)\sim\mathrm{Unif}(0,1)$ gives
$\int_0^1(y-\tfrac12)^2\,dy=1/12$, while taking the essential supremum gives
$1/4$; the same midpoint estimator attains both values.  Estimator
randomization only increases the nonnegative first term above.
\end{proof}

\begin{corollary}[Individual treatment effect]
\label{cor:ite-minimax}
For $\tau=Y(1)-Y(0)$, the two models imply
$\tau_+(y)=0$ and $\tau_-(y)=1-2y$.  Every estimator $\widehat\tau$ corresponds
bijectively to $\widehat Y_1=y+\widehat\tau$, so the sharp conditional,
integrated, and worst-query minimax risks remain respectively
$(y-\tfrac12)^2$, $1/12$, and $1/4$.
\end{corollary}

\clearpage

\section{Conditional-Mean Diagnostics}
\label{app:frozen-mean-checks}
\paragraph{Residual-correction test.}This appendix tests
\cref{prop:irreducible} on three large external corpora, separately from our
benchmarks. The proposition is a population statement, and finite networks need not be unbiased. For a
frozen fitted predictor and $R_t=Y_t-\widehat\mu(X_t)$, the projection
identity gives
\begin{equation}
  \sup_c\frac{\mathbb E\|R_t\|^2-
  \mathbb E\|R_t-c(X_t)\|^2}{\mathbb E\|R_t\|^2}
  =\frac{\mathbb E\|\mathbb E[R_t\mid X_t]\|^2}{\mathbb E\|R_t\|^2},
  \label{eq:residual-oracle}
\end{equation}
so a held-out correction $c$ that reduces residual energy detects a violation.
\label{subsec:frozen-mean-validation}
The five cells use DINOv2-S on AntMaze, HumanoidMaze and PushT,
and V-JEPA~2 on AntMaze and HumanoidMaze. The test uses unweighted
squared-loss predictors and disjoint fitting and report episodes.

\paragraph{Predictor fitting.}We freeze DINOv2-S or
V-JEPA~2, discard any pretraining predictor, and fit a fresh squared-loss mean
head on the official OGBench \citep{ogbench_park2025} AntMaze/HumanoidMaze \texttt{stitch}
corpora; a matched DINOv2 experiment uses the large PushT corpus. OGBench
predictors receive four past frames as context and predict only
$f(O_{t+1})$ from that context and $A_t$; PushT uses its checkpoint-compatible
five-control macro-action. AntMaze, HumanoidMaze, and PushT provide respectively
985k, 1.985M, and 1.441M episode-valid training windows.

DINOv2-S uses \texttt{vit\_small\_patch14\_dinov2} at 224 pixels.

V-JEPA~2 uses \path{facebook/vjepa2-vitl-fpc64-256} with repeated-frame
self-clips.

Humanoid/DINOv2 retains a $2\times2$ patch grid; the other image
features are globally pooled. PushT/DINOv2 uses 16 context frames and five
concatenated controls per macro-step. OGBench uses the official visual
\texttt{stitch} corpora, \texttt{stitch-val} report episodes and registered
third-person camera; PushT uses \texttt{push-t.lance}, reserving 20\% of its training split for
validation after the 80\% train split.
The action-concatenating MLP uses SiLU, LayerNorm and three width-1,024
hidden layers (two for maze V-JEPA~2). AdamW uses zero weight decay, gradient
clipping at 1 and bfloat16 precision. Linear warmup spans 5\% of the stage
(0.5\% for final temporal PushT stages), followed by cosine decay to 1\% of
the peak learning rate. Maze DINOv2 and V-JEPA~2 use peak rate
$3\times10^{-4}$ and batch/device 32/16 (AntMaze/Humanoid DINOv2) or 8
(V-JEPA~2); PushT/DINOv2 uses $3\times10^{-5}$ and batch/device 2.
Final-stage budgets are one epoch and can resume earlier fits. Training uses
stable-pretraining, like the benchmark predictors; seeds and the exact training
history accompany the code release.

\begin{figure}[htbp]
  \centering
  \includegraphics[width=\linewidth]{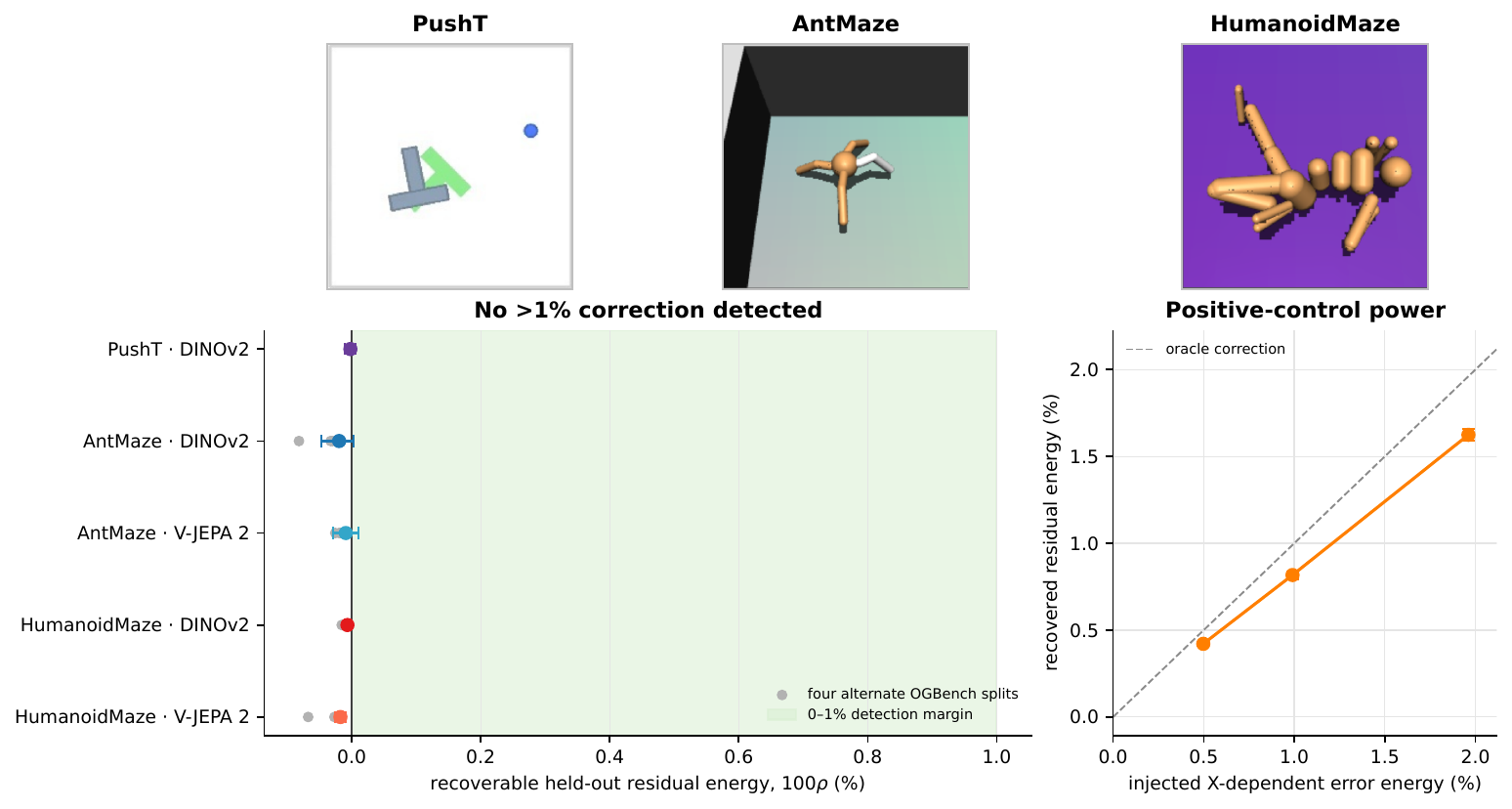}
  \caption{\textbf{Residual corrections and injected-bias controls.} ({\bf Left}) Primary episode-bootstrap 95\%
  intervals (colored) and four alternate OGBench splits (gray); the green band
  is the predeclared 0--1\% correction margin. ({\bf Right}) Deterministic
  $X$-dependent positive controls: realized oracle corrections of $.498\%$,
  $.990\%$, and $1.961\%$ yield trained-procedure gains of $.421\%$, $.817\%$,
  and $1.624\%$, respectively. Thumbnails are exact PushT model input and
  official OGBench \texttt{stitch} renders.}
  \label{fig:conditional-residual-oracle}
\end{figure}

\paragraph{Test information and uncertainty.}Write $X$
for the frozen information supplied to each world model. Holding
the fitted model fixed, we train a separate linear correction and a two-layer,
2048-wide MLP to predict its residual $R=Y-\widehat\mu(X)$ on one set of
episodes, then measure the fraction $\phi$ of residual energy removed on
disjoint episodes. By \eqref{eq:residual-oracle}, the unrestricted optimum is
the normalized energy in $\mathbb E[R\mid X]$; the achieved MLP gain is a lower
bound on that quantity and hence a powered falsification test, not a proof of
the unrestricted moment. We predeclare a 1\% correction margin, cluster-bootstrap
whole episodes, and use one-sided 99\% upper bounds for a Bonferroni family-wise
95\% statement over the five environment--encoder cells. Deterministic
$X$-dependent errors injected at 0.5\%, 1\%, and 2\% calibrate the procedure's
power (\cref{fig:conditional-residual-oracle}, right). No cell admits a
correction above the margin (\cref{tab:frozen-mean-residual}).

\begin{table}[htbp]\centering\small
\setlength{\tabcolsep}{4pt}
\begin{tabular}{llrrccc}
\toprule
Environment & Frozen encoder & Train & Test & $100\phi$ & 99\% & Alt.\ split \\
 & & windows & episodes & [95\% CI] & upper & range \\
\midrule
AntMaze & DINOv2-S & 985k & 250 & $-.019\;[-.048,.004]$ & $.006$ & $[-.082,-.017]$ \\
AntMaze & V-JEPA~2 & 985k & 250 & $-.009\;[-.029,.011]$ & $.014$ & $[-.025,-.016]$ \\
HumanoidMaze & DINOv2-S & 1.985M & 250 & $-.007\;[-.011,-.002]$ & $-.001$ & $[-.016,-.008]$ \\
HumanoidMaze & V-JEPA~2 & 1.985M & 250 & $-.017\;[-.026,-.009]$ & $-.008$ & $[-.067,-.022]$ \\
PushT & DINOv2-S & 1.441M & 2,989 & $-.002\;[-.010,.006]$ & $.007$ & n/a \\
\bottomrule
\end{tabular}
\caption{\textbf{Conditional residual-mean diagnostics.} Gains, intervals, and upper bounds are percentages
of zero-correction residual energy (recall \cref{eq:residual-oracle}). The
primary 95\% interval and one-sided 99\% upper bound are cluster-bootstrapped
over held-out episodes; the latter gives a Bonferroni family-wise 95\% statement over five
cells. Alternate-split ranges are robustness points on the same official
corpus, not independent datasets. Every cell's 99\% upper bound is below the
predeclared 1\% margin. CI: confidence interval.}
\label{tab:frozen-mean-residual}
\end{table}

\paragraph{Distribution shift.}
The primary PushT predictor admits an $18.92\%$ correction on a
chronologically held-out block with shifted simulator-state occupancy.
The primary inference therefore concerns the matched behavior-data support.

\paragraph{Patch-grid check.}As a supplementary
representation check, we repeat the DINOv2 tests while retaining a
$4\!\times\!4$ grid of patch tokens (6,144 target coordinates) instead of global
pooling. The same correction procedure remains nonpositive on AntMaze
($-.0145\%$, 95\% CI $[-.0202,-.0090]\%$) and HumanoidMaze ($-.0159\%$, 95\% CI
$[-.0196,-.0123]\%$), over 302.6M and 609.8M scalar residuals, respectively.
This supplementary check is outside the five-cell family-wise claim.

\clearpage

\section{Teaser Setup}
\label{app:teaser}
\Cref{fig:teaser} is an illustrative toy, not one of our benchmarks: a boat picks
its heading $a'$ and moves one boat length through water that drifts it. The
hidden current is an east-west mean flow plus eddies: a random divergence-free
field, the sum of three waves $10$ to $20$ boat lengths long whose pattern
drifts slowly in time, drawn anew for every episode. The boat moves through the
water while the water moves it, so the drift depends slightly on where the boat
goes. The top panel plots the distance to the center of the view
after the move and the bottom panel the drift off heading, i.e., the course over
ground minus the heading. Gray points are every possible future from the same
state, $300$ currents from the training law for each of $21$ headings, every $18^\circ$ (column
width: outcome density). Where the mean flow and the eddies align, the current
nears the boat's speed and can turn its course by up to $180^\circ$, so the
bottom panel shows the curves and the central $99\%$ of the gray points; the diamond is the one observed outcome, at the
executed heading $a_t$, also marked on the heading axis; the frames are the same
crop of $224$-pixel images. The left and
top-right frames are the two model inputs. The bottom-right frame is the true
counterfactual for heading $-126^\circ$ under the same current, rendered by the
function that produced the model inputs, which reproduces the observed frame
exactly; no model sees it. Arrows and outlines are drawn over the frames and
leave their pixels unchanged. Each arrow is the drift the current causes at its
location over the transition, and each outline draws the rendered hull, with
heading $-126^\circ$, at the position a model predicts. This heading is roughly
opposite the executed one, and both models' position errors there are close to
their medians over the $81$ headings of the curves, every $4.5^\circ$ ($0.52$ and $0.14$ boat lengths,
against medians of $0.47$ and $0.14$).
The JEPA uses
frozen MAE features; the counterfactual world model adds the observed minus predicted
position to its prediction, the exact correction for a uniform current and an
approximate one under eddies. Its
median error is $0.08$ against $0.24$ boat lengths and $3.1^\circ$ against
$17.6^\circ$, and over $1000$ fresh events the drift $R^2$ is $77\%$ against
$-7\%$. The event, the 80th percentile of mean-flow strength among $32$, was
chosen from the simulator before any prediction was computed.

\subsection{Current Field Ablation}
\label{app:boat-field}
\begin{table}[!b]\centering\footnotesize
\setlength{\tabcolsep}{3.5pt}
\begin{tabular}{rrrrrrrrrrrr}\toprule
\multicolumn{2}{c}{Eddies} & & \multicolumn{3}{c}{Relative risk} & \multicolumn{2}{c}{Drift $R^2$ (\%)} & \multicolumn{4}{c}{Observability gates, $R^2$ (\%)} \\ \cmidrule(lr){1-2}\cmidrule(lr){4-6}\cmidrule(lr){7-8}\cmidrule(lr){9-12}
$L$ & Rate & $\rho$ & CF & JEPA & Oracle & CF & JEPA & Drift head & Readout & Oracle & Copy \\ \midrule
1.25 & 0.3 & 0.487 & 0.640 & 1.168 & 1.019 & 20.6 & $-$9.9 & 92.58 & 99.96 & 97.2 & $-$7.3 \\
2.5 & 0.3 & 0.263 & 0.383 & 1.145 & 1.023 & 48.0 & $-$7.6 & 94.39 & 99.96 & 96.1 & $-$4.9 \\
5 & 0.3 & 0.082 & 0.229 & 1.132 & 1.007 & 70.1 & $-$7.7 & 94.81 & 99.96 & 95.6 & $-$4.1 \\
10 & 0.3 & 0.022 & 0.168 & 1.109 & 0.997 & 76.9 & $-$7.1 & 94.95 & 99.96 & 95.4 & $-$4.2 \\
20 & 0.3 & 0.005 & 0.141 & 1.119 & 0.993 & 80.1 & $-$9.6 & 94.90 & 99.96 & 95.3 & $-$4.2 \\
1000 & 0.3 & 0.000 & 0.131 & 1.115 & 0.991 & 80.9 & $-$8.9 & 94.69 & 99.96 & 95.2 & $-$4.3 \\
\midrule
10 & 0 & 0.022 & 0.154 & 1.121 & 0.999 & 78.9 & $-$8.6 & 94.92 & 99.96 & 95.3 & $-$4.2 \\
10 & 1 & 0.021 & 0.153 & 1.109 & 0.991 & 78.9 & $-$8.3 & 94.98 & 99.96 & 95.6 & $-$4.1 \\
10 & 3 & 0.020 & 0.164 & 1.129 & 0.986 & 80.7 & $-$10.9 & 93.84 & 99.96 & 96.4 & $-$4.0 \\
\bottomrule\end{tabular}
\caption{\textbf{Current field ablation on the title-page boat.} Shortest eddy wavelength $L$ (boat lengths), largest phase rate (radians per transition), interaction ratio $\rho$, relative risks of the counterfactual world model (CF), the trained predictor (JEPA) and the oracle predictor without the outcome, and drift $R^2$ over $1000$ fresh events with $16$ alternative headings each. Gates (\cref{app:benchmark-selection-summary}): held-out $R^2$ of the drift head and of the state readout (lowest coordinate), and position $R^2$ of the oracle predictor and of factual copying (highest coordinate). The oracle predictor exceeds $60\%$ and the north-south drift is recovered below $95\%$ in every row, so these results are diagnostic.}\label{tab:boat-field}
\end{table}
The ablation of \cref{sec:where-transport} keeps the boat, the mean flow and the eddy speed of the title page and changes one property of the eddies at a time. Each of the three waves has a wavelength drawn uniformly in $[L,2L]$ boat lengths and a phase that drifts at a rate drawn uniformly between minus and plus the stated phase rate; the title-page current has $L=10$ and phase rate $0.3$. All conditions use the same random draws, so changing $L$ or the phase rate rescales the same fields, and at $L=1000$ the eddies are uniform over the region the boat visits. Each condition collects $128$k factual episodes of two transitions and fits on frozen MAE-B features a state readout, a drift head for the realized two-dimensional drift of the current, and two factual predictors with weight decays $10^{-4}$ and $0.3$, of which it keeps the one with the lower factual validation loss. Each of the $1000$ fresh evaluation events is scored at $16$ alternative headings, every $22.5^\circ$. The oracle predictor without the outcome and residual reuse with exact means use conditional means over $512$ currents from the training law, shared by all headings of an event, and $\rho$ is estimated from the same draws. Relative risks divide the summed squared position error by the summed conditional variance of the position given the source state and heading, which is the risk of the oracle predictor, so this predictor scores one up to Monte Carlo error. The readout recovers the position at $99.96\%$ and the east-west drift at $97.5$--$97.9\%$, and factual copying, from the readout or from the true state, stays below $0\%$, but the north-south drift is recovered at $92.6$--$94.98\%$ and the oracle predictor reaches $86$--$97\%$ position $R^2$, so every condition is diagnostic (\cref{tab:boat-field}).

\end{document}